\documentclass[acmlarge]{acmart}
\usepackage{booktabs} 
\usepackage{makecell}
\usepackage{arydshln}
\usepackage{pifont}

\usepackage{colortbl} 
\usepackage[ruled]{algorithm2e} 

\SetAlFnt{\small}
\SetAlCapFnt{\small}
\SetAlCapNameFnt{\small}
\SetAlCapHSkip{0pt}

\usepackage[textsize=tiny]{todonotes}
\usepackage{xspace}
\usepackage{xcolor}
\newcommand{\blue}[1]{\textcolor{black}{#1}}

\begin{document}

\title{Camera Trajectory Generation: A Comprehensive Survey of Methods, Metrics, and Future Directions}


\author{Zahra Dehghanian}
\affiliation{%
  \institution{Sharif University of Technology}
   \country{Iran}
  }
\email{zahra.dehghanian97@sharif.edu}

\author{Pouya Ardekhani}
\affiliation{%
  \institution{Sharif University of Technology}
     \country{Iran}
}
\email{pouya.ardehkhani02@sharif.edu}
\author{Amir Vahedi}
\affiliation{%
 \institution{Sharif University of Technology}
    \country{Iran}
 }
 \email{amir.vahedi123@sharif.edu}

\author{Hamid Beigy}
\affiliation{%
 \institution{Sharif University of Technology}
    \country{Iran}
 }
 \email{beigy@sharif.edu}
 
 \author{Hamid R. Rabiee}
\affiliation{%
 \institution{Sharif University of Technology}
    \country{Iran}
 }
 \email{rabiee@sharif.edu}

\renewcommand\shortauthors{Dehghanian et. al.}

\begin{abstract}

Camera trajectory generation is a cornerstone in computer graphics, robotics, virtual reality, and cinematography, enabling seamless and adaptive camera movements that enhance visual storytelling and immersive experiences. Despite its growing prominence, the field lacks a systematic and unified survey that consolidates essential knowledge and advancements in this domain. This paper addresses this gap by providing the first comprehensive review of the field, covering from foundational definitions to advanced methodologies. We introduce the different approaches to camera representation and present an in-depth review of available camera trajectory generation models, starting with rule-based approaches and progressing through optimization-based techniques, machine learning advancements, and hybrid methods that integrate multiple strategies. Additionally, we gather and analyze the metrics and datasets commonly used for evaluating camera trajectory systems, offering insights into how these tools measure performance, aesthetic quality, and practical applicability. Finally, we highlight existing limitations, critical gaps in current research, and promising opportunities for investment and innovation in the field. This paper not only serves as a foundational resource for researchers entering the field but also paves the way for advancing adaptive, efficient, and creative camera trajectory systems across diverse applications.


\end{abstract}

%

%

\keywords{Camera Trajectory Generation, Automatic Camera Control, Virtual Cinematography}

\maketitle

\section{Introduction}
Virtual cinematography involves the cinematic projection of scenes occurring in a 3D graphical environment onto a flat screen, with a virtual camera serving the role of a physical one. A key component of virtual cinematography is camera trajectory generation. It is a pivotal area of research in computer graphics, robotics, virtual reality, and cinematography \cite{elson2007lightweight,pandya2014review}; where precise and adaptive camera movements significantly enhance user experiences and address both aesthetic and practical demands. Informally, camera trajectory refers to the continuous path a camera follows in three-dimensional space, encompassing its position, orientation, and motion over time \cite{liu2024splatraj}. The formal definition is provided in Section \ref{representation}. This process entails designing and calculating camera paths by integrating mathematical models, computational methods, and aesthetic principles, ensuring the motion is seamless, adaptable, and purpose-driven within dynamic settings.

Historically, camera trajectory generation has evolved from basic rule-based systems \cite{he1996virtual, christie2009camera} rooted in traditional cinematographic principles to sophisticated, data-driven models that integrate machine learning and real-time adaptability \cite{burg2021real}. This evolution has been driven by the increasing demands for computational efficiency, dynamic scene responsiveness, and aesthetic coherence across both virtual and real-world contexts. Various representation and modeling approaches for camera trajectory generation, such as the 7-degree-of-freedom (7-DOF) framework \cite{Christie2008camera}, Toric space \cite{Lino2015}, and drone-specific adaptations \cite{galvane2018directing}, have been proposed to represent the camera in distinct ways. Each approach offers specific advantages and limitations, rendering them suitable for particular applications depending on factors such as flexibility, computational efficiency, and the specific requirements of the given task. Recent advancements, including the application of deep learning and emerging trends like diffusion models, have facilitated the development of adaptive and context-aware systems, significantly enhancing the capabilities of camera trajectory generation \cite{massaglia2023dreamshot}.
Beyond its technical contributions, camera trajectory generation has broad practical applications, spanning autonomous drones \cite{nageli2017multi}, surveillance systems \cite{fiengo2006optimal}, gaming \cite{burelli2011towards}, and film production \cite{yang2024direct}. 

While these advancements have significantly enhanced virtual cinematography, challenges persist. These include the seamless integration of computational, perceptual, and aesthetic constraints, which are crucial for further improving user immersion and visual experiences, visual storytelling, and the adaptability of camera systems in dynamic scenarios.
By aligning artistic vision, technical precision, and user-focused design, research in camera trajectory generation bridges technology and art, offering solutions to real-world challenges while elevating creative practices.

A notable gap in the current body of research is the absence of a comprehensive survey that consolidates the diverse methodologies and techniques proposed in this field. To address this, we present a detailed survey that unifies foundational principles, state-of-the-art (SOTA) methodologies, and cutting-edge advancements. 
It focuses on the theoretical and methodological advancements in camera trajectory generation, emphasizing SOTA techniques and foundational principles. The research spans diverse applications in computer graphics, virtual reality, robotics, and cinematography
By analyzing research from the past 20 years, it synthesizes key methodologies, emerging trends, and unresolved challenges to guide future innovation.

We systematically reviewed related work from reputable sources, including peer-reviewed journals, conference proceedings, and technical reports, using academic databases such as IEEE Xplore, ACM Digital Library, and SpringerLink with keywords 'camera trajectory generation,' 'automatic camera control,' and 'virtual cinematography' to ensure wide-ranging coverage. This method facilitated a comprehensive integration of theoretical advancements and practical applications across diverse fields.

The remainder of this paper is organized as follows. Section \ref{representation} examines camera trajectory representation frameworks across three abstraction levels, addressing trade-offs between usability and precision while highlighting strategies for balancing expressiveness, computational efficiency, and user-system compatibility. Section \ref{camera_movement_sys} focuses on camera movement systems and their integration with computational frameworks. Section \ref{algorithms_section} discusses trajectory generation techniques, emphasizing real-time adaptability and aesthetic considerations. Section \ref{metric_section} reviews evaluation metrics, ranging from quantitative measures to qualitative assessments, while Section \ref{datasets_section} surveys key datasets and their contributions to the field. Section \ref{limitations_section} synthesizes findings and identifies open research challenges, paving the way for future advancements. Finally, the conclusion summarizes key insights and underscores the significance of continued innovation in camera trajectory generation.

\section{Representation} \label{representation} 
Camera trajectory generation involves creating a shot, or a sequence of shots, that form a scene under specific constraints. These constraints must be translated into a unique set of camera parameters specifying its position, orientation, and movement over time \cite{zhang2021camera}. Managing these parameters, in addition to time, is tedious and overly complex for non-technical users. Utilizing high-level descriptions, such as natural language-like shot annotations, offers a more accessible and user-friendly way for non-experts to specify constraints compared to manually managing precise camera parameters like position and orientation over time.

\blue{Camera intrinsics including focal length, focal distance, aperture, and camera extrinsics including position and orientation are critical parameters in camera modeling and image formation \cite{zhang2021camera}. Focal length determines the magnification and field of view of a camera lens, while focal distance refers to the distance between the lens and the focused subject. Aperture controls the amount of light entering the lens and affects depth of field \cite{zhang2021cameraCal}. Extrinsic parameters define the camera’s position and orientation relative to a world coordinate system \cite{zhang2021cameraExt}, whereas intrinsic parameters describe the internal characteristics of the camera, such as focal length and principal point \cite{zhang2021cameraCal}. Together, these parameters enable precise camera calibration and projection modeling}

The constraint representation should be as compact and expressive as possible, capable of covering all existing and potential scenarios. A key challenge lies in establishing a one-to-one correspondence between the intermediate representation and precise camera parameters. At higher abstraction levels, certain details might be omitted, leading to ambiguity where a single representation could correspond to multiple parameter configurations. Several works have addressed automating the parameter retrieval process, contributing the automatic conversion of shot annotations into fully realized shots \cite{ronfard2015prose, louarn2018automated, louarn2020interactive}. 

We can categorizes representations into three levels of abstraction First, high-level representations use natural language for intuitive descriptions. Second, mid-level representations rely on structured formal languages. Third, low-level representations employ precise mathematical definitions for detailed control.
There is an inherent trade-off between the expressiveness and usability of camera trajectory representations and their ease of conversion into precise camera parameters. As the level of abstraction moves closer to natural language, the representation becomes easier to use and more intuitive for non-specialists \cite{liu2024chatcam}. However, this increased accessibility often comes at the cost of precision and the complexity of converting the representation into an accurate camera trajectory. Conversely, lower-level representations provide a higher degree of precision and are more straightforward to translate into real camera parameters but are harder for humans to understand and use \cite{ronfard2015prose, Galvane2015, Lino2012}. Striking the right balance between ease of use and technical rigor is essential for designing representations that meet the needs of both human users and computational systems.


These levels of abstraction will be further elaborated upon in the subsequent sections. The completeness and parameter retrieval of each abstraction level are also examined.
\subsection{High-Level Natural Language Representation} 
High-level natural language representation refers to employing natural language descriptions to specify camera trajectories in an intuitive and accessible manner. This approach leverages the expressiveness of human language to allow users, including non-technical ones, to define constraints and desired outcomes for camera movements without requiring direct manipulation of complex mathematical parameters or low-level settings.
Recent advancements in the field of large language models (LLMs) have significantly enhanced their capacity to understand natural languages, leading to notable achievements such as LLaMA 3, GPT-4o, and Gemini 1.5 \cite{ dubey2024llama, hurst2024gpt, reid2024gemini}. One promising approach involves utilizing high-level natural language descriptions to generate desired camera trajectories, anticipating that the system will create these trajectories in virtual or real environments based on the constraints specified in the linguistic descriptions. While the expressiveness of natural languages ensures the completeness of this approach, retrieving exact parameters remains challenging due to the complex nature of language comprehension by computers. This challenge can be mitigated by leveraging emerging LLMs \cite{liu2024chatcam, he2024cameractrl}.

The ChatCam model \cite{liu2024chatcam} is an example from this family of approaches, aiming to enable camera control through natural language interactions. The approach employs CineGPT, a GPT-based autoregressive model, for text-conditioned camera trajectory generation, complemented by an Anchor Determinator for precise trajectory placement.

Also, CameraCtrl \cite{he2024cameractrl}, a plug-and-play module enables precise camera control in text-to-video generation by using this representation. These module can integrate with existing video diffusion models, such as AnimateDiff \cite{guo2023animatediff}, without affecting frame quality or temporal consistency. 

Hou et al. \cite{hou2024training} introduce CamTrol, a training-free framework for camera control in video diffusion models. The approach leverages 3D point cloud representations for explicit camera motion modeling and employs noise layout priors to guide video generation. 

\subsection{Mid-Level Shot Annotation Representation}
A formal language offers an alternative approach to representing camera trajectories, providing a structured and rule-based method for defining descriptions and restricting the descriptions to adhere to this language, instead of relying on high-level natural language.
The completeness of this approach highly depends on the formal language used to describe the constraints. On the other hand, because we are dealing with formal language, there is a formal grammar representing the language, thus shot annotations can be easily derived from the grammar to retrieve the parameters easily and quickly \cite{bares2000virtual, van2009movie, liang2012script, ronfard2015prose, louarn2018automated, louarn2020interactive}. 
Most contributions in this category focus on linguistic specifications for generating camera trajectories, primarily utilizing mid-level shot annotations that are later translated into fully realized shots.

The Movie Script Markup Language (MSML) \cite{van2009movie} is a camera specification language designed to provide a structured format for screenplay narratives in television and film production. It incorporates timing and animation models for synchronization and production control and uses XML serialization. Developed in collaboration with industry professionals, MSML has been implemented in proof-of-concept systems, showcasing its applicability to practical scenarios.

\begin{figure}[h]
    \centering
        \includegraphics[width=0.6\linewidth]{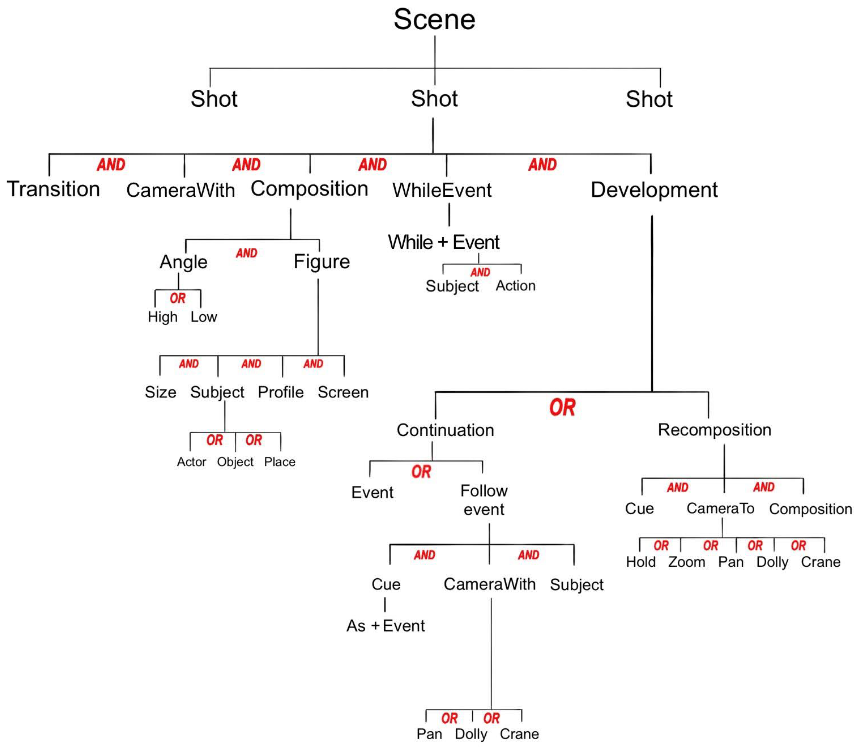}
    \caption{Tree representation of the PSL grammar \cite{ronfard2015prose}.}
    \label{fig:and_or_grammer}
\end{figure}

The Prose Storyboard Language (PSL) \cite{ronfard2015prose} is a method designed for annotating movie shots using a formal context-free language and its associated grammar. PSL enables the structured annotation of shots, providing a systematic approach to describing scenes through a well-defined formal language. The grammar of PSL forms an AND-OR tree, as illustrated in Figure \ref{fig:and_or_grammer}.

 Any sentence in PSL must adhere to the same grammar. Like any formal grammar, there are multiple terminals and non-terminals. Terminals in PSL are divided into two categories: generic terminals and specific terminals. Generic terminals include terms such as “pan,” “dolly,” and “enter.” Specific terminals include character names, places, and objects. Non-terminals consist of categories of shots, image composition, image development, and other elements.

\begin{figure}[h]
    \centering
        \includegraphics[width=0.6\linewidth]{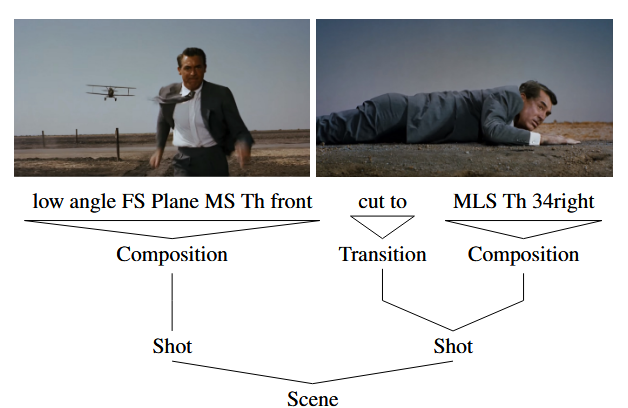}
    \caption{Prose storyboard language description of two iconic shots in Alfred Hitchcock’s North By Northwest \cite{ronfard2015prose}.}
    \label{fig:psl_example_1}
\end{figure}

To describe an entire movie, a unique PSL sentence is assigned to each shot. Every PSL sentence address two properties of the shot: spatial structure and temporal structure. Spatial structure focuses on the composition of an individual movie frame, while temporal structure captures events in a sequence of frames. Therefore, each shot can be described with a complete PSL sentence that includes at least one composition and an arbitrary number of screen events. An example of PSL description is shown in Figure \ref{fig:psl_example_1} 

The Prose Storyboard Language (PSL) is intended to represent a director's vision by providing a method for annotating shots across pre-production, production, and post-production stages \cite{ronfard2015prose}. PSL allows for describing existing movies as an ordered sequence of sentences, one per shot, enabling parameter retrieval based on its formal grammar. While the structured nature of PSL simplifies parameter retrieval, the absence of a systematic approach for extracting parameters from PSL sentences is identified as a limitation.

Following PSL, Film Editing Patterns (FEP) \cite{wu2018thinking} is a language designed to formalize film editing practices, supporting virtual cinematography by encoding constraints on elements such as shot size, angle, and actor positioning. The framework facilitates automated style analysis and prototyping of creative 3D sequences. Evaluations involving professionals and amateurs suggest that FEP is particularly useful for novice users, providing pedagogical and practical benefits. However, the framework's flexibility for expert users is limited, and there is potential for enhancing editing functions and enabling more customizable patterns.

Even though both PSL and FEP are utilized for shot creating, they differ significantly in their methodology and focus. The FEP language emphasizes cinematographic visual properties, such as shot sizes, angles, and actor layouts, to formalize film editing techniques and improve creative workflows in 3D animation by encoding stylistic patterns (e.g., intensify, opposition) and their application in editing tools \cite{wu2018thinking}. Meanwhile, the PSL adopts a descriptive syntax to provide structured, human-readable annotations for each shot, capturing spatial and temporal structures, with particular attention to shot development and transitions. PSL enables a more granular representation of events and compositions, catering to both manual annotation and machine interpretation \cite{ronfard2015prose}.

Louarn et al. proposed an extension of the Prose Storyboard Language (PSL) to facilitate automated staging in virtual cinematography \cite{louarn2018automated}. The extension introduces enhancements such as camera identification, enabling the specification of complex constraints involving multiple cameras, and scene identification, which supports the description of continuity constraints for character and camera placement and orientation. Additionally, it incorporates three generic terminals—entity, object, and region—along with associated constraints, expanding PSL’s expressive capacity for representing and staging complex scenes.

The extended PSL representation has been applied to automate camera staging in 3D virtual environments through pruning the Potential Location-Rotation Set (PLRS) \cite{louarn2018automated}. By incorporating additional features into the traditional PSL, the extended language accommodates a broader range of constraints. However, the system faces limitations, including restricted support for multiple target constraints and challenges in dynamic scene handling. It is currently limited to constraints between two entities and requires further development to effectively express complex cinematographic rules and evaluate constraints over time for moving entities.

In subsequent work, Louarn et al. utilized the same extended PSL for interactive staging and shooting in virtual cinematography \cite{louarn2020interactive}. They introduced a system that takes a 3D virtual environment and constraint specifications in extended PSL as inputs, then selects the position and orientation of entities in the scene as output. The system operates in a loop of three stages.

The process involves three key stages: the Pruning Stage refines each entity's PLRS using a Geometric Pruning Operator, producing a dependency graph. The Elicitation Stage utilizes this graph and each entity’s domain to generate candidate solutions by sampling within specified constraints. In the Interactive Stage, users can modify entities and navigate the environment, triggering a new elicitation phase to ensure updated solutions meet requirements. This approach's advantage is its interactive capability, absent in prior methods. However, it regenerates the dependency graph with each interaction, disrupting solution continuity. Additionally, like other constraint-based methods, it struggles to identify conflicting constraints when a solution cannot be found, limiting its effectiveness in such cases.

\subsection{Low-Level Mathematical Representation}
At the lowest level of abstraction, camera trajectories can be described using mathematical representations. Methods such as 7-DOF \cite{Christie2008camera} and Toric space \cite{Lino2012} can be employed to provide precise and mathematically sound descriptions of camera movements. These approaches ensure accuracy and rigor, making them ideal for scenarios requiring fine-grained control over camera behavior. In the following subsection, we will delve into the details of these methods, exploring their principles, applications, and limitations.

\subsubsection{7-DOF Modeling}

\begin{figure}[h]
    \centering
    \includegraphics[width=0.3\linewidth]{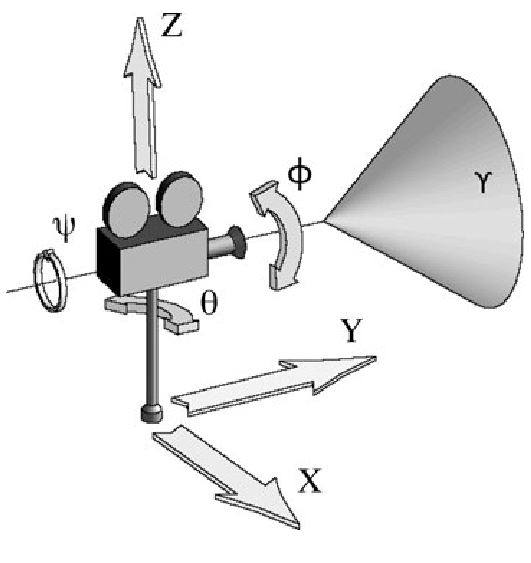}
    \caption{A simple camera model based on Euler angles; tilt ($\phi$), pan ($\theta$), and roll ($\psi$) \cite{Christie2008camera}.}
    \label{fig:camera-model}
\end{figure}

Camera modeling in computer graphics often aims to address the challenges of dynamic environments and precise visual representation \cite{Christie2008camera}. One of the most well-known and widely used low-level representations is the 7-DOF model \cite{Christie2008camera}, which includes three parameters for Cartesian coordinates $(x_c, y_c, z_c)$ \cite{stewart2012calculus}, three Euler angles $(\phi_c, \theta_c, \psi_c)$ \cite{foley1996computer}, and one intrinsic parameter for the field of view $\gamma_c$ \cite{hartley2003multiple}, as shown in Figure \ref{fig:camera-model}. This approach was motivated by the complexity of ensuring accurate camera placement while accommodating constraints like occlusion and motion in multidimensional datasets. Occlusion constraints are designed to ensure that critical elements in a scene remain visible and are not blocked by other objects. Motion constraints ensure that the camera's movement is smooth and logical, especially in dynamic scenes where objects or the environment may change over time. By modeling the camera with these degrees of freedom, the authors aimed to create a flexible framework for visualization and multimodal systems \cite{eisenhauer2008degrees}. 

By explicitly accounting for relationships between visual elements, spatial configurations, and user perspectives, the framework surpasses conventional models in adaptability and precision, dynamically maintaining visual coherence and contextual alignment in complex, interactive systems \cite{Christie2008camera}. 
This adaptability is achieved through a mathematical representation that transforms world coordinates into a local camera basis, as shown in Equation \ref{eq:1}:

\begin{equation} \label{eq:1}
\begin{pmatrix}
x' \\
y'
\end{pmatrix}
=
P(\gamma_c) \cdot T(x_c, y_c, z_c) \cdot R(\phi_c, \theta_c, \psi_c) \cdot 
\begin{pmatrix}
x \\
y \\
z \\
1
\end{pmatrix},
\end{equation}

where $x', y'$ are the projected coordinates on the 2D screen, and $(x, y, z)$ represent the object's 3D coordinates in the world space. Here, $R$ incorporates the Euler angles, $T$ translates the camera's position, and $P$ adjusts the projection based on the field of view. 

The 7-DOF camera model excels in flexibility and precision, using its degrees of freedom in position, orientation, and field of view to address challenges like occlusion avoidance and aligning visual elements with linguistic references. By dynamically positioning the camera to maintain unoccluded views and accurately linking spatial configurations with linguistic descriptors, it proves invaluable for multimodal The 2D manifold representation revolutionizes camera composition by transforming the problem into an efficient algebraic framework. This framework represents the solution space as a spindle torus, a specific type of toroidal surface characterized by its unique topology and geometry. The spindle torus arises naturally in problems where a point or subject is constrained by angles and distances relative to a central axis or plane, such as in camera positioning for visual composition.

\subsubsection{Spherical Surface}
\begin{figure}[h]
    \centering
    \includegraphics[width=0.3\linewidth]{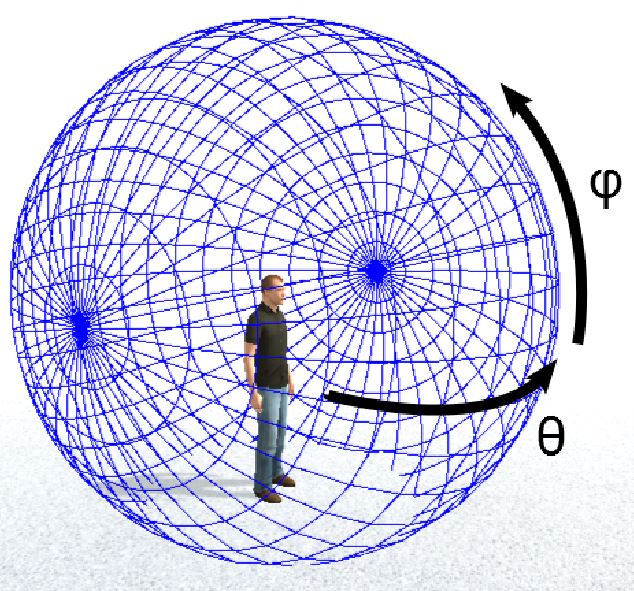}
    \caption{Spherical surface used to model a camera for single-target configurations, showing the character's vantage angles $(\theta, \phi)$ in spherical coordinates \cite{Galvane2015}.}
    \label{fig:spherical-surface}
\end{figure}

As shown in Figure \ref{fig:spherical-surface}, this approach enables smooth transitions between initial and final camera configurations while preserving framing constraints \cite{galvane2015camera}. The uniqueness lies in its algebraic simplicity and ability to handle single-target configurations effectively, which is particularly useful in scenarios requiring precise tracking of a single moving subject.

The spherical surface model's primary advantage is its computational efficiency, as it reduces the complexity of determining optimal camera positions for single-target tracking. Additionally, it facilitates smoother transitions compared to more generalized manifold surfaces. However, a notable limitation is its restriction to single-character scenarios, as it cannot handle interactions or occlusion with multiple targets. This limitation makes it less suitable for more dynamic or multi-character environments.

In summary, the drone-specific spaces offers a tailored approach for aerial cinematography but faces challenges in balancing computational efficiency with the demands of dynamic drone operation, particularly in cluttered or rapidly changing environments. Future work could involve developing adaptive algorithms that dynamically adjust safety parameters based on environmental inputs or using predictive control models for smoother transitions between camera configurations. Exploring lightweight neural network models for real-time decision-making and collision avoidance could further enhance the utility and flexibility of this method in drone cinematography.

\subsubsection{Toric Space}
The concept of 2D manifolds has revolutionized camera composition by reframing it as an algebraic problem, enabling more efficient solutions \cite{Lino2012}. This method models the solution space as a spindle torus, a distinctive toroidal structure with unique geometrical and topological features. The spindle torus naturally emerges in scenarios where a point or object is constrained by angular and distance parameters relative to a central plane or axis, which is particularly relevant in tasks like camera positioning for composition.

Within this framework, the spindle torus is described using angular parameters $\phi$ and $\theta$, forming a continuous surface that represents potential camera configurations adhering to fixed distance and alignment constraints with respect to the subject. This organized representation streamlines the process of identifying optimal camera parameters, eliminating the need for computationally heavy iterative approaches \cite{Lino2012}. Unlike general-purpose 7-DOF methods, which are applicable in environments without predefined targets, Toric spaces rely on the presence of targets for functionality. This target dependency facilitates precise subject placement within the frame by leveraging the geometrical properties of the spindle torus, significantly lowering computational demands. Moreover, the algebraic model tackles Blinn’s spacecraft problem \cite{Blinn1988} by optimizing camera orientation and positioning under constraints like fixed distance and direction. Such methods are crucial for applications that demand detailed and efficient visual composition.

In the 2D manifold representation model, the camera position $P_{\phi,\theta}$ is parameterized by two angles: $\phi$, defining the vertical plane, and $\theta$, defining the arc within this plane. The relationship is mathematically expressed in Equation \ref{eq:2}.
\begin{equation} \label{eq:2}
P_{\phi,\theta} = (q_{\phi} \cdot \vec{IO_0}) + \vec{I},
\end{equation}

where $q_{\phi}$ represents the rotation by $\phi$ radians around the axis $\vec{AB}$, $\vec{IO_0}$ is the vector connecting the midpoint $\vec{I}$) to the center of the inscribed circle $\vec{O_0}$ (specifically for$\phi = 0$), and $\vec{I}$ is the midpoint of the segment joining the two subjects. Here, $\vec{I}$ and $\vec{O_0}$ are not parameters but derived entities based on the geometric configuration: $\vec{I}$ is explicitly the midpoint of segment $\vec{AB}$, and $\vec{O_0}$ is the center of the inscribed circle determined by the 2D manifold constraints. This representation encapsulates all feasible camera positions that satisfy the exact on-screen projection constraints.

By reducing the search space from six dimensions to two (2-DOF), the method significantly lowers computational costs, making it highly efficient for real-time and complex environments. Its parametric nature supports integrating visual properties like vantage angles and object sizes, enhancing versatility. However, its focus on exact on-screen compositions may limit flexibility in scenarios with broader or competing constraints \cite{Lino2012}.

\begin{figure}[h]
    \centering
    \includegraphics[width=0.5\linewidth]{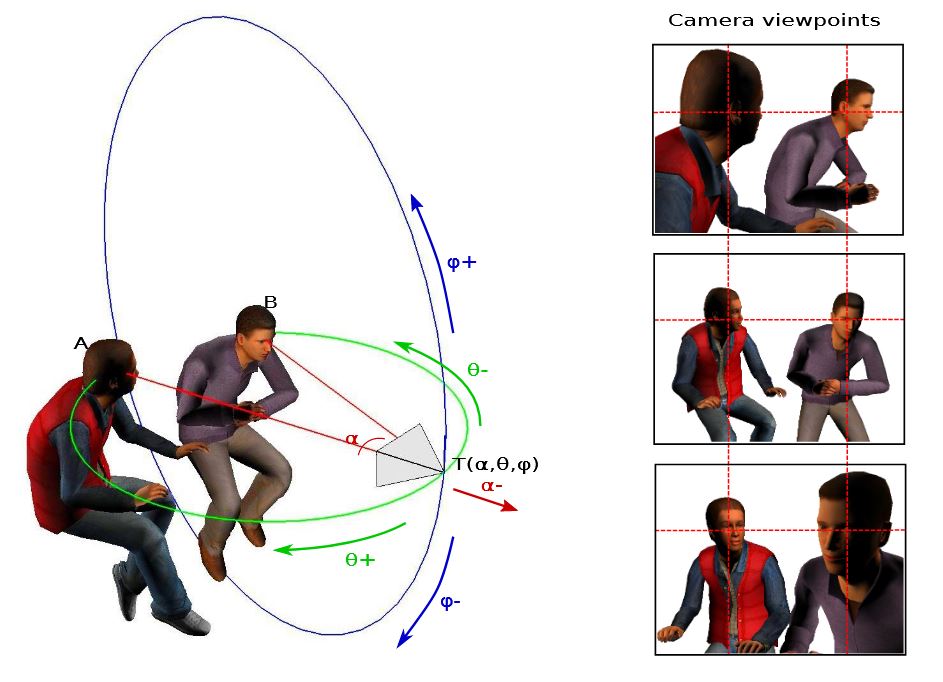}
    \caption{Representation of the Toric space. The manifold is parametrized by $(\alpha, \theta, \phi)$, defining camera positions around two targets \cite{Lino2015}.}
    \label{fig:toric-space-fig}
\end{figure}

The Toric space model is a generalization of the 2D manifold representation \cite{Lino2012} into a three-dimensional search space \cite{Lino2015} defined by the triplet of Euler angles $(\alpha, \theta, \phi)$ describe horizontal and vertical angles around the targets. This representation simplifies the camera control problem by reducing a 7-DOF search space to a 4-DOF space for scenarios involving two targets. Using this model, any camera positioned on this manifold can view the two targets with specified on-screen compositions. The conversion of a camera's Toric representation $T(\alpha, \theta, \phi)$ to its Cartesian representation $C(x, y, z)$ is given by the Equation \ref{eq:3}.

\begin{equation} \label{eq:3}
C = A + (q_\phi \cdot q_\theta \cdot AB) \cdot \sin(\alpha + \theta/2),
\end{equation}

where $q_\phi$ and $q_\theta$ are quaternions representing rotations by $\phi$ and $\theta$ respectively. Quaternions are a mathematical tool for representing 3D rotations. They are defined as a set of four numbers $q = (w, x, y, z)$, where $w$ is the scalar part, and $x, y, z$ form the vector part .The vector $AB$ is derived from the difference in the positions of the two targets, and $A$ corresponds to the location of the first target. As shown in Figure \ref{fig:toric-space-fig}, this model provides a compact and computationally efficient means of defining camera placement while maintaining visual properties.

The Toric space model was developed to overcome limitations in earlier camera control frameworks, such as their reliance on exact on-screen positioning and inefficiencies in handling soft framing \cite{Lino2012}. By reducing the complexity of the search space and enabling rapid computation of camera positions, the Toric space provides a more versatile approach to virtual camera control. It directly incorporates visual properties like vantage angles (relative viewing angle around a target, defined by a reference direction and a permissible deviation, used to specify the desired orientation of a camera toward the target.), target sizes, and on-screen positions within its parameterization, addressing many challenges of prior models. However, its reliance on point-based target representations restricts its ability to manage occlusion or complex multi-target relationships \cite{Lino2015}. Extending the model to include occlusion-aware strategies or adaptive parameterization could improve its applicability in diverse scenarios. A line of research for future extension, may integrate machine learning-based predictive models for dynamic framing or combine the Toric space with real-time depth analysis to enhance its effectiveness in intricate virtual environments.

\subsubsection{Drone Toric Space} \label{DTS}
Unlike static or ground-based camera setups, drones operate in three-dimensional airspace and must account for different factors. These complexities demand a specialized framework that not only ensures compliance with cinematographic principles but also integrates the physical realities of drone navigation \cite{galvane2018directing}. The Drone Specific Space addresses these challenges by extending conventional camera models with additional parameters tailored to the specific requirements of drone cinematography, offering a robust solution for dynamic and aerial filming scenarios.

The Drone Toric Space (DTS) extends the Toric space model to address the unique requirements of cinematographic drone control because it builds upon the foundational principles of the Toric Space while incorporating additional considerations for drone-specific constraints. It introduces a 7D parameterization $q(x, y, z, \rho, \gamma, \psi, \lambda)$, where $(x, y, z)$ denotes the drone's position in Cartesian space, $(\rho, \gamma, \psi)$ are the Euler angles for roll, pitch, and yaw, and $\lambda$ defines the gimbal tilt. This model integrates physical constraints like collision avoidance and minimum safety distances with cinematographic principles such as framing and smooth transitions, ensuring physically feasible and visually coherent drone movements \cite{galvane2018directing}.

\begin{figure}[h]
    \centering
    \includegraphics[width=0.3\linewidth]{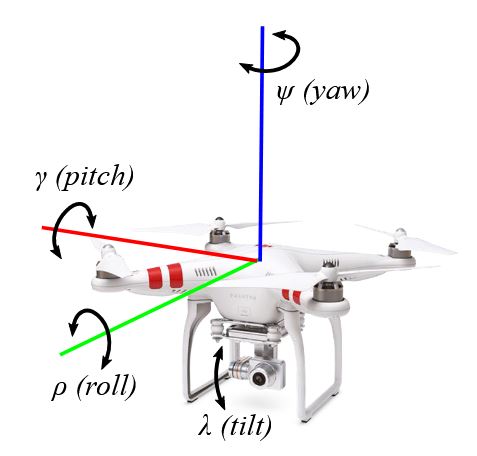}
    \caption{Drone configuration in the DTS model, showcasing its 7D parameterization \cite{galvane2018directing}.}
    \label{fig:dts-config}
\end{figure}

Figure \ref{fig:dts-config} demonstrates this configuration, highlighting how safety and physical constraints are embedded. Unlike the Toric space, the DTS incorporates collision avoidance by enforcing a minimum safety distance around targets and maintaining feasible trajectories through dynamic path planning. This ensures physical safety while accommodating real-time cinematographic adjustments. 

\begin{figure}[h]
    \centering
    \includegraphics[width=0.5\linewidth]{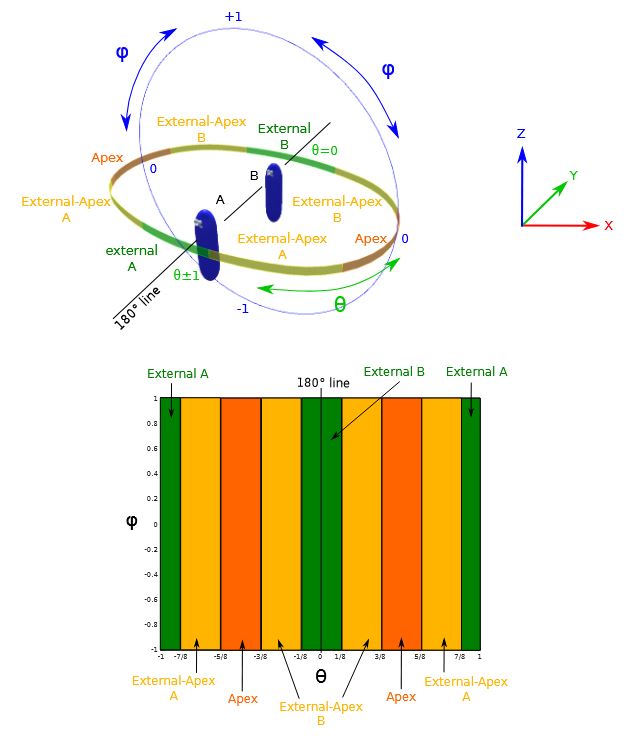}
    \caption{Drone Toric Space parameterization, highlighting regions for camera positioning and framing \cite{galvane2018directing}.}
    \label{fig:dts-features}
\end{figure}

The DTS model introduces significant advancements for drone cinematography by offering predefined camera regions (e.g., external, apex) for framing targets dynamically, as illustrated in Figure \ref{fig:dts-features}. These regions help maintain visual consistency while allowing smooth transitions between cinematic shots. The system also ensures collision-free paths and adaptability to environmental changes, making it ideal for real-time filming of moving targets. However, its complexity increases computational demands, and its reliance on fixed parameters like safety distances may limit flexibility in highly dynamic or cluttered environments \cite{liu2017planning}. Nevertheless, the DTS remains a robust solution for managing drone trajectories while balancing physical and visual constraints.

To adapt the Toric space framework for real-time environments, \cite{burg2020real} introduces several critical enhancements focused on computational efficiency and dynamic adaptability. Traditional Toric space methods faced significant challenges in processing dynamic scenes, as visibility computations often relied on computationally intensive ray-casting \cite{roth1982ray} or static pre-computation \cite{oskam2009visibility}, which made real-time application impractical. The improvements in this work involve the use of GPU-accelerated techniques, such as shadow mapping \cite{williams1978casting, everitt2001hardware} and anisotropic blurring \cite{galvane2015camera}, to compute visibility and occlusion anticipation in Toric space efficiently. By utilizing GPU-based techniques, the system generates an "anticipation map" to predict occlusions within a specified time frame. This map, paired with a motion model, enables dynamic camera adjustments that ensure smooth transitions, minimize visibility loss, and allow Toric space to function effectively in real-time, even in complex, highly occluded scenes.

\subsubsection{Pl\"{u}cker Coordinates}
In this approach, a camera is represented using Plücker coordinates \cite{zhang2024raydiffusion}, which describe it as a collection of rays instead of relying on conventional global parameters. Each ray is characterized by its direction and moment vectors, providing a flexible and detailed way to model cameras. This representation supports over-parameterization, where additional variables enable modeling of both classical and non-perspective camera systems, including those with complex imaging geometries \cite{grossberg2001general, schops2020having}. By assigning each pixel to a corresponding ray, the method effectively utilizes localized features, offering greater granularity compared to traditional models.

The motivation for adopting this representation arises from the challenges posed by sparsely sampled views, where establishing reliable correspondences between image features is often difficult \cite{Snavely2006PhotoTourism,Zhou2023SparseFusion}. By representing cameras as a collection of rays, this method complements transformer-based architectures, which excel in set-level processing and patch-wise analysis \cite{Dosovitskiy2021Transformers}. Furthermore, this approach naturally accommodates probabilistic modeling, an essential capability for addressing uncertainties inherent in sparse-view pose estimation tasks \cite{Wang2023PoseDiffusion}.

\begin{figure}[h]
    \centering
    \includegraphics[width=0.7\linewidth, height=0.4\linewidth]{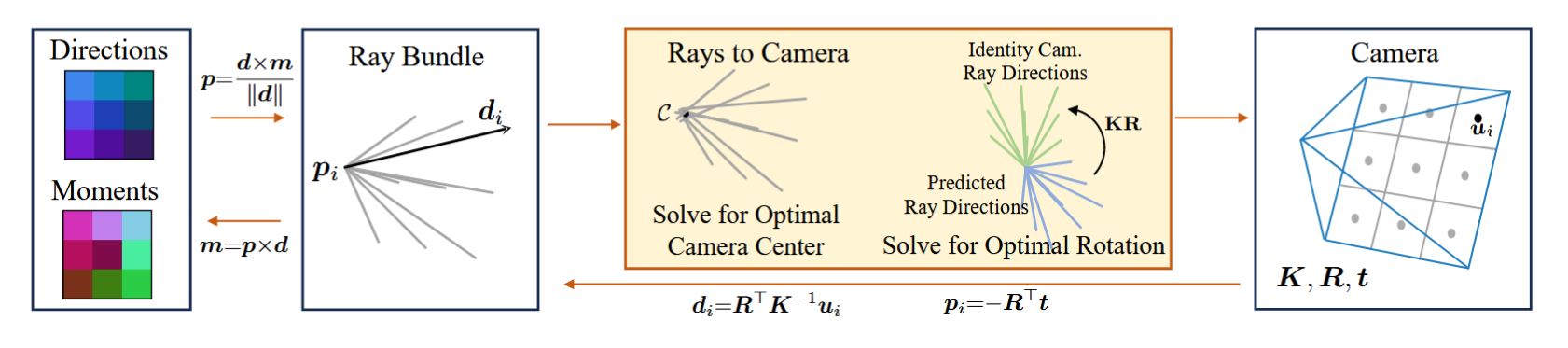}
    \caption{Conversion process for Pl\"{u}cker coordinates \cite{zhang2024raydiffusion}.}
    \label{fig:camera_rays}
\end{figure}

Mathematically, the Pl\"{u}cker representation encodes each ray $r$ as:
\begin{equation} \label{eq:pluc}
r = \langle d, m \rangle, \quad m = p \times d,
\end{equation}

where $d \in \mathbb{R}^3$ is the direction vector, $m \in \mathbb{R}^3$ is the moment vector, and $p$ represents a point on the ray. The parameters $d$ and $m$ ensure the ray remains agnostic to the choice of $p$. To compute the rays from a known camera, the directions and moments are derived as:
\begin{equation} \label{eq:pluccart}
d = R^\top K^{-1} u, \quad m = (-R^\top t) \times d,
\end{equation}

where $R$, $t$, and $K$ denote the rotation matrix, translation vector, and intrinsics matrix of the camera, respectively \cite{zhang2024raydiffusion}. Term $u$ represents the 2D pixel coordinates in the image plane. These coordinates are typically expressed in normalized device coordinates (NDC) \cite{everitt2001interactive}, scaled to fit within a specific range, such as $[-1, 1]$ or $[0, 1]$, depending on the application. Figure~\ref{fig:camera_rays} illustrates the conversion between the classical camera representation and the ray-based model.

Representing a camera using Pl\"{u}cker coordinates introduces complexity and over-parameterization by modeling it as a bundle of rays. While this enables flexibility for diverse camera models, it demands intensive computation and complicates calibration. Converting these rays back to traditional parameters also involves optimization, which can reduce precision in applications needing high geometric accuracy \cite{zhang2024raydiffusion}.

\subsubsection{TUM Trajectory (3D Motion of Camera Over Time)}
The TUM camera trajectory format \cite{sturm12iros} is a standardized way to represent the movement of a camera through 3D space over time, often used in computer vision and robotics research. It captures both the position and orientation of the camera at each timestamp, using a 7-element vector. This vector includes the $timestamp$ in seconds (or frames), followed by the camera's translation ($x$, $y$, $z$ coordinates) and its orientation represented as a quaternion ($qx$, $qy$, $qz$, $qw$). 

A quaternion is a mathematical term used to represent rotations in three-dimensional space, consisting of four components: one real part and three imaginary parts. It is typically written as \eqref{eq:quat}, where $w$ is the scalar component, and $x$, $y$, $z$ are the vector components. Quaternions are particularly useful because they offer several advantages over other rotation representations, such as Euler angles. They help avoid issues like gimbal lock (the loss of one degree of freedom in a multi-dimensional mechanism at certain alignments of the axes) and allow for smooth, continuous interpolation between orientations. In the case of the TUM camera trajectory format, quaternions efficiently capture the camera's orientation, providing a compact and stable way to describe rotations in 3D space without redundancy or ambiguity.

\begin{equation} \label{eq:quat}
q = w + xi + yj + zk
\end{equation}

This compact format allows for a precise description of the camera's trajectory, which is crucial for evaluating and comparing different motion estimation algorithms. Another key advantage is its utility in benchmarking and evaluating algorithms in areas like visual odometry \cite{aqel2016review}, SLAM \cite{zhang2021survey}, and related fields, as it provides reliable ground truth data for comparing predicted camera trajectories. It is often used alongside RGB-D datasets, such as the TUM RGB-D dataset \cite{sturm12iros}, for more comprehensive evaluation.

\section{Movement System} \label{camera_movement_sys}
Camera movement systems are essential in computer vision and graphics, defining how cameras are manipulated to capture scenes. The term "camera movement" refers to the types of motions that cameras can perform, enabling diverse views of a scene \cite{christie2009camera}. These parameters collectively determine the position and orientation of the camera in a 3D space. The specific type of camera movement directly impacts how trajectories are planned and optimized, as it influences both the setup and the design of the system. In this section, we explore the most critical types of camera movement systems, emphasizing their characteristics and the importance of understanding camera setups for effective design and implementation.

These systems, whether in virtual or real-world environments, are classified as fixed or non-fixed. Fixed systems, characterized by stationary positions, are ideal for applications like surveillance or UAV monitoring, offering stability and simplified trajectory planning. Non-fixed systems, common in virtual environments, allow free movement within a defined space, making them suitable for dynamic applications such as video games \cite{burelli2016game}. In these games, non-fixed cameras adapt based on the perspective: first-person cameras synchronize with the player’s position and orientation, while third-person cameras provide external views that can be free or constrained. Additionally, during non-interactive sequences, cameras focus on highlighting key narrative elements without player control.

In the following subsections, we explore two specialized types of camera movement systems: Pan-Tilt-Zoom (PTZ) cameras and Gimbal-Mounted cameras. The first subsection focuses on PTZ systems, which enable dynamic adjustments in horizontal (pan), vertical (tilt), and focal length (zoom) movements, making them highly effective for real-time applications such as surveillance and broadcasting. The second subsection examines gimbal-mounted cameras, which leverage gyroscopic feedback and motorized stabilization to maintain smooth and steady imaging, particularly in UAV applications. These specialized systems showcase unique capabilities that cater to specific scenarios requiring precise control and adaptability in camera movement.

\subsection{Pan-Tilt-Zoom Camera} \label{PTZ}
\begin{figure}[h]
    \centering
    \includegraphics[width=0.3\linewidth]{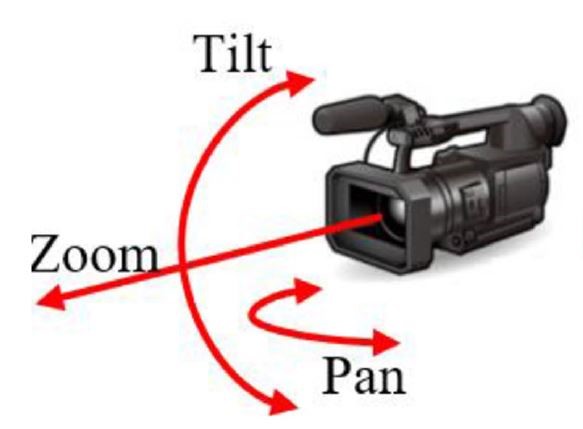}
    \caption{Camera motion of fixed PTZ Cameras \cite{bak2023camera}.}
    \label{fig:ptza}
\end{figure}

The pan-tilt-zoom (PTZ) camera movement system is useful particularly in scenarios where a fixed camera is employed. This system facilitates three primary motions: pan, tilt, and zoom, as depicted in Figure \ref{fig:ptza}. 

Pan refers to the horizontal rotation of the camera, enabling the tracking of objects moving laterally within a scene \cite{vineyard2008setting}. This movement that the subject remains within the frame during dynamic scenarios, such as sports events or live performances \cite{zhu2009trajectory, chen2015mimicking}. Similarly, tilt involves vertical rotation of the camera, which allows for capturing objects moving along the vertical axis or for emphasizing towering structures or high-angle perspectives .

Zoom, on the other hand, adjusts the focal length of the camera lens to magnify or reduce the size of the subject in the frame. This capability is often used to create emotional or dramatic tension by directing the viewer’s attention to specific elements of the scene \cite{brown2012cinematography, vineyard2008setting}. By integrating these motions, PTZ cameras offer a flexible approach to trajectory generation, as the system's operations are computationally lightweight and suitable for real-time adjustments in applications such as surveillance \cite{kumar2009real}, broadcasting \cite{chen2015mimicking}, and cinematography \cite{pattanayak2024automating}.

\begin{figure}[h]
    \centering
    \includegraphics[width=0.3\linewidth]{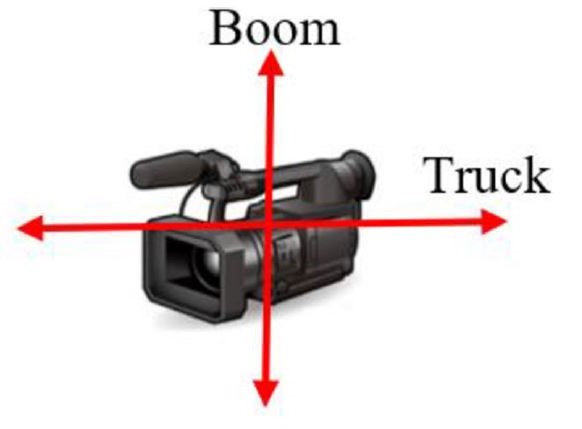}
    \caption{Camera motion of non-fixed PTZ Cameras \cite{bak2023camera}.}
    \label{fig:ptzb}
\end{figure}

Compared to non-fixed camera systems like boom or truck movements, as illustrated in Figure \ref{fig:ptzb}, PTZ cameras offer a simpler yet effective approach for generating diverse trajectories. Truck movements shift the field of view laterally, useful for dynamic tracking shots, while boom movements provide vertical adjustments for varied perspectives \cite{brown2012cinematography}. Although these non-fixed motions are valuable in cinematic contexts, the rotational and zoom capabilities of PTZ systems serve as a compact and versatile alternative for achieving complex camera trajectories without requiring physical relocation \cite{vineyard2008setting}.

\subsection{Gimbal Mounted Camera} \label{gimbal}
\begin{figure}[h]
    \centering
    \includegraphics[width=0.3\linewidth]{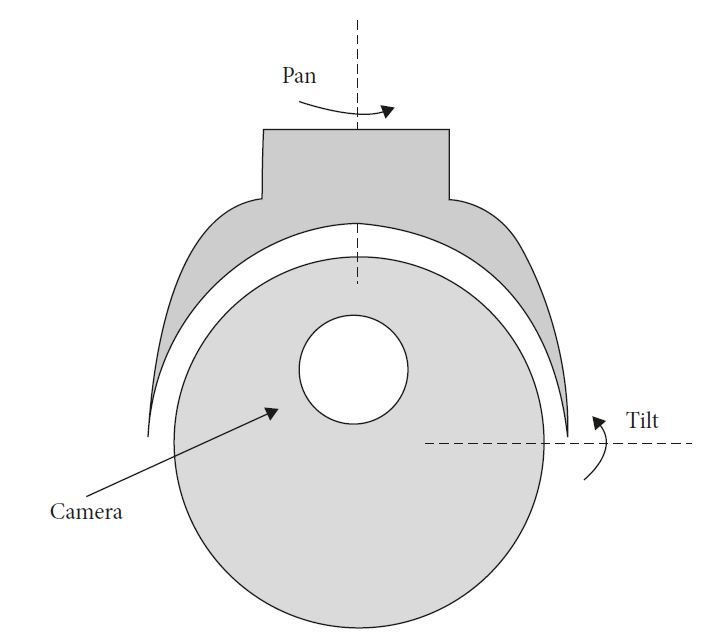}
    \caption{Overview of yaw-pitch gimbal \cite{cong2021stability}.}
    \label{fig:gimbalcam}
\end{figure}

Gimbal-mounted camera systems are widely used in unmanned aerial vehicles (UAVs) to stabilize and control camera movement during flight. These systems typically consist of a motorized structure that allows adjustments in two key directions: yaw (horizontal rotation) and pitch (vertical tilt), as shown in Figure \ref{fig:gimbalcam}. 

The camera is integrated within the gimbal, with its lens oriented outward, enabling precise control over its movement and stabilization. However, this design introduces challenges, such as an unbalanced mass due to the inclusion of the camera. This imbalance directly affects the pitch angle, making it a critical parameter to optimize for smooth operation and stability.

\begin{figure}[h]
    \centering
    \includegraphics[width=0.3\linewidth]{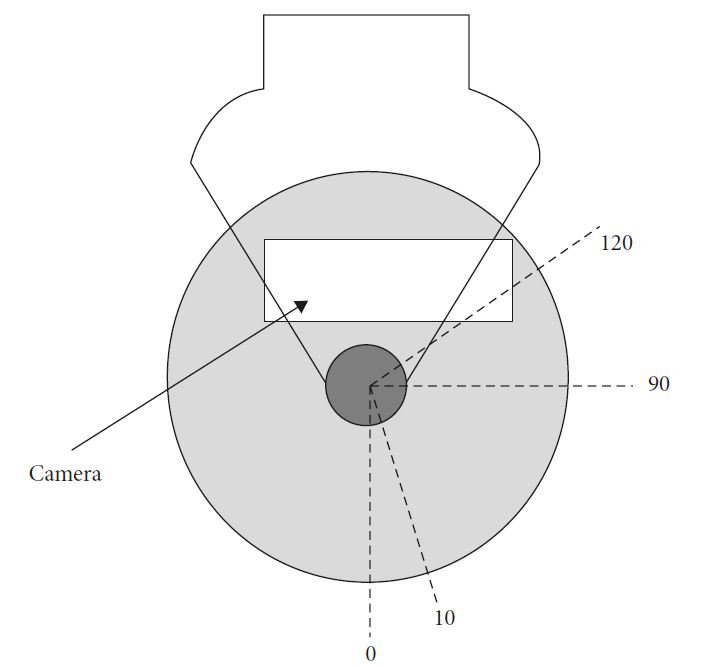}
    \caption{Pitch angle limit \cite{cong2021stability}.}
    \label{fig:gimbalcamlim}
\end{figure}

Gimbal systems integrate gyroscopes to measure movement speeds and interact with motor torque, creating a control loop that stabilizes camera movements and minimizes disturbances \cite{cong2021stability}. This motor and gimbal integration ensures smooth operation, but certain design limitations persist. For instance, the camera frame is obscured at pitch angles beyond 120 degrees, and images invert at negative pitch angles (less than 0 degrees), as shown in Figure \ref{fig:gimbalcamlim} \cite{cong2021stability}. These constraints demand precise calibration to maintain proper image orientation and smooth, blur-free camera motion, highlighting the need for responsive and accurate control systems.

Gimbal-mounted camera systems are particularly valued in UAVs for their ability to maintain image stability during rapid or irregular movements. The combination of precise gyroscopic feedback, motorized control, and careful pitch angle calibration ensures high-quality imaging in dynamic aerial environments, making these systems indispensable for UAV applications.

\section{Algorithm} \label{algorithms_section}
Algorithms are essential for generating precise and efficient camera trajectories across applications like cinematography, graphics, and robotics \cite{bonatti2020autonomous, gebhardt2021optimization}. By automating trajectory planning, they address challenges such as complex environments, computational efficiency, and real-time constraints \cite{burg2020real, burg2021real, nageli2017real}. Bridging artistic principles with technology, algorithms enhance storytelling, user immersion, and visual coherence. Advances in rule-based, optimization, and learning-based methods have expanded the capabilities of camera systems, enabling creative and adaptable trajectory generation \cite{wang2024dancecamanimator, wang2024dancecamera3d}.

This section categorizes the prominent algorithms into four groups. Rule-based methods rely on predefined cinematic principles and heuristics, offering reliability but limited flexibility. Optimization techniques formulate trajectory generation as a problem of maximizing shot quality while balancing constraints and objectives. Machine learning approaches leverage data-driven models to learn complex motion patterns, introducing adaptability and creativity. Finally, hybrid methods integrate multiple strategies, combining the strengths of rule-based, optimization, and learning techniques to achieve enhanced performance and versatility. The following subsections discuss each category in detail, highlighting their foundational principles, strengths, and limitations. 

\subsection{Rule-Based}
Rule-based methods for camera trajectory generation rely on established cinematography principles rather than optimization or learning-based techniques. These approaches utilize traditional cinematic rules, expert insights, and well-defined heuristics, such as camera placement and guidelines \cite{christie2009camera, chen2014autonomous}. These approaches offer a practical and computationally efficient solution. However, their rigidity poses a limitation, as they strictly adhere to predefined rules, making adaptation and creativity challenging. Modifications often require revising or replacing these rules. Despite their inflexibility, rule-based methods provFide reliability and efficiency, particularly in scenarios with limited computational resources. The following section discusses key contributions in this domain.

The first significant contribution to the application of cinematography principles for generating camera trajectories is presented in \cite{he1996virtual}, where the authors introduced the concept of the Virtual Cinematographer (VC), a system designed to generate real-time camera trajectories in virtual 3D environments. The VC incorporates cinematographic expertise using film idioms, implemented as a hierarchy of finite state machines, each suited to specific scene types. These idioms control shot selection and transition timing to effectively depict unfolding events. The paper details the filmmaking heuristics embedded in the system and demonstrates its application in a virtual "party" scenario. However, the system’s applicability is constrained to a specific scenarios, limiting its broader generalizability.

Tomlinson et al. \cite{tomlinson2000expressive} introduced a behavior-based autonomous cinematography system designed for interactive 3D environments. The system employs ethologically-inspired mechanisms, such as sensors, motivations, and hierarchical action-selection, to select optimal camera shots in real-time. It integrates seamlessly with virtual actors, enabling information exchange to create a cohesive and enriched environment. However, challenges include maintaining adaptability to unpredictable actor behaviors and ensuring user comfort through effective coordination with the user interface. While limitations exist, the work establishes foundational principles for interactive cinematography systems.

Mezouar and Chaumette (2003) propose a method for generating camera trajectories in image-based control systems through the use of smooth collineation paths connecting initial and desired viewpoints \cite{mezouar2003optimal}. The approach aims to reduce energy consumption and acceleration while ensuring robustness against modeling errors and noise. A key feature of this method is its ability to operate without prior camera calibration or a predefined scene model. Furthermore, the framework incorporates a potential field-based planning scheme to manage trajectory constraints, enabling effective tracking and adaptability in complex visual servoing tasks. However, the paper does not address potential limitations related to scalability or applicability in more intricate scenarios.

Christie et al. \cite{Christie2008camera} provide an review of camera control techniques aimed at enhancing viewer engagement in virtual environments. The paper addresses a range of methods, including viewpoint computation, motion planning, and editing, grounded in cinematographic principles to meet diverse application requirements. A key focus is on constraint-based and optimization-based approaches, offering detailed insights into camera placement and movement strategies. The study also explores occlusion management and the cognitive and aesthetic dimensions of camera expressiveness. However, reliance on geometric abstractions may limit the handling of complex 3D scenes, particularly in occlusion management and precise positioning.

A prototype system was introduced for real-time rendering and automatic camera control in augmented virtual environments based on sparse video inputs \cite{silva2011automatic}. The system combines multiple video streams with a 3D scene model to facilitate free-viewpoint visualization and automatic object tracking. Notable features include real-time foreground-background segmentation, view-dependent texture mapping, and camera color calibration. The approach is particularly suited for surveillance and event analysis applications. However, the paper does not address potential challenges related to scalability or the system's performance under varying environmental conditions, which may affect its generalizability.

\begin{figure}[h]
    \centering
        \includegraphics[width=0.5\linewidth]{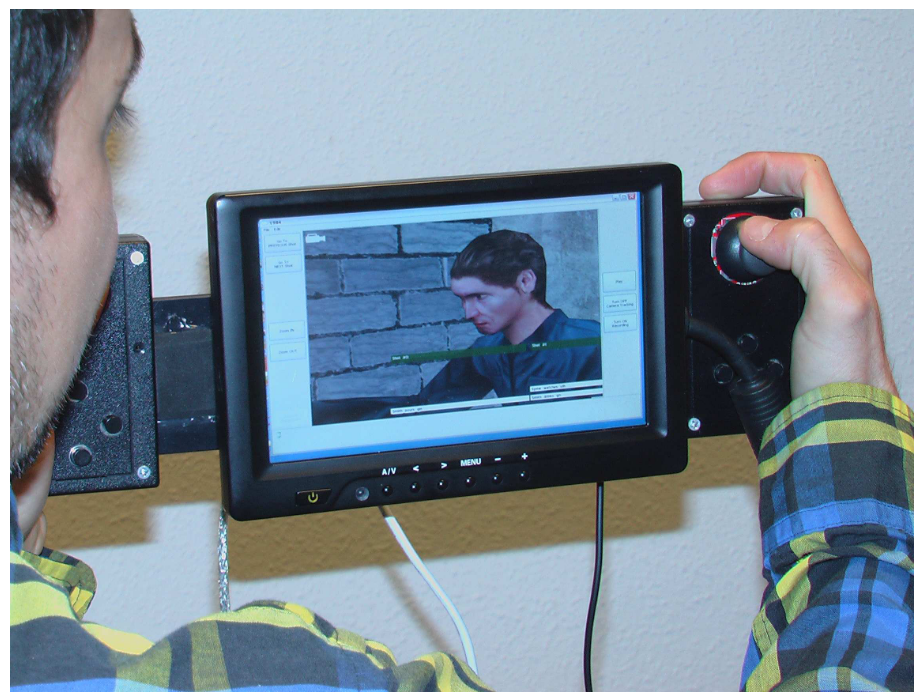}
    \caption{The hand-held virtual camera device with custom-built dual handgrip rig and button controls, a 7-inch LCD touch-screen \cite{lino2011director}.}
    \label{fig:director_lens}
\end{figure}

Lino et al. \cite{lino2011director} propose a system to support the filmmaking process through an interactive assistant that uses a motion-tracked hand-held device for virtual cinematography. This approach facilitates rapid exploration of cinematographic options and efficient production of computer-generated films. However, the reliance on pre-defined cinematic knowledge limits its adaptability to unexpected scenarios, potentially constraining creative judgment. While effective for guided filmmaking, the system may not always align with the user’s vision in novel or unconventional contexts. The hand-held virtual camera device is shown in Figure \ref{fig:director_lens}.

\blue{In a paper published in 2013, an approach was introduced to address the challenges of autonomous camera control in dynamic 3D environments \cite{galvane2013steering}. The study employs Reynolds’ steering behaviors \cite{reynolds1999steering} to control multiple autonomous cameras in crowd simulations. The proposed system models cameras as intelligent agents that dynamically transition between scouting and tracking modes, optimizing their positioning to maximize event visibility while minimizing occlusions. By leveraging steering forces and torques, the framework ensures adaptive, collision-free camera behaviors, producing diverse and informative shots.}

Quentin Galvane et al. \cite{galvane2014narrative} propose a system for automated cinematic replays in dialogue-based 3D games, focusing on narrative-driven camera control. The method assesses characters' narrative importance to inform camera framing, diverging from traditional action- or idiom-based approaches. It includes modules for assigning camera specifications based on narrative weight and for animating cameras smoothly across scenes. By utilizing toric \cite{Lino2015} and spherical models \cite{Lino2012, galvane2015camera}, the system produces dynamic and visually coherent cinematic shots.

\begin{figure}[h]
    \centering
        \includegraphics[width=0.5\linewidth]{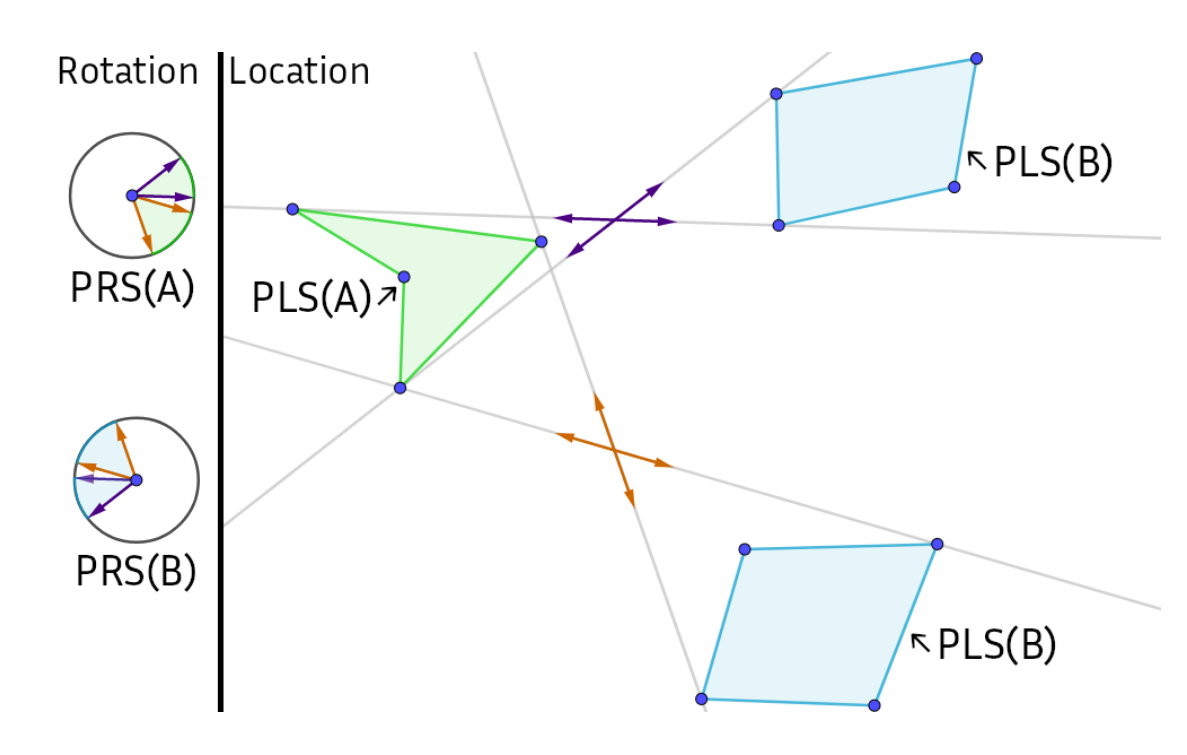}
    \caption{PLRS for two entities A (in green) and B (in blue) \cite{louarn2018automated}.}
    \label{fig:plrs}
\end{figure}

The often-overlooked challenge of object placement, or staging, in virtual cinematography was tackled through the introduction of a staging language, presented as an extension of Prose Storyboard Language (PSL) \cite{ronfard2015prose, louarn2018automated}. This language coordinates the simultaneous positioning of characters and cameras through geometric pruning and sampling operators, combined with fixed-point computation, to generate multiple staging solutions. The pruning operators are applied to the PLRS, shown in Figure \ref{fig:plrs}.

Building on this work, the staging language was further extended to incorporate temporal relationships, facilitating the simultaneous manipulation of cameras, lights, objects, and actors \cite{louarn2020interactive}. The iterative pruning operators and graph-based problem decomposition enhance cinematic precision and adaptability, with an interactive system allowing fine-tuning and exploration. However, challenges remain, including scalability in dynamic environments, graph regeneration disrupting solution continuity, and diagnosing conflicting constraints.

Jovane et al. \cite{jovane2020topology} address camera placement and movement in 3D virtual environments using a topology-driven approach. This method utilizes navigation mesh analysis to create abstract skeletal representations of the environment, which are then used to generate camera positions and trajectories organized in graph structures with visibility data. The system dynamically selects optimal cameras and paths based on artistic guidelines, making it suitable for real-time applications. While the approach allows for diverse and adaptive camera behaviors in dynamic scenarios, its lack of event-specific contextual knowledge may limit narrative alignment. Additionally, further development is needed to incorporate high-level controls and stylistic diversity for more expressive cinematographic applications.

Yoo et al. \cite{yoo2021virtual} propose an automated approach to creating virtual camera layouts in 3D animation by replicating the cinematic attributes of a reference video. The method extracts key cinematic elements, such as framing, camera movements, and subject features, to generate adaptable layouts for both human-like and exaggerated characters. User evaluations suggest the generated layouts are similar to those created by professionals, while reducing layout creation time, especially for novices. Although the system is effective for initial layout development, its reliance on extracted features may limit adaptability in dynamic or unconventional scenarios.

Rule-based methods for camera trajectory generation offer a reliable framework grounded in established cinematographic principles, ensuring practical application and computational efficiency. Their strengths lie in leveraging predefined rules to produce consistent results, particularly in real-time and resource-constrained scenarios. However, the inherent rigidity of these methods limits adaptability and creative flexibility, requiring manual updates to accommodate novel contexts or evolving cinematic needs. Innovative systems like the Virtual Cinematographer and topology-driven approaches enhance real-time applicability, yet challenges persist in scaling to dynamic or complex environments, such as occlusion and dynamic environment \cite{he1996virtual, jovane2020topology}. Future advancements should prioritize integrating adaptive and hybrid techniques to balance reliability, creativity, and user-driven flexibility.

Rule-based methods rely on well-established cinematographic principles and predefined heuristics to generate camera trajectories. These methods offer computational efficiency and reliability, particularly in constrained scenarios where flexibility is less critical. However, their rigidity limits adaptability to novel contexts, requiring manual updates to accommodate changing requirements. Table \ref{tab:RBCTG} highlights notable contributions in this area. 

\newcommand{\rowheightmethods}{0.5ex} 

\begin{table*}%
\caption{Overview of Rule-Based Methods for Camera Trajectory Generation}
\label{tab:RBCTG}
\begin{center}
\begin{tabular}{lccc}
  \toprule
  Method & \makecell[c]{Real World} & Virtual & \makecell[c]{Camera Movement} \\ \midrule

  \rule{0pt}{\rowheightmethods} \cite{he1996virtual} & - & Animation & Non-Fixed \\

  \rule{0pt}{\rowheightmethods} \cite{tomlinson2000expressive} & - & Animation & Non-Fixed \\

  \rule{0pt}{\rowheightmethods} \cite{mezouar2003optimal} & Human-Based & - & Non-Fixed \\

  \rule{0pt}{\rowheightmethods} \cite{silva2011automatic} & Human-Based & - & Non-Fixed \\

  \rule{0pt}{\rowheightmethods} \cite{lino2011director} & - & Animation/Games & Non-Fixed \\
  
    \rule{0pt}{\rowheightmethods} \cite{galvane2013steering} & - & Animation/Games & Non-Fixed \\

  \rule{0pt}{\rowheightmethods} \cite{chen2014autonomous} & - & - & - \\

  \rule{0pt}{\rowheightmethods} \cite{galvane2014narrative} & - & Games & Non-Fixed \\

  \rule{0pt}{\rowheightmethods} \cite{ronfard2015prose} & Human-Based & - & - \\

  \rule{0pt}{\rowheightmethods} \cite{louarn2018automated} & - & Animation/Games & Non-Fixed \\

  \rule{0pt}{\rowheightmethods} \cite{louarn2020interactive} & - & Animation/Games & Non-Fixed \\

  \rule{0pt}{\rowheightmethods} \cite{jovane2020topology} & Human-Based & Animation/Games & Non-Fixed \\

  \rule{0pt}{\rowheightmethods} \cite{yoo2021virtual} & - & Animation & Non-Fixed \\

  \bottomrule
\end{tabular}
\end{center}
\bigskip\centering
\footnotesize\emph{Note:} All the entries are entered based on evidence or our evaluation.
\end{table*}%

\subsection{Optimization}
Optimization techniques for camera trajectory generation often express shot properties as objectives to maximize or to minimize, with metrics evaluating the quality of shots based on the scene's graphical model and user-defined criteria \cite{bonatti2020autonomous}. Classical methods include deterministic approaches, such as gradient-based \cite{bengio2000gradient} and Gauss-Seidel techniques \cite{tewari2021overview}, alongside non-deterministic strategies like genetic algorithms \cite{wright1991genetic}, Monte Carlo methods \cite{kroese2012monte}, and stochastic local search \cite{hoos2018stochastic}. While pure optimization techniques can produce solutions where properties are partially satisfied, they risk unbalanced outcomes, with some objectives dominating others \cite{mca28050100}. Conversely, purely constraint-based methods \cite{Meseguer2003} can compute complete sets of solutions but are computationally intensive and struggle with over-constrained problems. A practical alternative lies in constrained optimization, combining enforceable constraints and optimizable properties to balance feasibility and quality \cite{galvane2015continuity}. Hybrid approaches that integrate constraint-based methods with optimization offer effective solutions, often leveraging geometric operators to narrow the search space before applying optimization techniques. In this section, we provide an overview of the various methods proposed in the field of camera trajectory generation, highlighting their underlying principles, strengths, and limitations. 

\subsubsection{7-DOF Optimization Problems}
The Optimization of camera trajectories can be formulated in a 7-DOF search space. The objective is to determine a camera configuration $q \in Q$, where $Q$ denotes the space of all possible configurations, that maximizes a fitness function \cite{Christie2008camera}. This can be mathematically expressed in Equation \ref{eq:fitness_optimization}.

\begin{equation}
\label{eq:fitness_optimization}
\text{maximize } F(f_1(q), f_2(q), \ldots, f_n(q)) \quad \text{s.t. } q \in Q,
\end{equation}

where each function $f_i: \mathbb{R}^7 \to \mathbb{R}$ evaluates the fitness of a specific property of the configuration, and $F: \mathbb{R}^n \to \mathbb{R}$ combines these fitness values into a single scalar output. A commonly used formulation for $F$ is a weighted sum \cite{Marler2010}, defined in Equation \ref{eq:weighted_sum}.

\begin{equation}
\label{eq:weighted_sum}
F(f_1(x), f_2(x), \ldots, f_n(x)) = \sum_{i=1}^n w_i f_i(x),
\end{equation}

where $w_i$ represents the weight associated with the $i$th property, allowing user preferences to influence the optimization process.

Exploring the continuous 7-DOF search space can be simplified through discretization \cite{latombe2012robot}, transforming it into a manageable grid. The CONSTRAINTCAM framework \cite{bares2000model} was extended with a global optimization strategy that exhaustively evaluates configurations based on an aggregated fitness value, as described in Equation \ref{eq:weighted_sum}. A typical discretization divides the search space into a $50 \times 50 \times 50$ grid for positions, $15^\circ$ angular increments for orientation, and 10 levels for the field of view. To enhance efficiency, feasible regions are identified by intersecting individual property regions, and the grid resolution is iteratively reduced. The process terminates when a predefined quality threshold is met or the minimal resolution is reached, ensuring efficient exploration while adhering to the constraints in Equation \ref{eq:fitness_optimization}.

An incremental solving approach for automating camera control in real-time target-tracking applications was introduced to manage shot properties such as relative elevation, size, visibility, and screen position while ensuring frame coherence to avoid abrupt movements \cite{halper2001camera}. This system employs an algebraic incremental solver to adjust camera configurations by incrementally satisfying screen constraints and selectively relaxing subsets when necessary. Look-ahead techniques are used to refine parameters based on anticipated object motion \cite{halper2001camera}. Similarly, Bourne and Sattar~\cite{bourne2005applying} proposed a local search optimization method to preserve object-relative properties like height, distance, orientation and ensure smooth camera paths.

The problem of computing optimal viewpoints in 3D environments is common in applications across computer graphics and robotics \cite{scott2003view}. For instance, image-based modeling requires selecting a minimal set of cameras to cover all visible surfaces for texture mapping \cite{debevec2023modeling}. Early work by Kamada and Kawai~\cite{kamada1988views} inspired many approaches by maximizing the projected area to surface area ratio. Solutions often use classical solvers, such as simulated annealing~\cite{stuerzlinger1999annealing}, or heuristic methods that populate environments with cameras and apply coverage metrics to evaluate solutions~\cite{fleishman2000automatic}. A coverage metric evaluates how effectively selected viewpoints or cameras capture the required surfaces or areas of a 3D environment, considering visibility, resolution, and overlap criteria. Viewpoint entropy \cite{vazquez2003entropy}, maximizes the information captured in a minimal set of views. Other research explores cognitive aspects like scene understanding and attention \cite{viola2006geometry}, who try to augment geometry with object importance to compute characteristic views using visibility and importance metrics. For scene exploration, heuristic optimization methods compute automatic camera paths by attracting the camera to unexplored areas based on physical models~\cite{stoev2006paths}. Initial configurations in these methods are guided by viewpoint quality estimations using total surface curvature and projected area.

While optimization techniques in this section provide precise trajectories and a more realistic camera model, many of these automated solutions are considered impractical. The algorithms operate in a seven-dimensional space, which is virtually infinite, leading to high computational complexity \cite{Lino2015}. Additionally, the search process demands substantial computational power, making it unsuitable for real-time systems or hardware with strict resource limitations. As a result, these methods often fail to meet the necessary delay constraints for safe, real-time use \cite{ ranon2014improving}. Despite these challenges, 7-DOF algorithms offer valuable benefits in terms of camera abstraction and interpretability, which sets them apart from alternative methods that employ different approaches \cite{Taketomi2017}.

\subsubsection{Low Dimension Optimization Problems (LDO)}
The optimization problem addressed in \cite{Lino2012} aims to improve the computational efficiency of virtual camera control, specifically for satisfying exact on-screen positioning of multiple subjects. Traditional methods, such as those relying on high-dimensional 7-DOF search spaces, encounter issues due to the computational cost of exploring large regions of the solution space, which limits practical applications. The proposed approach \cite{Lino2012} reduces this complexity by representing the solution space as a 2D manifold for two subjects and extending it algebraically to three or more subjects. This manifold is parameterized by meaningful angles, simplifying the optimization process while maintaining accuracy. 

The primary issue arising from traditional methods relying on high-dimensional searches is addressed through an optimization approach leveraging the Toric space \cite{Lino2015}. This technique reduces the search space from 7-DOF to 4-DOF. By employing an interval-based pruning algorithm (as shown in \ref{eq:toricequ}), the method incrementally narrows the solution space through constraints on angles ($\alpha$, $\theta$, and $\phi$) and field of view, ensuring that only regions meeting all necessary properties are retained.

\begin{equation} \label{eq:toricequ}
\min_{\alpha, \theta, \phi} \sum_{i} w_i \cdot \text{Error}_i(\alpha, \theta, \phi),
\end{equation}

where $\text{Error}_i$ quantifies the deviation of a visual property from its desired value, and $w_i$ is the weight assigned to that property. This cost function balances competing constraints to find optimal camera positions. While the approach is computationally efficient, it may struggle in highly over-constrained scenarios where no feasible solution exists \cite{Lino2015}.

The optimization approach in \cite{Galvane2015} addresses the challenge of generating smooth and realistic camera motions for dynamic scenes while satisfying aesthetic and physical constraints. It begins by interpolating a raw camera trajectory based on user-defined framing properties, which is then smoothed using a cubic Bézier curve \cite{arijon1976grammar}. A two-step optimization refines this trajectory, minimizing positional errors and ensuring smooth transitions in velocity, controlled acceleration, and accurate orientation adjustments.

The work in \cite{ren2023automatic} automates camera control in dynamic settings by integrating PTZ mechanics \ref{PTZ} with DNN-based visual sensing. Traditional systems lack real-time adaptability, often relying on predefined paths. The process begins with visual detection using DNNs \cite{samek2021explaining}, followed by target tracking and estimation via Kalman filters \cite{khodarahmi2023review}. Trajectories are dynamically planned with PID control \cite{borase2021review}, adjusting pan, tilt, and zoom to maintain aesthetic composition within physical constraints, such as angular velocity and acceleration limits.

Research in this area has primarily focused on altering the camera's representation or fixing some of the dimensions to reduce the overall search space. The use of Toric space has been particularly dominant due to its efficient mathematical representation and its ability to be transformed into Cartesian coordinates. However, several challenges persist in this domain. One key issue is that many algorithms achieve lower-dimensional solutions by either simplifying certain parameters or fixing them, which reduces the search space but often leads to compromises in flexibility \cite{burg2020real}. Additionally, some methods impose constraints to target specific problems or a fixed number of objectives, limiting their general applicability \cite{burg2020real}.

\subsubsection{Drone Trajectory Optimization (DTO)}
Creating camera trajectories for drones involves two distinct tasks with unique requirements. The first is object tracking, which ensures the camera remains focused on the target at all times without losing sight of it. The second is cinematography, which emphasizes aerial filming to achieve visually appealing shots \cite{bonatti2019autonomous}. A key distinction in drone-based filming is that the camera and drone are most often coupled, meaning that optimizing the drone's trajectory inherently optimizes the camera's path or the trajectory of the camera are often considered the trajectory of the drone. Optimization problems are widely used in drone applications due to the need for fast, real-time responses. Machine learning methods are less prevalent in this domain, as most drones lack the computational hardware required to run complex models efficiently, and such methods often introduce significant latency, making them unsuitable for time-sensitive tasks. In this section, we explore optimization techniques tailored to aerial vehicles, addressing these challenges effectively.

\blue{In a paper introduced in 2016 \cite{gebhardt2016airways}, a computational framework has been developed to plan quadrotor trajectories by integrating high-level user objectives with physical feasibility constraints. Optimization-based methods are employed to generate flight paths that adhere to user-defined goals, such as smooth aerial videography or complex maneuvers, without requiring expertise in low-level control systems. A 3D design interface allows intuitive specification and iterative refinement of trajectories. Constraints from cinematography, physical dynamics, and collision avoidance are incorporated to ensure practical applicability across use cases, including drone racing and robotic light-painting.}

The optimization problem in \cite{roberts2016generating} addresses the challenge of generating dynamically feasible trajectories for quadrotor cameras, which must satisfy velocity and control force limits while preserving the visual layout of user-specified paths. This is critical because infeasible trajectories can result in unsafe quadrotor operation or deviation from intended paths. The proposed solution optimizes the progress curve \( s(t) \), re-timing the trajectory to ensure physical feasibility with minimal deviation from the user’s input. The algorithm discretizes the camera path, enforcing constraints on velocity, acceleration, and control forces through a non-convex optimization frameworkas shown in \ref{eq:robertgen}.

\begin{equation} \label{eq:robertgen}
\begin{aligned} 
\min_{S,V} & \sum_{i} (\dot{s}_i - \dot{s}^{\text{ref}}_i)^2 \\
\text{subject to} & \quad s_{i+1} = s_i + (M s_i + N v_i) \frac{\Delta s_i}{\dot{s}_i}, \\
& \quad v_{\min} \leq v_i \leq v_{\max}, \quad \dot{s}_i > 0, \\
& \quad u_{\min} \leq U(s_i) \leq u_{\max}, \\ \quad \dot{q}_{\min} \leq \dot{Q}(s_i) \leq \dot{q}_{\max},
\end{aligned}
\end{equation}

where \( \dot{s}^{\text{ref}}_i \) represents the desired progress curve derivatives, let $S$ be the concatenated vector of all $s_i$ values along the path, let $V$ be the concatenated vector of all $v_i$ values along the path., and \( U(s_i) \) and \( \dot{Q}(s_i) \) represent control forces and velocity constraints, respectively.

The challenge of balancing dynamic feasibility in drone motion, such as adhering to velocity and acceleration limits, with cinematographic constraints like framing targets and ensuring smooth transitions, is addressed in \cite{nageli2017real}. The proposed solution involves an optimization process that minimizes a composite cost function, representing deviations from desired shot parameters while respecting both physical and cinematic constraints, such as framing, collision avoidance, visibility, and pose alignment. This unified framework effectively integrates aesthetic and physical considerations, allowing drones to execute precise and visually appealing movements.

\begin{equation} \label{eqnegalreal}
\min_{\mathbf{x}, \mathbf{u}, \mathbf{s}} w_N^\top c(\mathbf{x}_N, \mathbf{u}_N) + \sum_{k=0}^{N-1} w^\top c(\mathbf{x}_k, \mathbf{u}_k) + \lambda \| \mathbf{s}_k \|_\infty,
\end{equation}

subject to:
\begin{align*}
    \mathbf{x}_0 &= \mathbf{x}_0^\text{init}, & \text{(Initial State)} \\
    \mathbf{x}_{k+1} &= f(\mathbf{x}_k, \mathbf{u}_k), & \text{(Dynamics)} \\
    r_{ct}^\top \Omega r_{ct} &> 1 - s_k, & \text{(Collision Avoidance)} \\
    r_{ct} &= g(\mathbf{x}_k), & \text{(Geometric Relationship)} \\
    \mathbf{x}_k &\in \mathcal{X}, & \text{(State Constraints)} \\
    \mathbf{u}_k &\in \mathcal{U}, & \text{(Input Constraints)} \\
    \mathbf{s}_k &\geq 0, & \text{(Slack Constraints)}
\end{align*}

The cost function \( c(\mathbf{x}_k, \mathbf{u}_k) \) is defined as:
\begin{equation} \label{eqnegalrealcost}
c(\mathbf{x}_k, \mathbf{u}_k) = \begin{bmatrix}
    c_\text{image}, c_\text{size}, c_\text{angle}, c_\text{coll}, c_\text{vis}, c_\text{pose}
\end{bmatrix}^\top_{(\mathbf{x}_k, \mathbf{u}_k)},
\end{equation}


The cost function minimizes the terminal cost is \( w_N^\top c(\mathbf{x}_N, \mathbf{u}_N) \), cumulative stage costs \( \sum_{k=0}^{N-1} w^\top c(\mathbf{x}_k, \mathbf{u}_k) \), and a penalty term \( \lambda \| \mathbf{s}_k \|_\infty \) to handle constraint relaxation through slack variables. The system starts at an initial state \( \mathbf{x}_0^\text{init} \) and evolves via dynamics \( \mathbf{x}_{k+1} = f(\mathbf{x}_k, \mathbf{u}_k) \). Collision avoidance is enforced by requiring \( r_{ct}^\top \Omega r_{ct} > 1 - s_k \), where \( r_{ct} = g(\mathbf{x}_k) \) defines geometric relationships, with slack \( \mathbf{s}_k \) ensuring feasibility. States \( \mathbf{x}_k \) and controls \( \mathbf{u}_k \) must adhere to feasible sets \( \mathcal{X} \) and \( \mathcal{U} \), respectively, while slack variables \( \mathbf{s}_k \) are constrained to be non-negative penalty to balance accuracy, smoothness, and constraint relaxation.

The work in \cite{nageli2017multi} extends the optimization framework from \cite{nageli2017real} to address challenges in cluttered environments. Using a non-linear Model Predictive Contouring Control (MPCC) \cite{lam2010model}, it integrates framing objectives, path accuracy, and collision avoidance into the cost function, enabling real-time trajectory re-planning. The method accounts for dynamic constraints and uses slack variables to handle infeasibilities, ensuring smooth, collision-free motion suitable for high-quality cinematography, even with multiple drones.


\blue{A study published in 2018 \cite{gebhardt2018optimizing} introduced an optimization-based approach for generating smooth and visually appealing quadrotor camera trajectories. The problem was formulated as an infinite-horizon optimization framework, where a weighted cost function \( J_i \) was minimized to balance positional accuracy, motion smoothness, and timing control. This cost function incorporates terms for positional reference tracking, orientation alignment, jerk minimization, timing progress, and control regularization, with adjustable scalar weight parameters to achieve a trade-off between these objectives. The optimization problem is solved under constraints, including system dynamics, bounds on states and control inputs, and progress variables. This formulation ensures that the generated trajectories adhere to user-defined spatial and temporal requirements while maintaining aesthetic smoothness. The formula for this optimization method is detailed in Equation \ref{optimization11}.}

\begin{align} \label{optimization11}
\min_{x, u, \Theta, v} \, \sum_{i=0}^{N} \, w_p c^p(\theta_i, \mathbf{r}_i) + w_\psi c^\psi(\theta_i, \psi_q, i, \psi_g, i) + w_\phi c^\phi(\theta_i, \phi_q, i) + \nonumber \\ w_j c^j(\dddot{\mathbf{r}}, \dddot{\psi_q}, \dddot{\phi_q}, i) + w_\text{end} c^\text{end}(T) + w_\text{len} c^\text{len}(N, \Delta t) + w_v \| \mathbf{v} \|^2,
\end{align}

\text{subject to}

\begin{align*}
    \mathbf{x}_0 &= k_0, & \text{(initial state)} \\
    \mathbf{\Theta}_0 &= 0, & \text{(initial progress)} \\
    \mathbf\Theta_N &= L, & \text{(terminal progress)} \\
    \mathbf{x}_{i+1} &= A x_i + B u_i + g, & \text{(dynamical model)} \\
    \mathbf{\Theta}_{i+1} &= C \Theta_i + D v_i, & \text{(progress model)} \\
    \mathbf{x}_\text{min} &\leq x_i \leq x_\text{max}, & \text{(state bounds)} \\
    \mathbf{u}_\text{min} &\leq u_i \leq u_\text{max}, & \text{(input limits)} \\
    \mathbf 0 &\leq \Theta_i \leq \Theta_\text{max}, & \text{(progress bounds)} \\
    \mathbf 0 &\leq v_i \leq v_\text{max}, & \text{(progress input limits)}
\end{align*}

where the scalar weight parameters $w_p, w_\psi, w_\phi, w_j, w_\text{end}, w_\text{len}, w_v > 0$ are adjusted for a good trade-off between positional fit and smoothness.

The optimization problem in \cite{bonatti2020autonomous} focuses on generating smooth, and visually appealing trajectories for drones filming dynamic actors, addressing issues such as obstacle avoidance, occlusion prevention, and adherence to artistic cinematography principles. They argued that traditional methods either neglect critical artistic objectives or fail in real-world scenarios with noisy localization and dynamic obstacles. This approach decouples the drone and camera motions, leveraging a gimbal \ref{gimbal} for fine adjustments. The proposed solution formulates the trajectory optimization as minimizing a composite cost function \( J(\xi_q) \) defined as Equation \ref{eq:J}.

\begin{equation}
\label{eq:J}
\begin{aligned}
J(\xi_q(t)) &= J_\text{smooth}(\xi_q(t)) 
+ \lambda_1 J_\text{obs}(\xi_q(t)) \\
&\quad + \lambda_2 J_\text{occ}(\xi_q(t), \xi_a(t)) 
+ \lambda_3 J_\text{shot}(\xi_q(t), \xi_a(t)), \\ 
\xi_q^*(t) &= \arg\min_{\xi_q(t) \in \Xi} J(\xi_q(t)), 
\quad \forall t \in [0, t_f].
\end{aligned}
\end{equation}

where \( J_\text{smooth} \) ensures trajectory smoothness, \( J_\text{obs} \) penalizes proximity to obstacles, \( J_\text{occ} \) reduces occlusion between the camera and the actor \( \xi_a \), and \( J_\text{shot} \) enforces adherence to artistic shot guidelines. $\xi_q(t)$ are the trajectory of the quadrotor (drone) represents its position in 3D space over time, $\xi_a(t)$ in the otherhand are trajectory of the actor describes their position over time. subject to boundary constraints and the drone's dynamic feasibility. The optimization process utilizes a covariant gradient descent \cite{zucker2013chomp} approach to iteratively minimize \( J(\xi_q) \), ensuring efficient convergence while accounting for noise in actor predictions. 


The study in \cite{rousseau2018quadcopter} tackles the challenge of generating smooth quadcopter trajectories for cinematic applications by minimizing jerk to enhance video quality. A bilevel optimization approach is employed: the first step adjusts velocity references within vertical and lateral limits, and the second step computes a minimum-jerk trajectory via quadratic programming. To manage complex flight plans, a receding waypoint horizon is used, iteratively computing trajectories over shorter segments to ensure smooth transitions and constraint adherence.


A method for dynamically sampling 3D environments with a visibility-aware roadmap is presented in \cite{galvane2018directing}, addressing the challenge of adapting to moving obstacles. The approach uses a composite distance metric combining cinematographic properties, such as target distance and angles, with spatial constraints. Path planning operates in a 4D parameter space, integrating the DTS \ref{DTS} for visual properties and altitude for spatial consistency, and employs the A* algorithm \cite{oskam2009visibility}. Trajectories are refined to \( C^4 \)-continuity to ensure smoothness and minimize abrupt changes in drone dynamics. \( C^4 \)-continuity refers to a mathematical property of a trajectory where the path and its first four derivatives (position, velocity, acceleration, jerk, and snap) are continuous.


An algorithm for real-time chasing a moving target in dense environments is presented in \cite{jeon2019online}. The approach ensures safety, visibility, and adherence to physical constraints by coupling the drone and gimbal camera trajectories, prioritizing target visibility. It refines a preplanned sequence of safe waypoints and corridors into a continuous trajectory using a convex optimization framework. Represented as piecewise polynomials, the trajectory minimizes a cost function, as detailed in Equation \ref{eq:online}.

\begin{equation}
\label{eq:online}
\min_{p_n} \sum_{n=1}^{N} \left( \int_{t_{n-1}}^{t_n} \|\dddot{\mathbf{x}}_c(\tau)\|^2 d\tau + \lambda \|\mathbf{x}_c(t) - \mathbf{x}_n\|^2 \right),
\end{equation}

where \( p_n \) represents the optimized waypoints or control points of the MAV's trajectory to ensure smoothness, safety, and visibility during motion planning, \( \mathbf{x}_c(t) \) represents the drone's position at time \( t_n \), \( \mathbf{x}_n \) is the \( n \)-th waypoint, and \( \dddot{\mathbf{x}}_c(\tau) \) is the jerk (third derivative of position). The cost function consists of two terms: the integral of squared jerk to ensure smooth motion, and a penalty term \( \lambda \|\mathbf{x}_c(t_n) - \mathbf{x}_n\|^2 \) to minimize deviations from the preplanned waypoints. The optimization in \cite{jeon2019online} incorporates constraints on initial conditions, trajectory continuity up to the second derivative, and adherence to safety corridors, formulating the problem as a quadratic programming task solved efficiently with interior-point methods \cite{gondzio2012interior}.


In the context of autonomous cinematography, Sabetghadam et al. \cite{sabetghadam2019optimal} solved the problem as a nonlinear optimization task, minimizing a cost function that combines control effort, camera smoothness, and terminal tracking objectives, as formulated in \eqref{eq:sabet}.

\begin{equation} \label{eq:sabet}
\min_{x_0, \dots, x_N, u_0, \dots, u_N} \sum_{k=0}^{N} \left( w_1 \| u_k \|^2 + w_2 J_\theta + w_3 J_\psi \right) + w_4 J_N,
\end{equation}

where \( J_\theta\) and \( J_\psi\) penalize angular camera movements, \( J_N \) enforces the final state’s proximity to the desired position and velocity, and \( u_k \) represents control inputs. The optimization is subject to constraints, such as, Enforces system kinematics \( x_{k+1} = f(x_k, u_k) \) to maintain trajectory feasibility, Limits \( v_Q, u_k \) within drone specifications, Keeps the drone at least \( r_\text{col} \) distance away from obstacles and, Maintains gimbal angles within mechanical limits. The optimization is solved iteratively in a receding horizon framework which is an approach to solving optimization problems over a time horizon that dynamically adapts to changes in the system. In this method, the system plans trajectories over a fixed prediction horizon, executes the initial part of the plan, and then re-optimizes as new information about the system's state and environment becomes available. 

The framework in \cite{bonatti2019towards} addresses the limitations of relying on predefined maps or precise localization by integrating actor localization, real-time LiDAR mapping, and trajectory planning. A composite cost function guides the trajectory planner, optimizing for smoothness to ensure stability and video quality, shot quality to adhering to cinematic guidelines like angle and distance, safety to avoiding collisions, and occlusion to minimizing visual obstructions using covariant gradient descent \cite{zucker2013chomp}.


Building on this, the approach in \cite{bonatti2019autonomous} redefines artistic shot selection as a sequential decision-making problem using deep reinforcement learning (RL). By modeling it as a Contextual Markov Decision Process (C-MDP) \cite{krishnamurthy2016pac}, the system maps scene context to optimal shot parameters in real time. The RL algorithm optimizes a reward function evaluating artistic quality metrics like smoothness, visibility, and obstacle avoidance, enabling adaptive and aesthetically refined drone behavior for high-quality cinematography.

The work in \cite{katoch2019edge} optimizes camera trajectories to minimize motion blur and preserve edge features critical for enhancing OCR accuracy \cite{mittal2020text}. It employs fourth-order polynomial trajectories that balance kinematic constraints with edge preservation, ensuring smooth motion with controlled velocity and acceleration. These trajectories maximize time at critical positions to enhance edge sharpness, and a tunable parameter allows fine-tuning between motion smoothness and edge clarity, improving real-time OCR performance.

In the realm of object tracking, \cite{jeon2020detection} stats that the primary focus must be on improving the detectability of a target during a drone cinematographer's chasing motion. The proposed optimization actively adjusts the drone's motion to ensure the target is distinguishable in the drone's view. The optimization process involves two main steps. First, a detectability-aware discrete path is generated by solving a directed acyclic graph (DAG) \cite{digitale2022tutorial} problem. The graph nodes represent candidate viewpoints, and edges are evaluated for both distance traveled and a detectability metric that quantifies the separability of the target and background in the color space. The optimization aims to minimize the cumulative travel distance while maximizing the detectability score. This process is mathematically represented in Equation \ref{eq:jeon}.

\begin{equation}\label{eq:jeon}
\min_{\sigma} \sum_{i=0}^{N-1} \| \mathbf{x}_{c,i} - \mathbf{x}_{c,i+1} \| + \lambda \sum_{i=1}^{N} L(\mathbf{x}_{c,i} \mid \hat{\mathbf{T}}_{a,i}),
\end{equation}

Subject to the constraints: $\|\mathbf{x}_{c,i} - \mathbf{x}_{a,i}\| = r_d$, ensuring the drone maintains a fixed distance from the target, and $\|\mathbf{x}_{c,i} - \mathbf{x}_{c,i+1}\| \leq r_{\text{max}}$, bounding the maximum inter-step travel distance. Here, $\mathbf{x}_{c,i}$ denotes the drone's position, $\hat{\mathbf{T}}_{a,i}$ is the predicted target pose, and $L(\cdot)$ represents the detectability cost function. Additionally, a smooth and dynamically feasible trajectory is generated using quadratic programming \cite{chen2016tracking}, which interpolates the discrete path while minimizing high-order derivatives for smooth motion, ensuring real-time applicability in dynamic scenarios.

The method in \cite{burg2020real} ensures smooth, predictable camera movements while avoiding occlusions in complex 3D environments. It generates an occlusion anticipation map (A-map) to predict future occlusions and adjusts the camera's motion using a physics-driven model. When local solutions fail, strategies like look-ahead searches \cite{agarwal2018modellearninglookaheadexploration, 8796031} or “cuts” \cite{10.5555/3059320.3059326} provide optimal viewpoints, maintaining continuous, unobstructed views in dynamic scenes.

\blue{The focus of \cite{ashtari2020capturing} was to enable drones to autonomously capture subjective first-person view (FPV) shots by imitating human camera operator motion for immersive cinematography. The proposed method models human walking dynamics and uses a constrained optimization framework to compute drone control commands that replicate these motions while adhering to user-defined trajectories and the drone’s physical constraints. Operating in real time, it allows interactive parameter adjustments and seamless transitions between shot styles in various environments.}

The approach in \cite{gebhardt2021optimization} tackles challenges in aerial cinematography by optimizing trajectories to maintain proper framing of 3D targets like landmarks while adhering to user intentions. By integrating compositional rules like the Rule of Thirds \cite{malevs2012compositional, amirshahi2014evaluating} and penalizing deviations from user-specified target positions, the method ensures targets stay fully visible in the frame. Using infinite horizon contour-following equations \cite{gebhardt2018optimizing} in a multi-objective optimization framework, it balances smooth motion, framing, and visibility for high-quality aerial video footage.

The method in \cite{yu2022bridging} addresses the challenge of aligning virtual camera content with both aesthetic and script fidelity requirements. Prior approaches often prioritize aesthetic rules at the expense of accurately reflecting the script's intent. To overcome this, the authors propose a unified framework that minimizes a weighted sum of aesthetic distortion ($D_a$) and fidelity distortion ($D_f$), as formalized in Equation \ref{eq:bridge}. Using dynamic programming \cite{bellman1966dynamic}, this recursive approach ensures that decisions about the current frame's camera configuration do not depend on earlier choices, allowing the use of dynamic programming for efficient computation.


\begin{equation}\label{eq:bridge}
\begin{aligned}
    D_k(z_{k-q}, \ldots, z_k) &= \min_{z_{k-q-1}, \ldots, z_{k-1}} \bigg\{ D_{k-1}(z_{k-q-1}, \ldots, z_{k-1}) \\ 
    &\quad + \frac{\lambda}{T} [\alpha O(c_k) + \beta] \\ 
    &\quad + (1 - \lambda) \big[\omega_0 V(c_k) + \omega_1 C(c_k) + \omega_2 A(c_k) \\ 
    &\qquad + \omega_3 S(c_k, c_{k-1}) + \omega_4 M(c_k, c_{k-1})\big] \\ 
    &\quad + (1 - \lambda) \cdot \\
    &\quad (1 - \omega_0 - \omega_1 - \omega_2 - \omega_3 - \omega_4) \cdot \\ 
    &\quad U(u, c_k, c_{k-1}, \ldots, c_{k-q}) \bigg\}
\end{aligned}
\end{equation}

Each term in Equation \ref{eq:bridge} corresponds to different aspects: $D_{k-1}$ refers to the accumulated distortion up to the previous frame; $\lambda$ is a weighting factor balancing fidelity and aesthetic distortions; $O(c_k)$ quantifying occlusion; $V(c_k)$ character visibility distortion; $C(c_k)$ camera configuration distortion; $A(c_k)$  Action alignment distortion; $S(c_k, c_{k-1})$  Screen continuity distortion; $M(c_k, c_{k-1})$  Motion continuity distortion; $U(u, c_k, \ldots, c_{k-q})$: Shot duration distortion; $Z_k$ and other parameters are trainable.

CineMPC, introduced in \cite{pueyo2022cinempc}, optimizes both extrinsic  and intrinsic parameters of UAV-mounted cameras for autonomous cinematography. Using a non-linear Model Predictive Control (MPC) framework \cite{schwenzer2021review}, it minimizes a cost function balancing cinematic goals, physical constraints, and artistic guidelines. By solving for optimal movements over a finite time horizon, the system adapts to dynamic targets, producing smooth, cinematic-quality footage.

This section explored a range of algorithms designed for real-world drone applications, focusing on those that try to optimize delays while accounting for the drone's physical constraints and the problem's unique nature. Although these algorithms are efficient and can operate with minimal delay, they often struggle with accuracy, particularly in generating smooth trajectories. Most of the methods navigate between two or more points or targets to record footage, yet they frequently fall short when it comes to planning more complex, seamless paths that are essential for optimal drone operation.

Optimization-based techniques frame trajectory generation as an objective-driven process, using metrics to evaluate shot quality. Classical approaches, such as gradient-based methods and genetic algorithms, excel in balancing enforceable constraints and optimizable properties. While these methods are effective for applications like drone cinematography, where real-time responses are critical, challenges such as high computational demands and limited flexibility persist. Table \ref{tab:OCTG} outlines various optimization techniques, emphasizing their role in addressing dynamic and constrained environments.

\begin{table*}%
\caption{Overview of Optimization Methods for Camera Trajectory Generation Methods}
\label{tab:OCTG}
\begin{center}
\begin{tabular}{lcccc}
  \toprule
  Method & Type & \makecell[c]{Real World} & Virtual & \makecell[c]{Camera Movement} \\ \midrule

  \rule{0pt}{\rowheightmethods} \cite{kamada1988views} & \cellcolor[HTML]{FFC7CE}7-DOF & Human-Based & Animation/Games & Non-Fixed \\
  \rule{0pt}{\rowheightmethods} \cite{stuerzlinger1999annealing} & \cellcolor[HTML]{FFC7CE}7-DOF & Human-Based & Animation/Games & Non-Fixed \\
  \rule{0pt}{\rowheightmethods} \cite{fleishman2000automatic} & \cellcolor[HTML]{FFC7CE}7-DOF & Human-Based & - & Fixed \\
  \rule{0pt}{\rowheightmethods} \cite{bares2000model} & \cellcolor[HTML]{FFC7CE}7-DOF & - & Animation/Games & Non-Fixed \\
  \rule{0pt}{\rowheightmethods} \cite{halper2001camera} & \cellcolor[HTML]{FFC7CE}7-DOF & Human-Based & - & Non-Fixed \\
  \rule{0pt}{\rowheightmethods} \cite{vazquez2003entropy} & \cellcolor[HTML]{FFC7CE}7-DOF & Human-Based & Animation/Games & Non-Fixed/Fixed \\
  \rule{0pt}{\rowheightmethods} \cite{bourne2005applying} & \cellcolor[HTML]{FFC7CE}7-DOF & - & Games & Non-Fixed \\
  \rule{0pt}{\rowheightmethods} \cite{viola2006geometry} & \cellcolor[HTML]{FFC7CE}7-DOF & - & - & Fixed \\
  \rule{0pt}{\rowheightmethods} \cite{stoev2006paths} & \cellcolor[HTML]{FFC7CE}7-DOF & - & Animation/Games & Non-Fixed \\

  \rule{0pt}{\rowheightmethods} \cite{Lino2012} & \cellcolor[HTML]{C6EFCE}LDO & - & Animation/Games & Non-Fixed \\
  \rule{0pt}{\rowheightmethods} \cite{Lino2015} & \cellcolor[HTML]{C6EFCE}LDO & - & Animation/Games & Non-Fixed \\
  \rule{0pt}{\rowheightmethods} \cite{Galvane2015} & \cellcolor[HTML]{C6EFCE}LDO & - & Animation/Games & Non-Fixed \\
  \rule{0pt}{\rowheightmethods} \cite{ren2023automatic} & \cellcolor[HTML]{C6EFCE}LDO & Human-Based & - & PTZ \\

  \rule{0pt}{\rowheightmethods} \cite{roberts2016generating} & \cellcolor[HTML]{DDEBF7}DTO & Areal-Based & - & Gimbal Mounted \\
  
    \rule{0pt}{\rowheightmethods} \cite{gebhardt2016airways} & \cellcolor[HTML]{DDEBF7}DTO & Areal-Based & - & Gimbal Mounted \\
  
  \rule{0pt}{\rowheightmethods} \cite{nageli2017real} & \cellcolor[HTML]{DDEBF7}DTO & Areal-Based & - & Gimbal Mounted \\
  \rule{0pt}{\rowheightmethods} \cite{nageli2017multi} & \cellcolor[HTML]{DDEBF7}DTO & Areal-Based & - & Gimbal Mounted \\
  \rule{0pt}{\rowheightmethods} \cite{bonatti2020autonomous} & \cellcolor[HTML]{DDEBF7}DTO & Areal-Based & Animation/Games & Gimbal Mounted \\
  \rule{0pt}{\rowheightmethods} \cite{rousseau2018quadcopter} & \cellcolor[HTML]{DDEBF7}DTO & Areal-Based & - & Gimbal Mounted \\
  
    \rule{0pt}{\rowheightmethods} \cite{gebhardt2018optimizing} & \cellcolor[HTML]{DDEBF7}DTO & Areal-Based & - & Gimbal Mounted \\

  \rule{0pt}{\rowheightmethods} \cite{galvane2018directing} & \cellcolor[HTML]{DDEBF7}DTO & Areal-Based & - & Gimbal Mounted \\
  \rule{0pt}{\rowheightmethods} \cite{jeon2019online} & \cellcolor[HTML]{DDEBF7}DTO & Areal-Based & - & Gimbal Mounted \\
  \rule{0pt}{\rowheightmethods} \cite{sabetghadam2019optimal} & \cellcolor[HTML]{DDEBF7}DTO & Areal-Based & - & Gimbal Mounted \\
  \rule{0pt}{\rowheightmethods} \cite{bonatti2019towards} & \cellcolor[HTML]{DDEBF7}DTO & Areal-Based & - & Gimbal Mounted \\
  \rule{0pt}{\rowheightmethods} \cite{bonatti2019autonomous} & \cellcolor[HTML]{DDEBF7}DTO & Areal-Based & - & Gimbal Mounted \\
  \rule{0pt}{\rowheightmethods} \cite{katoch2019edge} & \cellcolor[HTML]{DDEBF7}DTO & Areal-Based & - & Gimbal Mounted \\
  \rule{0pt}{\rowheightmethods} \cite{jeon2020detection} & \cellcolor[HTML]{DDEBF7}DTO & Areal-Based & - & Gimbal Mounted \\
  \rule{0pt}{\rowheightmethods} \cite{burg2020real} & \cellcolor[HTML]{DDEBF7}DTO & Areal-Based & - & Gimbal Mounted \\
  \rule{0pt}{\rowheightmethods} \cite{ashtari2020capturing} & \cellcolor[HTML]{DDEBF7}DTO & Areal-Based & - & Gimbal Mounted \\
  \rule{0pt}{\rowheightmethods} \cite{gebhardt2021optimization} & \cellcolor[HTML]{DDEBF7}DTO & Areal-Based & - & Gimbal Mounted \\
  \rule{0pt}{\rowheightmethods} \cite{yu2022bridging} & \cellcolor[HTML]{DDEBF7}DTO & Areal-Based & - & Gimbal Mounted \\
  \rule{0pt}{\rowheightmethods} \cite{pueyo2022cinempc} & \cellcolor[HTML]{DDEBF7}DTO & Areal-Based & - & Gimbal Mounted \\

  \bottomrule
\end{tabular}
\end{center}
\bigskip\centering
\footnotesize\emph{Note:} All the entries are entered based on evidence or our evaluation.
\end{table*}%

\subsection{Machine Learning}
Camera trajectory generation has seen remarkable advancements through machine learning in recent years \cite{courant2025exceptional,jiang2024cinematographic,wang2024dancecamera3d}. Traditional methods based on optimization and handcrafted rules have progressively been complemented by data-driven approaches, which enable the automation of trajectory synthesis by learning complex patterns from examples. These methods offer greater flexibility and adaptability compared to traditional approaches, effectively addressing their shortcomings\cite{wang2024dancecamera3d}. By leveraging deep learning models, these methods not only incorporate cinematic principles and adapt to diverse constraints but also provide the ability to generate diverse and creative camera trajectories \cite{jiang2020example,dehghanian2025lenscraft}. This paradigm shift has expanded the creative capabilities of camera movement systems, enhancing their efficiency, with generative models serving as a cornerstone for these advancements \cite{courant2025exceptional,jiang2024cinematographic}. In the following, we examine the evolution of these methods.

One of the earliest efforts to apply machine learning to camera trajectory generation was presented by Chen et al. \cite{chen2016learning}, where Recurrent Random Forests were utilized to predict the pan angle of a camera in sports events. This study introduced a novel method for optimizing random forest models, wherein each prediction was dependent solely on the previous one. This dependency on the prior state ensured that the generated camera trajectory maintained the necessary smoothness and continuity. Simply put, this approach employed random forests within a Markovian structure to synthesize camera trajectories.

\begin{figure}[t]
     \centering
    \includegraphics[width=0.5\linewidth]{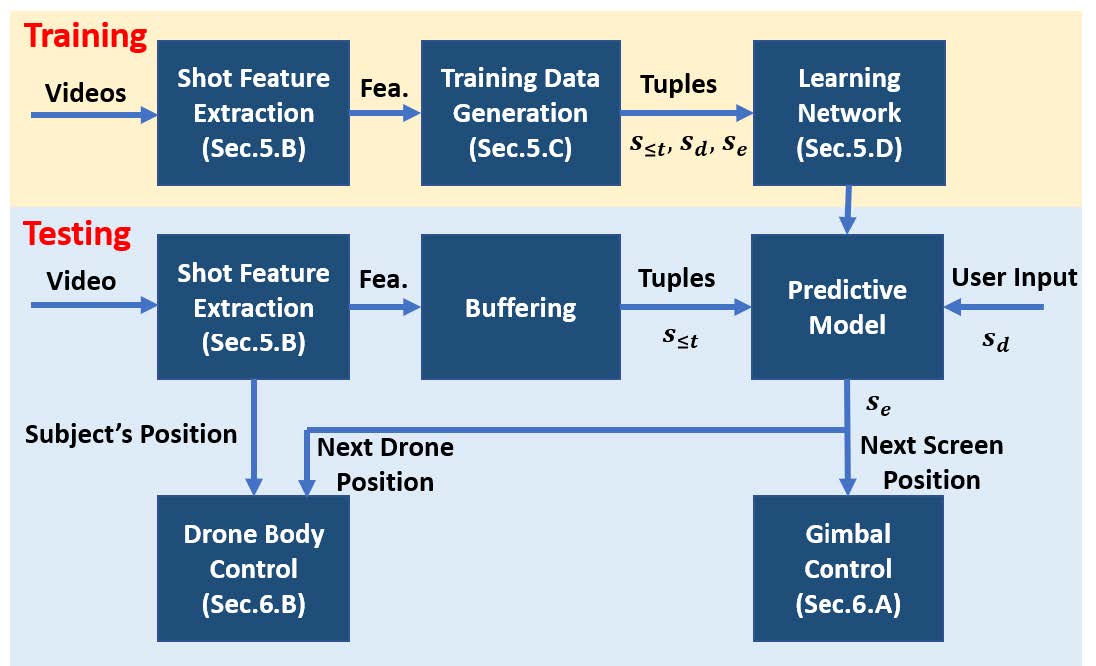}
    \caption{The framework of imitation filming \cite{huang2019learning}}
      \label{fig:imitation_filming}
\end{figure}

\blue{In the paper introduced in \cite{huang2019learning}, a data-driven learning-based approach is proposed to enable drones to autonomously capture cinematic footage by imitating professional camerawork. Unlike traditional methods that rely on predefined camera movements or heuristic planning (i.e., rule-based methods), the proposed framework employs supervised learning to predict future image composition and camera position, subsequently generating control commands to achieve professional shot framing. The framework of imitation filming introduced in this paper is illustrated in Figure \ref{fig:imitation_filming}.}

In their 2020 paper, Christos Kyrkou et al. \cite{kyrkou2020imitation} propose an end-to-end approach for active camera control using deep convolutional neural networks to address limitations of traditional multi-stage systems. Their model, named ACDCNet, combines visual detection and camera motion control in a single framework, using imitation learning to train the network on image-action pairs. The study demonstrates significant improvements in multi-target tracking, efficiency, and real-time performance compared to conventional methods.

\begin{figure}[t]
     \centering
    \includegraphics[width=0.8\linewidth]{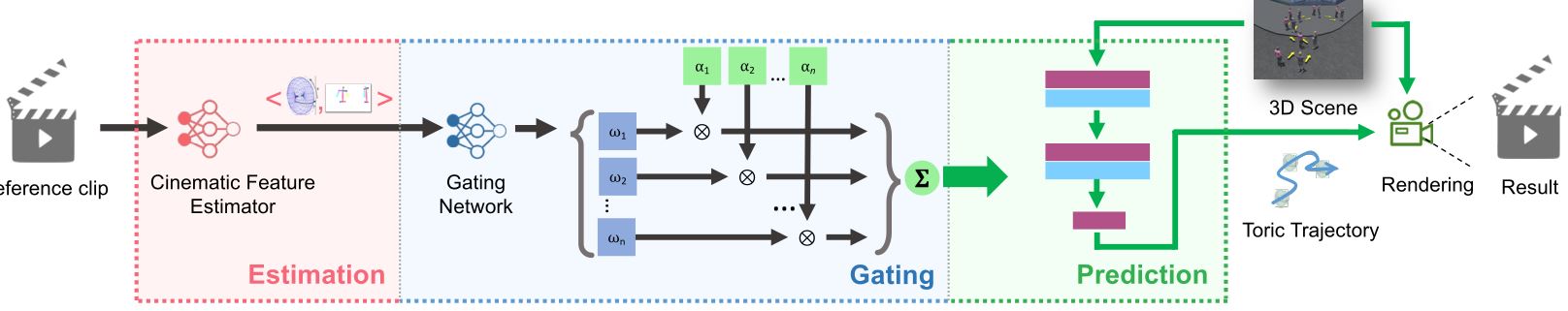}
    \caption{The model presented in the article \cite{jiang2020example} for transferring cinematic features from a reference video.}
      \label{fig: example}
\end{figure}

The 2020 paper Example-driven Virtual Cinematography by Learning Camera Behaviors \cite{jiang2020example} proposed a framework for transferring camera behaviors from one video to another. First, they extract a raw skeleton, followed by refinement method and then with a neural network to estimate the camera position in toric space. For trajectory generation, they utilized a mixture of experts framework, incorporating an LSTM followed by a fully connected layer as a gating network to determine the weighting of each expert. Each expert, implemented as a three-layer fully connected network, predicted new camera poses by processing character cinematic features from a 3D animation and information from past frames. Figure \ref{fig: example} shows the architecture of the model proposed in this paper.

The paper \cite{jiang2021camera} was published with the aim of adding more precise control over camera movement using key points. This research, building on the work in \cite{jiang2020example}, incorporates the ability to control the camera trajectory through key points rather than solely following a reference video.

\begin{figure}[t]
  
     \centering
    \includegraphics[width=0.8\linewidth]{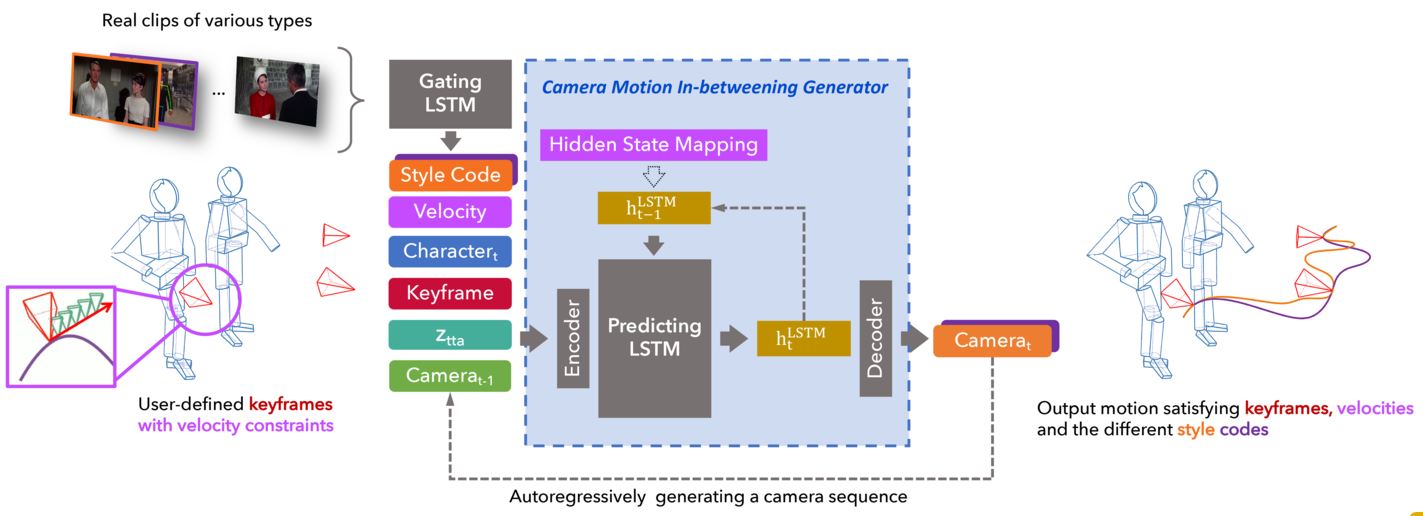}
    \caption{The architecture of the model \cite{jiang2021camera} for generating camera trajectories based on a reference video and key points. }
      \label{fig: keyframing}
\end{figure}

In their new architecture, the previous feature extraction model is still used to process the reference video, but the trajectory generation structure has been redesigned. Instead of employing a complex Mixture of Experts (MoE) architecture with multiple fully connected networks, an LSTM is used to extract embeddings from the reference video. This structural change simplifies the architecture and enhances the model's ability to understand the temporal features of camera movement. During the trajectory generation stage, the extracted embedding, along with the camera key point information, character positions, and the previous camera position, is fed into an LSTM network. This network operates in an autoregressive, step-by-step manner to generate the camera's positions. In Figure \ref{fig: keyframing}, the overall architecture of this network is illustrated. 

Kyrkou et al. (2021) proposed C3NET \cite{kyrkou2021c}, a lightweight neural network designed for real-time camera control through direct end-to-end learning from visual input to pan-tilt motion commands. Unlike traditional approaches that rely on multiple modules for detection, tracking, and control, C3NET learns to map raw image pixels directly to camera movement parameters without requiring explicit object detection or bounding box annotations. The network implicitly learns to identify targets and determine appropriate camera movements to keep them centered in the field of view. Their architecture consists of two main components: a feature extractor with convolutional blocks for processing visual information, and a fully connected controller subnetwork that maps these features to camera motion controls. 

A study in 2021 introduced trajectory tensors for Multi-Camera Trajectory Forecasting (MCTF), addressing limitations of traditional coordinate-based methods \cite{styles2021multi}. Unlike coordinate trajectories, which struggle with occlusions and multiple camera views, trajectory tensors represent object locations as heatmaps across cameras and timesteps, capturing spatial and temporal information in a unified form. This approach handles null trajectories, accounts for object scale, and models uncertainty in trajectory forecasting. The authors demonstrate its effectiveness using various models, including 3D-CNNs and CNN-GRU, which leverage the trajectory tensor representation for improved spatiotemporal forecasting.

In 2021 also, a deep reinforcement learning (RL) framework with an attention-based approach was proposed for virtual cinematography of 360-degree videos \cite{Wang2021AttentionBased}. This work aimed to replicate the viewpoint selection of professional cinematographers by integrating saliency detection and RL techniques. The proposed system utilized a DenseNet architecture to process both video content and saliency maps simultaneously. The RL component managed narrow field of view selection as a continuous action space, with a reward function designed to balance saliency, alignment with ground-truth views, and smoothness of camera transitions.

The paper Enabling Automatic Cinematography with Reinforcement Learning \cite{Yu2022Enabling} introduced a new RL approach using Proximal Policy Optimization (PPO) to train camera settings for virtual environments. The reward function was designed to optimize the camera's position and angle by minimizing the absolute difference from the ground truth, scaled by a factor of either 180 or 30 depending on the specific parameter. This approach effectively allowed the system to learn context-aware camera placements through reinforcement learning.

The 2023 paper, The Secret of Immersion: Actor-Driven Camera Movement Generation for Auto-Cinematography \cite{wu2023secret}, introduced a deep camera control framework designed to achieve actor-camera synchronization across three dimensions: frame aesthetics, spatial action, and emotional status. The approach begins with a user-provided initial camera position and utilizes the rule of thirds in a self-supervised manner to refine the camera's placement. This is achieved by incorporating a loss function based on the distance from the rule of thirds, along with minimizing differences in the generated trajectory. The framework further employs a generator trained using a combination of Mean Squared Error (MSE) loss, differences in features extracted by a VGG network, amplitude loss, and adversarial loss to learn and produce smooth and context-aware camera trajectories.

The paper Adaptive Auto-Cinematography in Open Worlds \cite{Yu2023Adaptive} addressed the unique challenges of user interaction in video games. Unlike traditional cinematographic approaches that emphasize cinematic rules, this method prioritized user interaction and the dynamic nature of open-world environments. The study highlighted the limitations of example-driven methods, particularly their inability to adapt to the uncertainty of targets, such as the main character in open-world games. To address these challenges, a GAN-based model was proposed to incorporate user interaction into the generation of camera trajectories. Additionally, new metrics were developed to evaluate the generated trajectories, accounting for the complexities of the task.

Building on this work, a follow-up study, Automated Adaptive Cinematography for User Interaction in Open Worlds \cite{Yu2024Automated}, enhanced the initial framework by introducing skeleton poses of the characters and their actions as conditions for the GAN model. This addition improved the ability of the model to generate contextually adaptive and realistic camera trajectories, further aligning the camera movement with the dynamic interaction of users and characters in open-world settings.

\begin{figure}[t]
  
     \centering
    \includegraphics[width=0.8\linewidth]{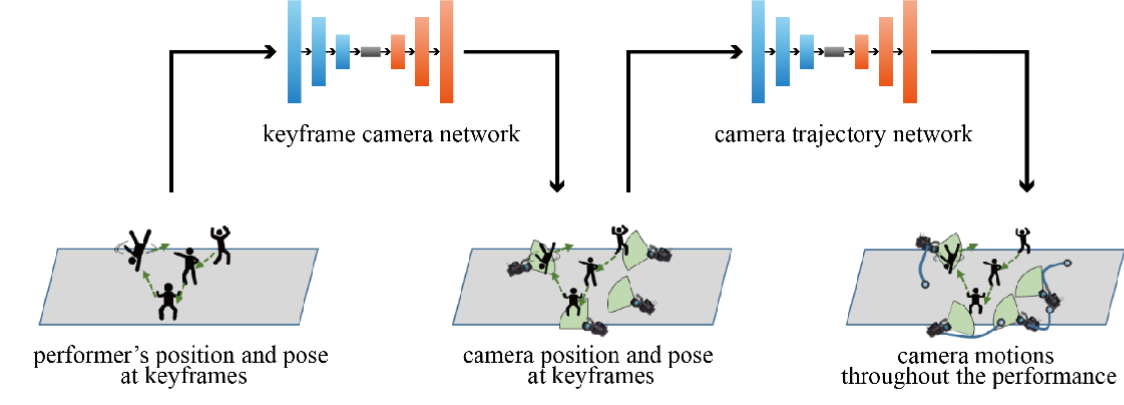}
    \caption{A two stage transformer based architecture proposed in \cite{Xie2023Camera}}
      \label{fig: transformer-two-stage}
\end{figure}

In \cite{Xie2023Camera}, a transformer-based approach was proposed for generating camera trajectories and motions in real-time environments. The method operates in two stages: first, it utilizes the performers' positions and orientations, as defined in the stage script, to set the initial placements and postures of the camera for the entire sequence. These initial positions serve as keyframes, predetermined by the script. In the second stage, the model uses these keyframes as input to generate smooth camera motion between them, adapting to the live placements and orientations of the performers. The network architecture integrates a Transformer with relative position encoding, which the authors state enables more effective learning of camera motion features in comparing to standard Transformer architectures. In figure \ref{fig: transformer-two-stage}

The year 2024 represented a turning point with the rise of diffusion models \cite{ho2020denoising}, whose growing popularity led to diverse applications ranging from direct use in generating camera trajectories \cite{jiang2024cinematographic, li2024director3d, courant2025exceptional} to indirect uses such as creating images with specific camera shot types \cite{Massaglia2024DreamShot}.

A study extending the work of \cite{jiang2021camera} was presented at the Eurographics conference \cite{jiang2024cinematographic}, introducing the use of diffusion-based models for camera trajectory generation for the first time. This system is capable of generating camera movements based on a complete or partial prompt that includes all or part of the standard framing, angle, and motion features, along with optional key points defined by the user at the beginning and end of the trajectory.

\begin{figure}[t]
     \centering
    \includegraphics[width=0.8\linewidth]{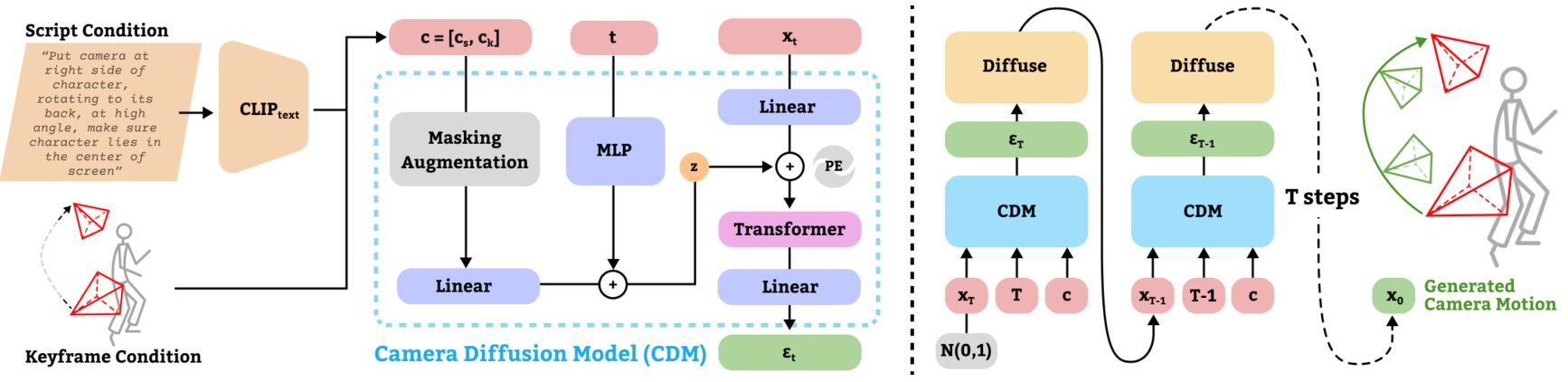}
    \caption{The architecture proposed in \cite{jiang2024cinematographic}, utilizing diffusion-based models with a transformer architecture.}
    \label{fig: CCD}
\end{figure}

In this architecture, the CLIP model \cite{radford2021learning} is used to encode textual descriptions, which are then combined with key point information. Unlike their previous studies \cite{jiang2020example,jiang2021camera} that relied on LSTM-based architectures, this method employs a diffusion-based model with a transformer architecture at each step of the generation process. The proposed architecture of this study is illustrated in Figure \ref{fig: CCD}. 

Another study, published in 2024 under the title E.T. \cite{courant2025exceptional}, introduced a new dataset for camera trajectory generation along with proposing three diffusion-based architectures.

The first proposed architecture, "Director A", utilizes a relatively simple approach to apply conditions; Here, textual descriptions and the subject's trajectory are added as context tokens to the transformer's input. 
In the second architecture, "Director B",  the conditions are concatenated into a single token vector and these vectors are then used to adjust AdaLN parameters before each self-attention and feed-forward layer.

In the final model, "Director C", the CLIP prompt embeddings and the subject's trajectory are combined and processed through two transformer encoder layers. This information is then applied to the main model via a cross-attention block, enabling the use of more intricate patterns in the conditions. These three architecture is  illustrated in Figure \ref{fig: ET-model}:

\begin{figure}[t]
     \centering
    \includegraphics[width=0.8\linewidth]{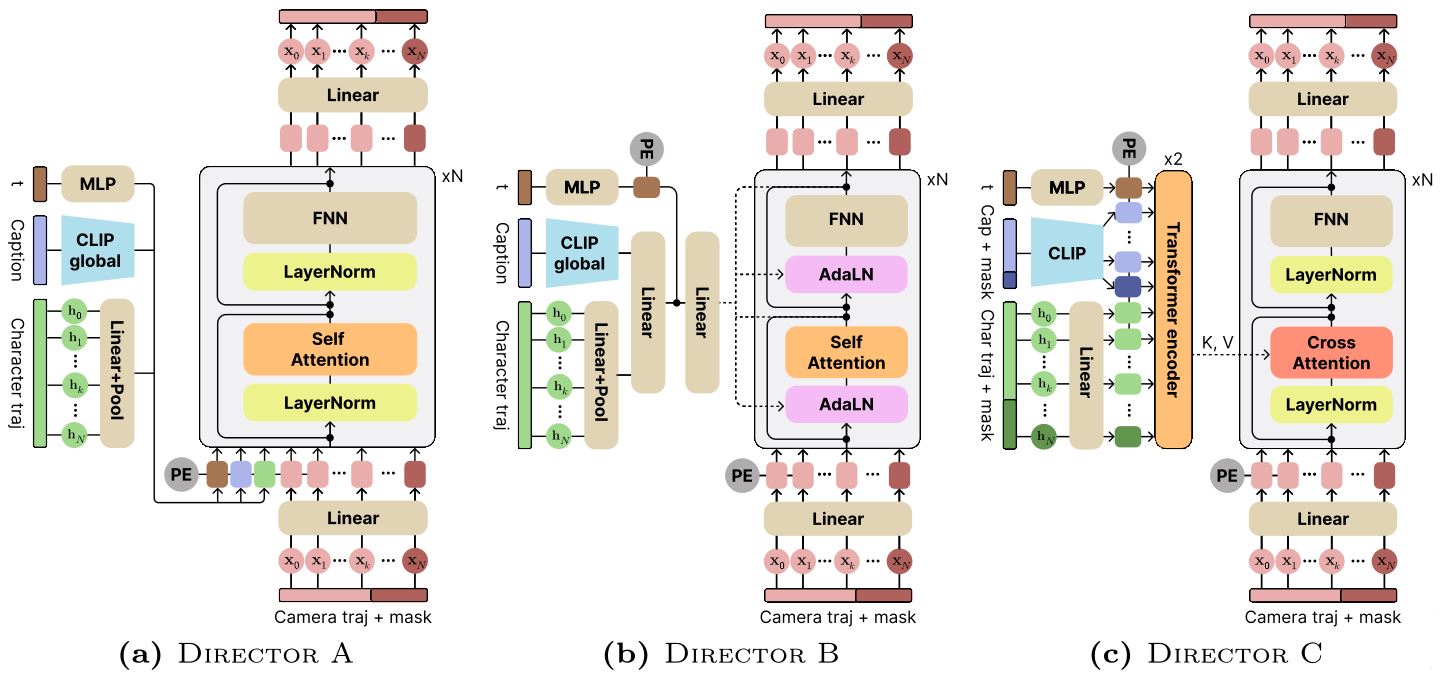}
    \caption{Architectures proposed in \cite{courant2025exceptional}}
    \label{fig: ET-model}
\end{figure}

In an upcoming study, LensCraft \cite{dehghanian2025lenscraft} tries to solve three critical challenges in virtual cinematography. First, it introduces a comprehensive cinematographic language paired with a dedicated simulation framework to generate balanced, high-quality, controlled training data through expert consultation - addressing the persistent issue of dataset bias and quality in existing systems. Second, it presents a dual-level representation system, allowing simultaneous conditioning on multiple inputs (text, keyframes, and reference trajectories) while maintaining cinematographic integrity. Also, the model's leverage progressive masking strategy and CLIP-based embedding approach enable it to learn meaningful interpolations between different camera movements while preserving semantic coherence. 


Next paper \cite{wang2024dancecamera3d} specifically focused on generating camera movements for Dance scenes, introducing a novel approach that combines musical information with the subject's motion to produce synchronized and context-aware camera trajectories.

\begin{figure}[t]
     \centering
    \includegraphics[width=0.8\linewidth]{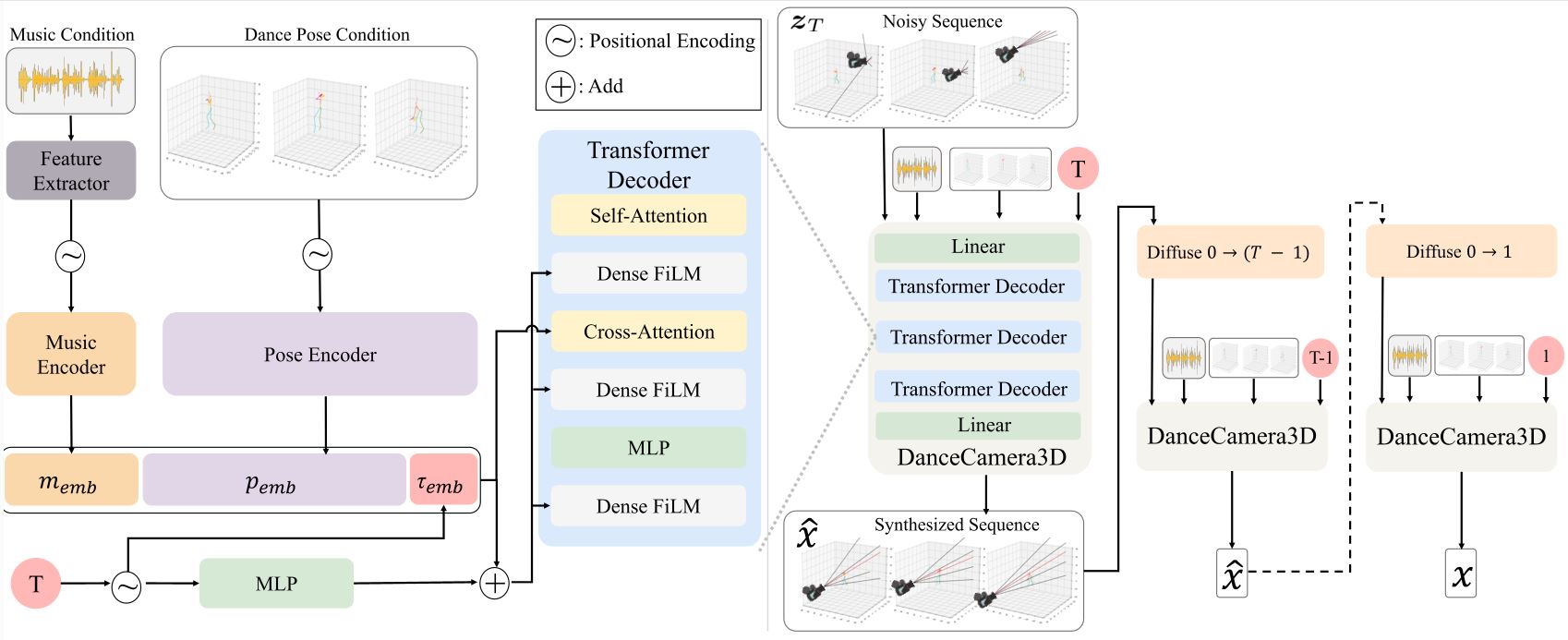}
    \caption{The DCM model architecture, based on a combination of transformer and diffusion networks \cite{wang2024dancecamera3d}.}
    \label{fig: DCM}
\end{figure}

The proposed architecture like previous models \cite{courant2025exceptional, jiang2024cinematographic}, utilizing a combination of transformer models and diffusion networks. Musical data and the subject's pose are embedded and combined, and then used this embedding in the transformer's cross-attention blocks. The model's final architecture consists of multiple sequential transformer decoders that execute the diffusion denoising process to generate the final camera trajectory. For conditioning, the model employs a Classifier Free Guidance-based approach \cite{ho2022classifier}, which is a well-known method for conditioning diffusion-based models. The architecture of the DCM model proposed in this research is illustrated in Figure \ref{fig: DCM}.

A recent continuation of the DCM model introduced the DanceCamAnimator framework \cite{wang2024dancecamanimator}, designed to address the limitations of the previous model by incorporating support for keyframing. This framework adopts a three-stage approach for generating camera movements in the context of music and dance, utilizing animator expertise to identify and produce keyframes as well as predict tween functions and tries to reduce the need for post-processing.

\begin{figure}[t] \centering 
{ \includegraphics[width=0.8\linewidth]{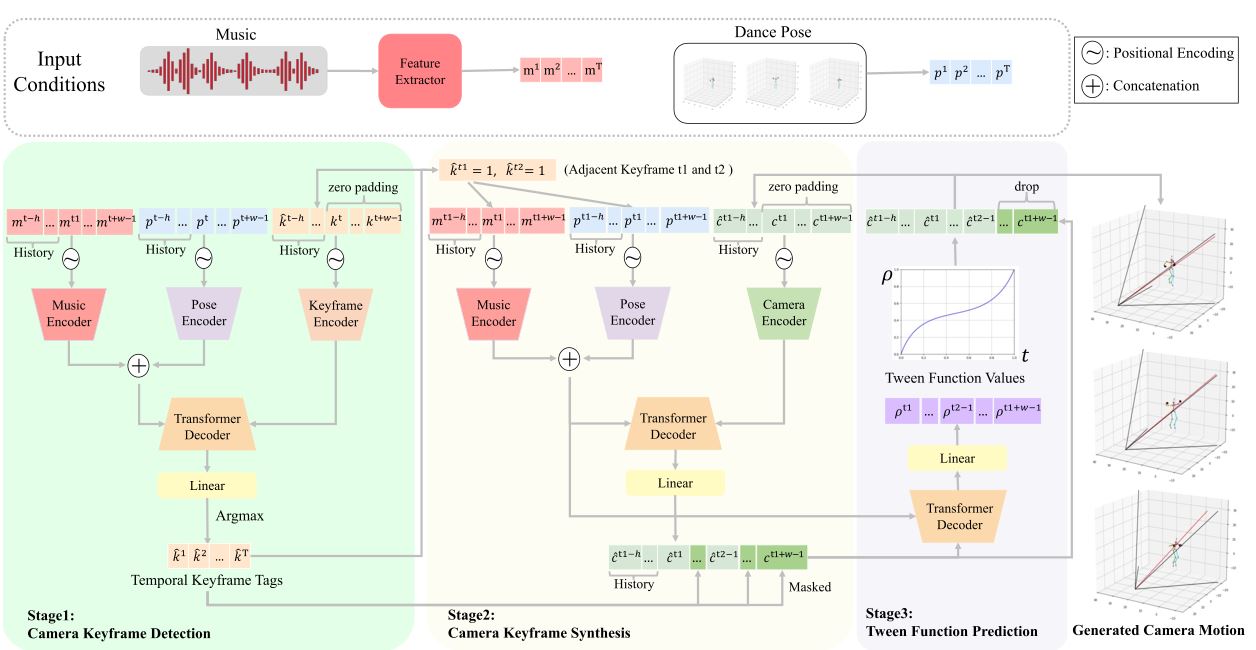} \caption{The architecture proposed in \cite{wang2024dancecamanimator} for modeling keyframes.} \label{fig: DanceCamAnimator} } \end{figure}

In the first stage, the model identifies camera keyframes by analyzing subject movements, musical representation, and the temporal history of key points to determine critical moments for significant camera adjustments. In the second stage, the model generates the camera’s position and movement for these keyframes. Finally, in the third stage, it predicts tween function values for in betweening keyframes to ensure smooth and natural transitions between them. Figure \ref{fig: DanceCamAnimator} depicts the stages of the DanceCamAnimator framework.

Jawad et al. \cite{jawad2024deep} explored camera control in robotic surgery by utilizing both dense neural network (DNN) and recurrent neural network (RNN) architectures trained on combined datasets of autonomous and human-operated camera trajectories \cite{jawad2024deep}. Unlike previous single-mode approaches, their method learned to merge the predictable behavior of rule-based systems with the adaptive nature of human operation, achieving the advantages of both. The DNN architecture demonstrated proficiency in basic tool tracking, while the RNN, excelled at learning timing-based camera zooming and complex motion patterns and achieved sub-millimeter accuracy, suggesting superior performance in real surgical scenarios where precise camera control is crucial. 

Some works address camera trajectory generation not as their primary focus but as a secondary or complementary task integrated within their frameworks to address other problems. The remainder of this section reviews these works.

Among these works Director3D \cite{li2024director3d} is a framework that integrates camera trajectory generation as part of a text-to-3D video generation process. The system, Director3D, begins by utilizing a Trajectory Diffusion Transformer \cite{peebles2023scalable} to model the distribution of camera trajectories from textual prompts. This phase, referred to as the "Cinematographer" step, generates adaptive camera paths tailored to the scene described in the input prompt. The generated trajectories serve as the input for subsequent steps, which involve creating a 3D scene and aligning it with the predefined camera motion.

Another framework that incorporates camera trajectory generation within a broader video generation task is MotionCtrl \cite{wang2024motionctrl}, which introduces a Camera Motion Control Module to effectively handle camera movements. This module extends the Denoising U-Net structure of the Latent Video Diffusion Model \cite{he2022latent} by integrating camera pose into second self-attention module and applying a fully connected layer to extract temporal features. These modifications allow the model to conditionally generate videos where the background and object movements align with the specified camera poses and trajectories.

\blue{The work in \cite{xie2023gait} addresses the task of generating aesthetically pleasing camera trajectories in synthetic 3D indoor scenes. The proposed method, GAIT, is a Deep Reinforcement Learning (DRL) framework that optimizes camera movements in a 5D space using a neural aesthetic model trained on crowd-sourced data. It employs a reward function integrating aesthetic evaluation, temporal smoothness, and diversity regularization to ensure smooth and diverse trajectories. GAIT uses visual DRL algorithms like DrQ-v2 \cite{zhou2024continuous} and CURL \cite{laskin2020curl}, leveraging data augmentation and contrastive learning to efficiently generate visually appealing and contextually diverse camera paths.}

Another approach addressing camera trajectory generation within a text-to-video framework is Direct-a-Video \cite{yang2024direct}. This model incorporates camera position generation by encoding three parameters: horizontal pan, vertical pan, and zoom ratio. 
The horizontal and vertical pan values are encoded using a Fourier embedder, while the zoom ratio directly passed through MLPs and then the resulting embeddings are combined to represent the camera movement in a temporal cross-attention mechanism to guide the generation of video sequences aligned with the specified camera movements and object interactions.

Next work integrates camera trajectory generation within a broader application is CinePreGen \cite{chen2024cinepregen}. This work introduces a previsualization framework and new coordinate system, CineSpace. This Space is based on Toric allows users to control camera movements for storyboarding purposes. Their framework offers 15 common rule-based options for defining camera trajectories. The camera dynamics are further enhanced by incorporating multi-masked IP-Adapter techniques and engine simulation, ensuring alignment with ground truth information throughout the rendering process.

Liu et al. \cite{xu2024camco} present a method for generating camera-controllable, geometry-consistent videos by integrating camera control into a pre-trained image-to-video diffusion model. They use Plücker coordinates for 6-DoF camera parameterization, enabling dynamic viewpoint adjustments across frames. A key innovation is the epipolar constraint attention mechanism, which ensures geometric consistency by aligning features between frames. The model is fine-tuned from Stable Video Diffusion (SVD), incorporating temporal noise scheduling and classifier-free guidance to maintain high-quality, temporally consistent videos while adhering to specified camera trajectories.

The approach introduced in \cite{kuang2024collaborative} builds upon CameraCtrl \cite{he2024cameractrl} and the consistency model from \cite{tseng2023consistent}, proposing a method for generating synchronized multi-view videos. The key innovation is the Cross-View Synchronization Module (CVSM), which uses masked attention and fundamental matrices to ensure structural consistency across video frames. This enables the model to generate temporally coherent videos from different camera trajectories while maintaining alignment across views. The model is trained on pairs of videos, leveraging datasets such as RealEstate10K and WebVid10M.

DreamCinema \cite{chen2024dreamcinema} is another framework that incorporates camera trajectory as part of a broader cinematic transfer process. This framework focuses on simplifying film creation by allowing camera movement transferring from source video and 3D character integration. It extracts camera trajectories from reference videos and optimizes them using motion-aware guidance and physical modeling with Bézier curves \cite{zhang1999cbezier}. The framework then continues its process to generate a new video, where the transferred camera movement is applied seamlessly to the newly created scenes.

\blue{The work in \cite{bar2024navigationworldmodels} addresses the task of camera trajectory generation for navigation in both known and unknown environments. It introduces the Navigation World Model (NWM), a machine learning-based approach that uses a novel Conditional Diffusion Transformer (CDiT) \cite{bar2024navigationworldmodels}. The NWM predicts future visual states based on past observations and navigation actions, allowing for the simulation of trajectories to achieve specified goals. The CDiT, a diffusion-based autoregressive model, is trained on diverse egocentric video datasets from human and robotic agents. Unlike standard diffusion transformers (DiTs), which compute self-attention over all input tokens with quadratic complexity, the CDiT employs a cross-attention mechanism for conditioning on past frames, reducing computational complexity to linear with respect to the number of context frames.}

The field of camera trajectory generation has witnessed remarkable progress through machine learning approaches, evolving from basic statistical models to sophisticated deep learning architectures. The transition from LSTM-based models to transformer architectures, and most recently to diffusion-based approaches, has significantly enhanced the quality and controllability of generated trajectories. These advancements have enabled more natural, context-aware camera movements while providing flexible conditioning mechanisms through text prompts, keyframes, and multi-modal inputs.

These approaches to camera trajectory generation offer several compelling advantages while facing certain notable challenges. On the positive side, these methods excel at learning complex cinematographic patterns directly from professional examples, capturing nuanced camera behaviors that would be difficult to encode through explicit rules. They also demonstrate remarkable adaptability, automatically adjusting to various scenes and contexts without requiring manual parameter tuning, and can generate diverse, creative camera movements that go beyond predefined templates. 

However, these benefits come with significant trade-offs: the models typically require large datasets of high-quality camera trajectories for training, which are often expensive and challenging to obtain. Additionally, computational costs can be substantial, particularly for sophisticated architectures like diffusion models, making real-time applications challenging. Perhaps most importantly, these approaches often struggle with long-term planning and maintaining global coherence over extended sequences, a crucial aspect of professional cinematography that traditional methods sometimes handle more effectively.

Machine learning has revolutionized camera trajectory generation by enabling data-driven approaches that learn from examples, providing flexibility and adaptability beyond traditional methods. Deep learning models integrate cinematic principles while adapting to complex constraints, facilitating creative and diverse trajectory generation. These methods, detailed in Table \ref{tab:MLCTG}, represent a paradigm shift, with generative models and neural rendering leading to significant advancements in camera trajectory generation.

\begin{table*}%
\caption{Overview of Machine Learning Methods for Camera Trajectory Generation Methods}
\label{tab:MLCTG}
\begin{center}
{\small
\begin{tabular}{lllll}
  \toprule
  Method & \makecell[l]{Real World} & Virtual & Metric & Dataset \\ \midrule

  \rule{0pt}{\rowheightmethods} \cite{chen2016learning} & Human-Based & - & Qual  & Not-Public \\
  
  \rule{0pt}{\rowheightmethods} \cite{huang2019learning} & Human-Based & - & Qual (User Study)  & \makecell[l]{Gathered from internet} \\

  \rule{0pt}{\rowheightmethods} \cite{wang2020attention} & Areal-Based & - & MO - MVD & Sports-360 - Pano2Vid \\ 
  
  \rule{0pt}{\rowheightmethods} \cite{kyrkou2020imitation} & Human-Based & - & Motion Error - FPS & Generated(Not-Public) \\
  \rule{0pt}{\rowheightmethods} & & & Target Tracking  \\

  \rule{0pt}{\rowheightmethods} \cite{jiang2020example} & - & Animation & Accuracy - MA & Synthetic \\ 

  \rule{0pt}{\rowheightmethods} \cite{jiang2021camera} & Human-Based & Animation & Silhouette Distance & Extracted From MovieNet\\
  \rule{0pt}{\rowheightmethods} & & & Trajectory Distance \\
  
  \rule{0pt}{\rowheightmethods} \cite{styles2021multi} & Human-Based & - & SIOU - Average Precision & WNMF \\
  \rule{0pt}{\rowheightmethods} & & & ADE - FDE \\

  \rule{0pt}{\rowheightmethods} \cite{Yu2022Enabling} & - & Animation & Accuracy & Not-Public \\


  \rule{0pt}{\rowheightmethods} \cite{Xie2023Camera} & Human-Based & - & MSE - Qual & MikuMikuDance(MMD) \\

  \rule{0pt}{\rowheightmethods} \cite{Yu2023Adaptive} & - & Games & MSE - Correlation Distance & Not-Public \\ 
  \rule{0pt}{\rowheightmethods} & & & Qual - Multifocus & \\

  \rule{0pt}{\rowheightmethods} \cite{wu2023secret} & Human-Based & Animation & MSE - RoTSft - AdjDis & Synthetic - Artist Design \\ 
  \rule{0pt}{\rowheightmethods} & & &  Hausdorff Distance - CosDA  & \\ 
  \rule{0pt}{\rowheightmethods} & & & LPIPS - FID - VisAcc - PCC & \\
  \rule{0pt}{\rowheightmethods} & & & SRCC - KRCC - AVA \\

  \rule{0pt}{\rowheightmethods} \cite{Yu2024Automated} & - & Games & MSE - Correlation Distance & MineStory \\ 
  \rule{0pt}{\rowheightmethods} & & & Qual - Multifocus & \\

  \rule{0pt}{\rowheightmethods} \cite{massaglia2023dreamshot} & Human-Based & - & CLIP-T Score - DINO - Qual & Not-Public \\ 
  
  \rule{0pt}{\rowheightmethods} \cite{xie2023gait} & Human-Based & - & Aesthetic Score - Qual & Replica \\ 
   \rule{0pt}{\rowheightmethods} & & & Training time - Avg Reward  & \\

  \rule{0pt}{\rowheightmethods} \cite{courant2025exceptional} & Human-Based & - & CLaTr-score - P - R - C - D & ET \\ 
  \rule{0pt}{\rowheightmethods} & & & FDCLaTr - Qual & \\
  
  \rule{0pt}{\rowheightmethods} \cite{dehghanian2025lenscraft} & Volume-Based & Animation & FID - P - R - C - D  & Synthetic\\ 
  \rule{0pt}{\rowheightmethods} & & & Clip-score - Qual \\
  
  \rule{0pt}{\rowheightmethods} \cite{li2024director3d} & Human-Based & - & NIQE - BRISQUE - Qual & MVImgNet - DL3DV-10K \\ 

  \rule{0pt}{\rowheightmethods} \cite{jiang2024cinematographic} & - & Animation & R Precision FID - Diversity & Synthetic \\ 
  \rule{0pt}{\rowheightmethods} & & & Qual - MultiModality & \\ 

  \rule{0pt}{\rowheightmethods} \cite{chen2024cinepregen} & Human-Based & - & Qual  & Not-Public \\

  \rule{0pt}{\rowheightmethods} \cite{chen2024dreamcinema} & - & Animation & PA - IoU - MPJPE - Qual & Not-Public \\ 

  \rule{0pt}{\rowheightmethods} \cite{wang2024dancecamera3d} & Human-Based & Animation & FID - Qual & DCM \\ 
  \rule{0pt}{\rowheightmethods} & & & Euclidean Distance \\

  \rule{0pt}{\rowheightmethods} \cite{wang2024dancecamanimator} & Human-Based & Animation & FID - Qual & DCM \\

  \rule{0pt}{\rowheightmethods} \cite{yang2024direct} & Human-Based & - & Flow Error Metric - Qual & Synthetic from MovieShot \\ 

  \rule{0pt}{\rowheightmethods} \cite{wang2024motionctrl} & Human-Based & - & FID - FVD & Realestate10k for Camera \\ 
  \rule{0pt}{\rowheightmethods} & & & Qual & WebVid for Object Trajectory \\

  \rule{0pt}{\rowheightmethods} \cite{xu2024camco} & Human-Based & - & FID - FVD - Pose accuracy & WebVid \\ 
  \rule{0pt}{\rowheightmethods} & & & COLMAP error rate  \\

  \rule{0pt}{\rowheightmethods} \cite{kuang2024collaborative} & Human-Based & - & FID - KID - CLIP-T - CLIP-F & WebVid10M, RealEstate10K \\ 
  \rule{0pt}{\rowheightmethods} & & & Rotation AUC - Transition AUC \\
  \rule{0pt}{\rowheightmethods} & & & Qual \\
  
  \rule{0pt}{\rowheightmethods} \cite{jawad2024deep} & - & - & ROS Latency & Published in \\
  \rule{0pt}{\rowheightmethods} & & & Base Prediction Time & \cite{eslamian2020development}  \\
  
  \rule{0pt}{\rowheightmethods} \cite{hou2024training} & Human-Based & - & FVD - FID - IS - ATE & Not Public \\ 
  \rule{0pt}{\rowheightmethods} & & & CLIP-SIM - RPE-T - RPE-R &  \\

  \rule{0pt}{\rowheightmethods} \cite{bar2024navigationworldmodels} & Human-Based & - & FVD - FID- PSNR - DreamSim & SCAND - TartanDrive - RECON  \\
  \rule{0pt}{\rowheightmethods} & & &  LPIPS - RPE - ATE & HuRoN- Ego4DitHub \\ 



  \bottomrule
\end{tabular}
}
\end{center}
\bigskip\centering
\footnotesize\emph{Note:} All the entries are entered based on evidence or our evaluation. (Qual = Qualitative)
\end{table*}%

\subsection{Hybrid}
Many problems in camera trajectory generation are approached by integrating multiple methods or combining different strategies to achieve better results. These approaches, often referred to as hybrid methods, leverage a mix of concepts and assumptions to optimize performance \cite{liu2024splatraj}. While some hybrid methods directly generate camera trajectories by producing a sequence of coordinates to position the camera in space, others take an indirect approach \cite{hu2024motionmaster, kirillov2023segment}. In the indirect case, the method does not output the trajectory itself \cite{azzarelli2024reviewing}, but instead generates products related to the trajectory or derived from. This section reviews various proposed methods that utilize a combination of techniques to either directly or indirectly generate camera trajectories within camera control systems.

The first notable approach was proposed by Bares et al. \cite{bares2000virtual} that introduced an environment for creating storyboard frames, known as the storyboard frame editor interface. The objective of model is to position the camera in a virtual 3D environment to realize the storyboard frame. This work does not explicitly deal with linguistic descriptions of the constraints; instead, the constraints are implicitly represented in the storyboard frames. 

A hybrid method for adaptive virtual camera control in computer games is presented in \cite{burelli2011towards}, aiming to enhance player experience by automatically adjusting the camera based on real-time gameplay conditions. This hybrid approach combines rule-based and machine learning techniques, inspired by gaze data collection methods \cite{bernhard2010empirical} but adapted to model the interplay between camera behavior, gameplay characteristics, and player actions. The process involves two steps: first, k-means clustering is used to group gaze-based data into distinct camera behaviors, iteratively adjusting clusters based on validity measures. Second, neural networks predict appropriate camera behaviors for different game areas, enabling nuanced and adaptive camera control tailored to player actions.

This study was later improved in \cite{Burelli2015} by replacing SVR and RF learning methods in \cite{burelli2011towards} with neural networks to model the relationship between player and camera behaviors more effectively. This advancement focused on predicting suitable camera profiles for future game segments, further enhancing the system's adaptability.

In subsequent work, a comprehensive survey on game cinematography systems was conducted \cite{burelli2016game}, addressing the design principles and methods for developing cinematic virtual camera control systems.

\begin{figure}[h]
    \centering
        \includegraphics[width=0.4\textwidth]{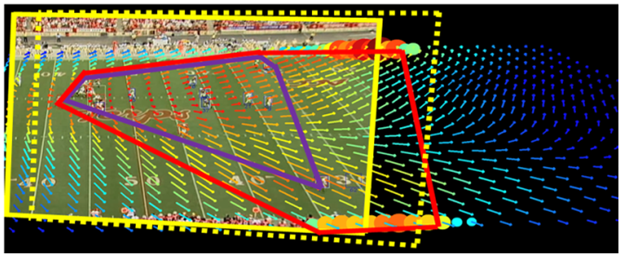}
    \caption{The convex hull formed by the player locations and merging points (red lines) indicates the field of view determined by GPR. \cite{kim2012detecting}.}
    \label{fig:kimcomparison}
\end{figure}

Kim et al.~\cite{kim2012detecting} proposes a method to detect regions of interest (ROIs) in dynamic scenes with PTZ cameras \ref{PTZ}, such as sports videos, addressing inefficiencies of prior Radial Basis Function (RBF) methods \cite{kim2010motion}. By using Gaussian Process Regression (GPR) \cite{kim2011gaussian}, the method constructs a stochastic motion field to capture global motion tendencies and filter low-certainty regions, improving robustness and efficiency. As illustrated in Figure \ref{fig:kimcomparison}, the GPR-based approach aligns predicted ROIs with actual camera movements more effectively, reducing computational overhead while requiring hyper-parameter tuning for optimal performance.

The method in \cite{chen2015mimicking} predicts the pan angle of a PTZ camera \ref{PTZ} based on player tracking data from basketball games, aiming to replicate human camera operator decisions. It combines multiple regression techniques—linear least squares \cite{bjorck1990least}, support vector regression \cite{smola2004tutorial}, and random forest regression \cite{biau2016random}—with feature vectors derived from player positions, heat maps, and spherical maps \cite{chen2015mimicking}. These inputs enable the learning algorithms to accurately predict camera movements, ensuring effective tracking of dynamic scenes.

\begin{figure}[h]
    \centering
        \includegraphics[width=0.4\textwidth]{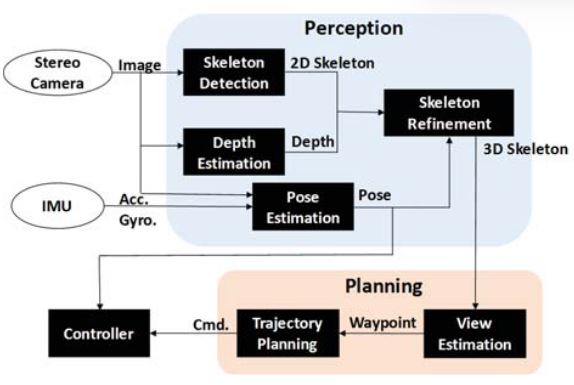}
    \caption{Overview of ACT system for cinematography \cite{huang2018act}.}
    \label{fig:huang2018act}
\end{figure}

An autonomous drone cinematography system is proposed in \cite{huang2018act}, designed to generate camera trajectories for action scenes by dynamically tracking human subjects. As shown in Figure \ref{fig:huang2018act}, the system detects 2D skeleton keypoints using stereo cameras and OpenPose \cite{cao2017realtime}, refining 3D poses with polynomial regression \cite{heiberger2009polynomial} for temporal consistency and smoothness. Camera viewpoints are selected based on predicted poses, and trajectories are optimized using polynomial functions while adhering to drone constraints such as velocity, acceleration, and safety distances. Real-time re-evaluation ensures continuous, feasible motion that integrates aesthetic and physical constraints.

\begin{figure}[h]
    \centering
        \includegraphics[width=0.48\textwidth, height=4.6cm]{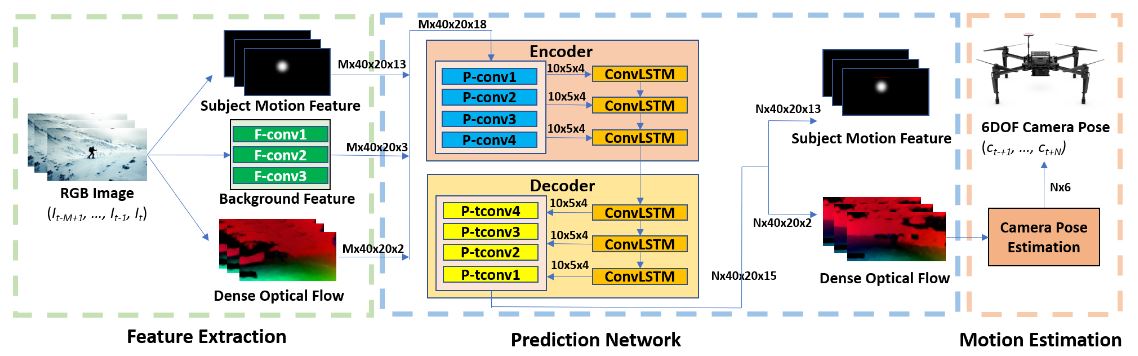}
    \caption{Imitation learning framework featuring three key modules \cite{huang2019learning}.}
    \label{fig:huang2019}
\end{figure}

An autonomous drone cinematography system capable of generating camera trajectories for action scenes by imitating human filming techniques is introduced in \cite{huang2019learning}. As shown in Figure \ref{fig:huang2019}, the framework consists of three modules: feature extraction, prediction network, and camera motion estimation. Features such as subject optical flow, background information, and prior camera motions are extracted from video frames. A Seq2Seq ConvLSTM network \cite{chen2015convolutional} predicts future camera and subject motions using these features. The predicted optical flow is then used to estimate real-time camera motion, ensuring smooth subject tracking and appropriate composition throughout filming.

\begin{figure}[h]
    \centering
        \includegraphics[width=0.48\textwidth, height=3.6cm]{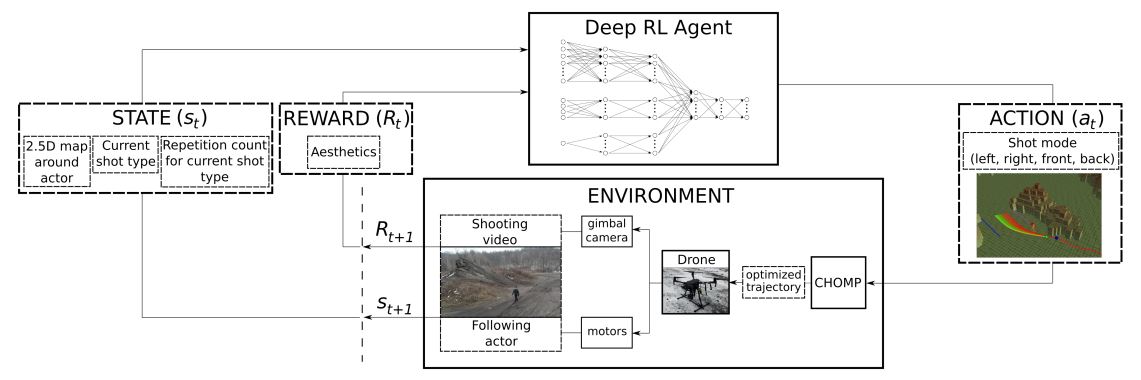}
    \caption{Overall System Flow of \cite{gschwindt2019can}.}
    \label{fig:gs2019}
\end{figure}

\blue{The \cite{gschwindt2019can} addresses automating drone camera trajectory generation for aesthetic aerial cinematography by replacing human input with a deep reinforcement learning (RL) agent. The agent uses a state representation (2.5D height maps, shot type, and repetition count) to select shot modes (e.g., left, right, front, back) and optimizes for rewards based on shot angle, actor presence, shot duration, and collision avoidance. Training combines hand-crafted and human-driven rewards in Microsoft AirSim simulations, generalizing to real-world tests. \ref{fig:gs2019} illustrates the RL framework, where the agent learns to generate smooth and visually pleasing trajectories autonomously.}

In the next work \cite{bonatti2021batteries} an intuitive interface is developed for controlling aerial cinematography by learning a semantic control space. The approach begins by generating diverse video clips based on minimal shot parameters, such as distance and tilt angle, which are then rated by participants to derive semantic descriptors. These descriptors form a reduced semantic space, enabling users to control the robot’s camera motion intuitively during deployment. By manipulating these high-level descriptors, users achieve natural camera control while maintaining a strong link between camera movements and the emotional content of the shot.


The approach in \cite{burg2021real} addresses real-time cinematic tracking in dynamic environments, focusing on generating smooth camera animations that follow a target's motion while avoiding occlusions and collisions. It anticipates the target's behavior using a simulated motion curve and selects a goal camera viewpoint based on predicted positions and prioritized viewpoints. Candidate trajectories are then generated and evaluated for smoothness, continuity, and collision avoidance. The method dynamically adjusts camera paths based on scene geometry, ensuring real-time adaptability and cinematic quality.


The methodology further improved in \cite{burg2022real} by incorporating physics-based simulations to model the target's behavior and predicting future positions and Additionally, leveraging GPU-based computations for efficient ray casting and collision detection, significantly speeding up the evaluation of camera animations.

A camera control system capable of making cinematographic decisions by learning from movie data is proposed in \cite{litteneker2022towards}. The system tackles the challenge of matching virtual camera movements to dynamic scenes with multiple actors by balancing factors like positions, angles, and relative motion to ensure aesthetically pleasing shot composition. Machine learning models are employed to learn a distance metric quantifying the similarity between desired intent and potential compositions. Optimization techniques then determine the optimal camera positions to achieve the user's cinematographic goals, even under complex scene dynamics.

\begin{figure}[h]
    \centering
        \includegraphics[width=0.48\textwidth, height=4.6cm]{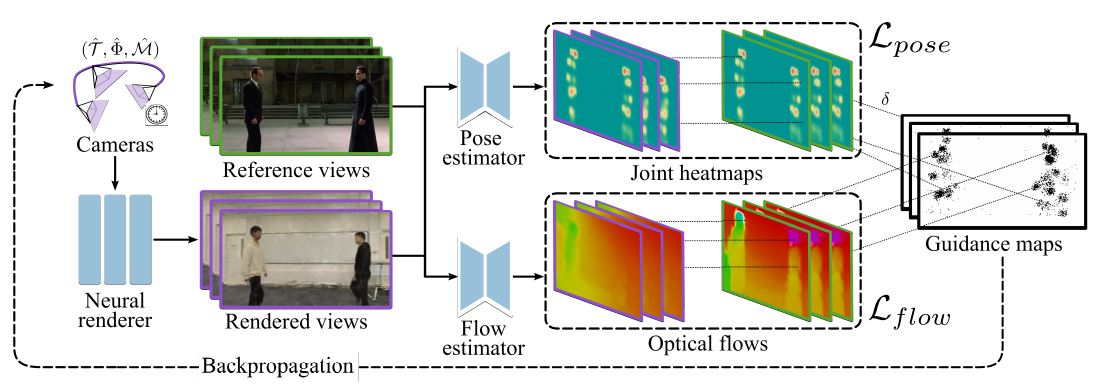}
    \caption{Overview of JAWS pipeline \cite{wang2023jaws}.}
    \label{fig:jaws}
\end{figure}

The method in \cite{wang2023jaws} transfers cinematic features such as camera motion, focal length, and timing from a reference video to a newly generated one. As shown in Figure \ref{fig:jaws}, it optimizes extrinsic and intrinsic camera parameters using the differentiability of neural representations through the Neural Radiance Fields (NeRF) network \cite{zhu2023deep, lin2024dynamic}. By refining cinematic features via backpropagation with guidance maps and optical flows, the approach ensures the generated video closely matches the visual style and motion characteristics of the reference clip.

\begin{figure}[h]
    \centering
        \includegraphics[width=0.5\textwidth, height=3.0cm]{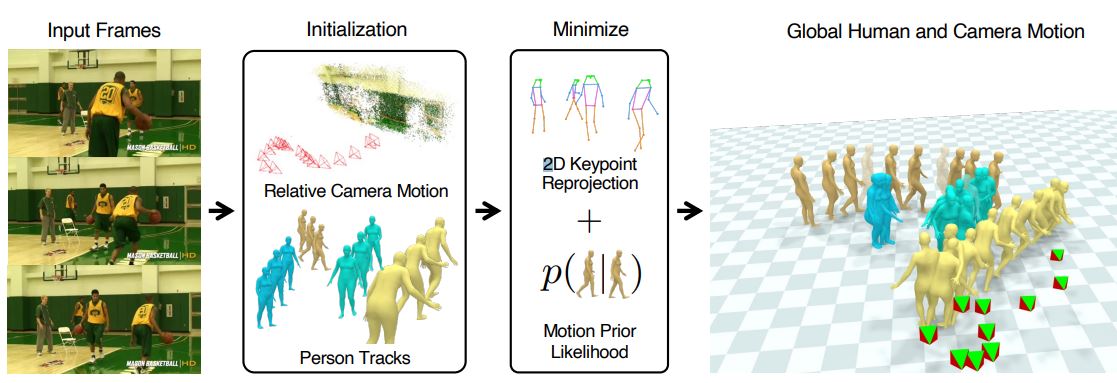}
    \caption{SLAHMR Framework \cite{ye2023decoupling}.}
    \label{fig:slah}
\end{figure}

\blue{The \cite{ye2023decoupling} addresses the task of reconstructing global human trajectories in a shared world frame from in-the-wild videos by decoupling human and camera motion. The proposed method, SLAHMR, estimates relative camera motion using SLAM and initializes human and camera trajectories through 3D human tracking. It then optimizes these trajectories by leveraging 2D video observations and learned human motion priors, aligning camera displacement with plausible human motion to resolve scene scale ambiguity. The process, depicted in \ref{fig:slah}, enables 4D trajectory recovery even in challenging, multi-person scenarios.}

\begin{figure}[h]
    \centering
        \includegraphics[width=0.48\textwidth, height=4.0cm]{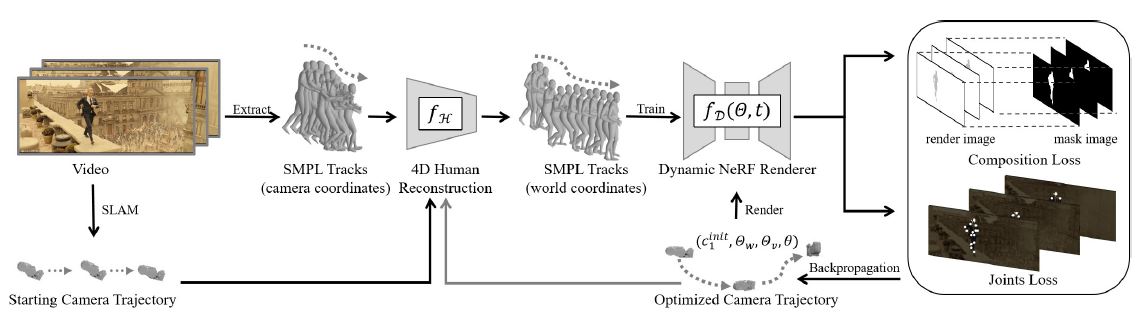}
    \caption{Overview of the approach in \cite{jiang2024cinematic}.}
    \label{fig:cinematicnerf}
\end{figure}

The approach in \cite{jiang2024cinematic} tackles the challenges of estimating camera trajectories and character motion in complex dynamic scenes, particularly where traditional methods like SLAM \cite{durrant2006simultaneous} struggle with dynamic elements and 3D representations. As shown in Figure \ref{fig:cinematicnerf}, the method employs NeRF and pose estimation \cite{zheng2023deep} as a differentiable renderer to estimate camera trajectories and character motion. It refines character motion using the Skinned Multi-Person Linear (SMPL) \cite{10.1145/3596711.3596800} human body model, effectively integrating neural rendering with motion tracking techniques for precise 3D results.

The method in \cite{hu2024motionmaster} addresses the challenge of efficient camera motion control in video generation, reducing the need for extensive training and computational resources. It employs a one-shot camera motion disentanglement technique to separate camera motion from object motion in a source video. The disentangled camera motion is then transferred to a new video, enabling flexible and resource-efficient camera control without the need for complex temporal camera module training.

\begin{figure}[h]
    \centering
        \includegraphics[width=0.48\textwidth, height=6.5cm]{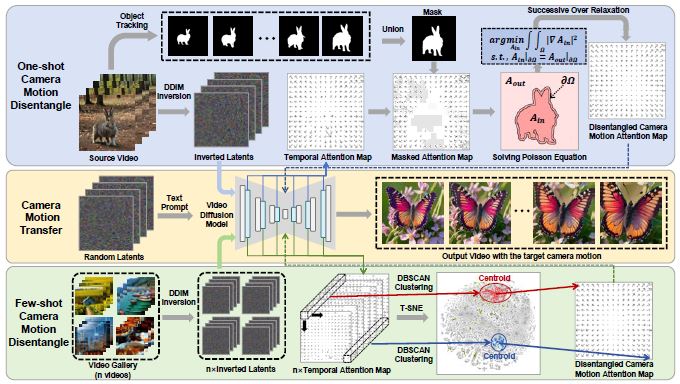}
    \caption{Main framework of \cite{hu2024motionmaster} method \cite{hu2024motionmaster}.}
    \label{fig:motiongen}
\end{figure}

The proposed model is designed to extract camera motion from either a single video or multiple videos with similar camera motions. This process is illustrated in Figure \ref{fig:motiongen}. First) One-shot camera motion disentanglement: The method begins by employing SAM \cite{kirillov2023segment} to segment moving objects in the source video and extract temporal attention maps from inverted latents. To separate camera motion from object motion, object regions in the attention map are masked, and camera motion within the mask is estimated by solving a Poisson equation. Second) Few-shot camera motion disentanglement: In cases involving multiple videos, the model extracts common camera motion from temporal attention maps across the given videos. For each position (x, y), k-neighboring attention map values across videos are clustered, and the centroid of the largest cluster is used to represent the camera motion at that position.

The SplaTraj framework, introduced in \cite{liu2024splatraj}, generates photogenic camera trajectories within environments represented by Gaussian Splatting models. It formulates the task as a trajectory optimization problem guided by user-specified semantic instructions. By integrating rendering-based costs such as target centering and ratio error, the method achieves smooth, object-centered views. Empirical evaluations highlight improvements in object placement, trajectory smoothness, and occlusion avoidance, advancing semantic-driven video generation within photorealistic environments.


Hybrid methods in camera trajectory generation offer several advantages by integrating multiple approaches, allowing for greater flexibility and efficiency in solving complex problems. These methods combine different techniques, such as machine learning, optimization, and neural rendering, to tackle challenges like dynamic scene tracking, real-time adaptation, and generating natural camera movements. However, hybrid methods also come with challenges, such as the need for high computational resources, complex parameter tuning, and the integration of diverse techniques that may not always align seamlessly. Despite these obstacles, the field of hybrid camera trajectory generation is still an area of active research, with significant potential for further improvements. As technologies like Neural Radiance Fields and DL continue to evolve, new opportunities for hybrid methods to enhance camera control systems in dynamic environments are emerging.

Hybrid methods combine rule-based, optimization, and machine learning techniques to achieve greater flexibility and efficiency in solving complex trajectory generation problems. These approaches address challenges like dynamic scene tracking and real-time adaptation, leveraging strengths across methodologies. Table \ref{tab:HCTG} illustrates various hybrid strategies, including direct trajectory generation and indirect methods.

\begin{table*}%
\caption{Overview of Hybrid Methods for Camera Trajectory Generation Methods}
\label{tab:HCTG}
\begin{center}
\begin{tabular}{lccc}
  \toprule
  Method & \makecell[c]{Real World} & Virtual & \makecell[c]{Camera Movement} \\ \midrule

  \rule{0pt}{\rowheightmethods} \cite{burelli2011towards} & - & Game & Fixed \\

  \rule{0pt}{\rowheightmethods} \cite{Burelli2015} & - & Game & Fixed \\

  \rule{0pt}{\rowheightmethods} \cite{kim2012detecting} & Human-Based & - & PTZ \\

  \rule{0pt}{\rowheightmethods} \cite{chen2015mimicking} & Human-Based & - & PTZ \\

  \rule{0pt}{\rowheightmethods} \cite{burelli2016game} & - & Game & - \\

  \rule{0pt}{\rowheightmethods} \cite{huang2018act} & Areal-Based & - & Gimbal Mounted \\

  \rule{0pt}{\rowheightmethods} \cite{huang2019learning} & Areal-Based & - & Gimbal Mounted \\
  
  \rule{0pt}{\rowheightmethods} \cite{gschwindt2019can} & Areal-Based & - & Gimbal Mounted \\

  \rule{0pt}{\rowheightmethods} \cite{bonatti2021batteries} & Areal-Based & - & Gimbal Mounted \\

  \rule{0pt}{\rowheightmethods} \cite{burg2021real} & Areal-Based & - & Gimbal Mounted \\

  \rule{0pt}{\rowheightmethods} \cite{burg2022real} & Areal-Based & Animation/Games & Gimbal Mounted \\

  \rule{0pt}{\rowheightmethods} \cite{litteneker2022towards} & - & Animation/Games & Non-Fixed \\

  \rule{0pt}{\rowheightmethods} \cite{wang2023jaws} & Human-Based & - & - \\
  
  \rule{0pt}{\rowheightmethods} \cite{ye2023decoupling} & Human-Based & - & - \\

  \rule{0pt}{\rowheightmethods} \cite{jiang2024cinematic} & Human-Based & - & - \\

  \rule{0pt}{\rowheightmethods} \cite{hu2024motionmaster} & Human-Based & - & - \\
  
  \rule{0pt}{\rowheightmethods} \cite{liu2024splatraj} & Human-Based & - & - \\

  \bottomrule
\end{tabular}
\end{center}
\bigskip\centering
\footnotesize\emph{Note:} All the entries are entered based on evidence or our evaluation.
\end{table*}%

\section{Metrics} \label{metric_section}
After gaining a thorough understanding of camera trajectory generation methods, it becomes necessary to evaluate their performance in order to assess the effectiveness of the underlying approaches. This evaluation relies on a comprehensive set of metrics that account for all relevant aspects of the camera trajectory. Metrics play a crucial role in this process by providing objective and reproducible standards for assessing the quality and functionality of generated trajectories. The methods employed for camera trajectory evaluation can be classified into general and specific metrics. Since a camera trajectory defines the path and orientation a camera follows through a scene, it significantly influences how visual narratives are communicated and perceived. Without standardized metrics, comparisons between different trajectory generation methods would remain inconsistent and inherently subjective.

Camera trajectory generation shares similarities with sequence analysis, as it involves evaluating temporal dependencies and continuity, akin to time series analysis. Techniques such as statistical correlation \cite{unterthiner2018towards, heusel2017gans} and predictive modeling \cite{yang2024direct, radford2021learning} can be adapted to assess trends and coherence in the generated trajectories, ensuring spatial consistency and enhancing audience engagement. These techniques can be considered as general metrics.

However, beyond these general methods of sequence analysis, comprehensive evaluation of camera trajectories requires domain-specific criteria \cite{courant2025exceptional}. The need for specialized metrics arises from the inherently multifaceted nature of these trajectories, which are influenced by various factors \cite{muller2007dynamic}. This necessity stems from the fact that camera trajectories are shaped by diverse aspects, including cinematic principles, temporal characteristics, interactions between scene components, and user prompts \cite{naeem2020reliable}. Consequently, there is a need for metrics capable of adequately addressing these complexities.

Despite the significant efforts devoted to developing purpose-specific metrics for evaluating particular aspects of camera trajectories, there remains a notable absence of general-purpose metrics capable of assessing all aspects of a camera trajectory comprehensively. As a result, qualitative evaluation methods continue to play a substantial role in this field.

The rest of this section is dedicated to quantitative and qualitative assessments. Quantitative metrics involve numerical evaluations, such as trajectory smoothness measured by minimizing jerk \cite{galvane2018directing} or acceleration variance \cite{nageli2017real}. Qualitative metrics, conversely, assess subjective aspects like the emotional impact \cite{bonatti2021batteries} of a trajectory or its alignment with storytelling goals \cite{wu2018thinking}. 

\subsection{Quantitative Metrics}

\subsubsection{\textbf{Peak Signal-to-Noise Ratio} \cite{korhonen2012peak, moreno2013towards}} \label{PSNR} \hfill

\noindent Peak Signal-to-Noise Ratio (PSNR) quantifies image or video quality by comparing a reconstructed version to the original. It expresses the maximum possible signal power relative to noise in logarithmic decibels (dB), with higher values indicating better quality.

\begin{equation}
\text{PSNR} = 10 \cdot \log_{10} \left( \frac{\text{MAX}^2}{\text{MSE}} \right)
\end{equation}

\begin{equation}
\text{MSE} = \frac{1}{n} \sum_{i=1}^{n} \left( x_i - \hat{x}_i \right)^2
\end{equation}
Where \(\text{MAX}\) is the maximum possible pixel value.

\subsubsection{\textbf{Structural Similarity Index} \cite{brunet2011mathematical}} \label{SSIM} \hfill

\noindent The Structural Similarity Index (SSIM) is a perceptual metric used to evaluate the similarity between two images. It assesses image quality based on structural information, luminance, and contrast, making it more aligned with human visual perception than traditional metrics like mean squared error.

The formula for SSIM is given by:
\begin{equation}
\text{SSIM}(x, y) = \frac{(2 \mu_x \mu_y + C_1)(2 \sigma_{xy} + C_2)}{(\mu_x^2 + \mu_y^2 + C_1)(\sigma_x^2 + \sigma_y^2 + C_2)}
\end{equation}

Where:
\begin{itemize}
    \item \( \mu_x \): Mean of image \( x \).
    \item \( \mu_y \): Mean of image \( y \).
    \item \( \sigma_x^2 \): Variance of image \( x \).
    \item \( \sigma_y^2 \): Variance of image \( y \).
    \item \( \sigma_{xy} \): Covariance between images \( x \) and \( y \).
    \item \( C_1 \) and \( C_2 \): Small constants to stabilize the division when the denominator is close to zero.
\end{itemize}

\subsubsection{\textbf{Dynamic Time Wrapping} \cite{muller2007dynamic, senin2008dynamic}} \label{q_metric_8} \hfill

\noindent Dynamic Time Warping (DTW) is a widely used algorithm for measuring the similarity between two temporal sequences that may vary in time or speed. Unlike simple distance metrics such as the Euclidean distance, DTW can handle time-series sequences that are misaligned due to temporal distortions. The core idea is to find an optimal alignment between two sequences by allowing non-linear mapping of time indices while minimizing a cumulative distance.

Given two time series \(X = \{x_1, x_2, \dots, x_N\}\) and \(Y = \{y_1, y_2, \dots, y_M\}\), where \(x_i, y_j \in \mathbb{R}\), the DTW distance is computed by constructing an \(N \times M\) cost matrix \(D\) and finding the warping path \(P = \{(i_1, j_1), (i_2, j_2), \dots, (i_L, j_L)\}\) that minimizes the cumulative cost. The cost matrix \(D\) is defined as:
\begin{equation}
    D(i, j) = \|x_i - y_j\|^2,
\end{equation}

where \(D(i, j)\) measures the squared distance between the elements \(x_i\) and \(y_j\).

The warping path \(P\) satisfies the following constraints:
\begin{enumerate}
    \item \textbf{Boundary Condition}: \(P(1) = (1, 1)\) and \(P(L) = (N, M)\).
    \item \textbf{Continuity}: If \(P(k) = (i, j)\), then \(P(k+1) \in \{(i+1, j), (i, j+1), (i+1, j+1)\}\).
    \item \textbf{Monotonicity}: The indices \(i\) and \(j\) in \(P\) must be non-decreasing.
\end{enumerate}

The objective of DTW is to minimize the cumulative cost over all valid warping paths:
\begin{equation}
    \text{DTW}(X, Y) = \min_P \sum_{(i, j) \in P} D(i, j).
\end{equation}

The optimal warping path is typically found using dynamic programming. The recurrence relation for the cumulative cost matrix \(C\) is given as:
\begin{equation}
    C(i, j) = D(i, j) + \min \{C(i-1, j), C(i, j-1), C(i-1, j-1)\},
\end{equation}

where \(C(i, j)\) represents the cumulative cost up to point \((i, j)\). The final DTW distance is then:
\begin{equation}
\text{DTW}(X, Y) = \sqrt{C(N, M)}.
\end{equation}

\begin{itemize}
    \item \(X, Y\): Input time-series sequences of lengths \(N\) and \(M\), respectively.
    \item \(D(i, j)\): Local cost between elements \(x_i\) and \(y_j\).
    \item \(C(i, j)\): Cumulative cost matrix.
    \item \(P\): Optimal warping path.
\end{itemize}

\subsubsection{\textbf{CLIP-Score} \cite{radford2021learning}} \label{q_metric_5} 
\hfill

\noindent CLIP-Score (\text{CLIP-S}) is a reference-free evaluation metric designed for assessing image-caption compatibility by leveraging the representations learned by the pre-trained CLIP model. Unlike traditional metrics that rely on comparisons between machine-generated captions and multiple human-authored references, CLIP-Score uses only the image and its candidate caption, aligning closely with how humans evaluate captions. It is computed as:

\begin{equation}
\text{CLIP-S}(c, v) = w \cdot \max\left(\cos\left(c, v\right), 0\right),
\end{equation}

where:
\begin{itemize}
    \item $c, v$: Normalized embeddings of the candidate caption and the image, respectively.
    \item $\cos(c, v)$: Cosine similarity between the embeddings.
    \item $w$: A rescaling factor, typically $w = 2.5$.
\end{itemize}

\subsubsection{\textbf{NIQE} \cite{mittal2012making}} \label{q_metric_6}

\hfill

\noindent The Natural Image Quality Evaluator (NIQE) is a no-reference image quality assessment metric. It operates in a completely blind manner, meaning it does not require any prior knowledge of distorted images or human opinion scores. Instead, NIQE uses Natural Scene Statistics (NSS) extracted from undistorted natural images to evaluate the quality of a given image. This approach makes NIQE distortion-agnostic and "opinion-unaware," relying solely on measurable deviations from the statistical regularities of natural images. NIQE evaluates the perceptual quality of frames within trajectories, identifying any unnatural distortions in the generated sequences. This ensures a realistic visual appeal for camera-generated sequences.

NIQE evaluates image quality based on the multivariate Gaussian (MVG) model and it is described as follows:
    
\begin{enumerate}
    \item \textbf{Preprocessing:} Local mean removal and divisive normalization are applied:
    \begin{equation}
    \hat{I}(i, j) = \frac{I(i, j) - \mu(i, j)}{\sigma(i, j) + 1},
    \end{equation}
    
    where $\mu(i, j)$ and $\sigma(i, j)$ are the local mean and standard deviation, respectively.
    
    \item \textbf{NSS Feature Extraction:} NSS features, including parameters of generalized Gaussian distributions (GGD) and asymmetric generalized Gaussian distributions (AGGD), are computed from patches.
    
    \item \textbf{Multivariate Gaussian Model:} A multivariate Gaussian model is fitted to the NSS features:
    \begin{equation}
    f_X(x_1, \ldots, x_k) = \sqrt{\frac{1}{(2\pi)^k \sqrt{|\Sigma|}} \exp \left( -\frac{1}{2} (x - \nu)^T \Sigma^{-1} (x - \nu) \right)},
    \end{equation}
    
    where $\nu$ and $\Sigma$ are the mean vector and covariance matrix of the pristine natural image corpus.
    
    \item \textbf{Quality Assessment:} The quality of a distorted image is expressed as the Mahalanobis distance:
    \begin{equation}
    D(\nu_1, \nu_2, \Sigma_1, \Sigma_2) = \sqrt{ (\nu_1 - \nu_2)^T \left( \frac{\Sigma_1 + \Sigma_2}{2} \right)^{-1} (\nu_1 - \nu_2) }.
    \end{equation}
\end{enumerate}

\subsubsection{\textbf{BRISQUE} \cite{mittal2012no}} \label{q_metric_7} \hfill

\noindent The Blind/Referenceless Image Spatial Quality Evaluator (BRISQUE) is a no-reference image quality assessment metric that quantifies perceptual quality by analyzing deviations from NSS in the spatial domain. Unlike distortion-specific approaches, BRISQUE leverages a distortion-generic framework using locally normalized luminance coefficients.

The locally normalized luminance coefficients, \( \hat{I}(i, j) \), are defined as:
\begin{equation}
    \hat{I}(i, j) = \frac{I(i, j) - \mu(i, j)}{\sigma(i, j) + C},
\end{equation}

where
\begin{equation}
    \mu(i, j) = \sum_{k=-K}^{K} \sum_{l=-L}^{L} w_{k, l} I(i+k, j+l),
\end{equation}

\begin{equation}
    \sigma(i, j) = \sqrt{\sum_{k=-K}^{K} \sum_{l=-L}^{L} w_{k, l} \left(I(i+k, j+l) - \mu(i, j)\right)^2}.
\end{equation}
The coefficients \( \hat{I}(i, j) \) are modeled using a Generalized Gaussian Distribution (GGD):
\begin{equation}
f(x; \alpha, \sigma^2) = \frac{\alpha}{2 \beta \Gamma(1/\alpha)} \exp\left(-\left(\frac{|x|}{\beta}\right)^\alpha\right),
\end{equation}

where \( \beta = \sigma \sqrt{\Gamma(1/\alpha)/\Gamma(3/\alpha)} \).

BRISQUE also models paired product coefficients along four orientations: horizontal, vertical, main diagonal, and secondary diagonal, using an Asymmetric Generalized Gaussian Distribution (AGGD).

\subsubsection{\textbf{Flow Error} \cite{yang2024direct}} \label{q_metric_9} \hfill

\noindent The \textit{Flow Error Metric} is designed to evaluate the quality of camera movement control in video generation. It quantifies the deviation between the optical flow from generated videos and the ground truth flow derived from specified camera movement parameters. Optical flow represents the motion of objects or the camera between consecutive frames, making this metric essential for assessing temporal dynamics and movement consistency.

This metric utilizes VideoFlow \cite{shi2023videoflow}, an optical flow estimation model, to extract flow maps from generated videos. The extracted flow maps are compared against the ground truth flow maps, which are computed based on the given camera movement parameters. The Flow Error Metric is defined as:
\begin{equation}
    \text{Flow Error} = \frac{1}{N} \sum_{(x, y, t)} \| \mathbf{F}_g(x, y, t) - \mathbf{F}_r(x, y, t) \|_2,
\end{equation}
where:
\begin{itemize}
\item \( \mathbf{F}_g(x, y, t) \) represent the optical flow at spatial location \((x, y)\) and time \(t\) in the generated video

\item \( N \) is the total number of flow vectors (pixels over all frames).

\item \( \mathbf{F}_r(x, y, t) \) denote the ground truth optical flow derived from camera movement parameters
\end{itemize}

\subsubsection{\textbf{Average Precision} \cite{zhu2004recall}} \label{q_metric_11} \hfill

\noindent The \textit{Average Precision (AP)} is a general-propose metric which evaluates the precision-recall trade-off across confidence thresholds, commonly used in object detection and classification tasks. It represents the area under the precision-recall curve.

Let \( \text{Precision}(r) \) be the precision at recall \( r \). The AP is defined as:
\begin{equation}
    \text{AP} = \int_0^1 \text{Precision}(r) \, dr,
\end{equation}

where the integral is approximated numerically by summing over discrete recall levels. Precision and recall are defined as:
\begin{equation}
    \text{Precision} = \frac{\text{TP}}{\text{TP} + \text{FP}}, \quad
\text{Recall} = \frac{\text{TP}}{\text{TP} + \text{FN}},
\end{equation}
with TP, FP, and FN representing true positives, false positives, and false negatives, respectively.

\subsubsection{\textbf{Average Endpoint Error} \cite{sharmin2012optimal}} \label{q_metric_12} \hfill

\noindent The Average Endpoint Error (AEE) metric is a quantitative measure used to evaluate the precision of predicted optical flows. It assesses the deviation of predicted motion vectors from the ground truth, particularly in the context of drone cinematography systems. It quantifies the ability of the system to replicate professional filming styles. Lower AEE values signify higher accuracy in the imitation of expert cinematography \cite{galvane2015camera}.

The AEE is mathematically defined as:
\begin{equation}
    \text{AEE} = \frac{1}{W.H} \sum_{i=1}^{W} \sum_{j=1}^{H} \sqrt{(u_{i,j} - u_{i,j}^{\text{GT}})^2 + (v_{i,j} - v_{i,j}^{\text{GT}})^2},
\end{equation}

where:
\begin{itemize}
    \item $W$ and $H$ are the width and height of the optical flow map, respectively.
    \item $(u, v)$ and $(u^{\text{GT}}, v^{\text{GT}})$ are the predicted and ground-truth optical flow components, respectively.
    \item $N$: Total number of pixels in the optical flow map.
\end{itemize}

\subsubsection{\textbf{Precision}\cite{naeem2020reliable}} \label{q_metric_16} \hfill

\noindent Precision quantifies the fidelity of the generated data by measuring the proportion of generated samples that lie within the manifold of real data. It evaluates how realistic the generated samples are with respect to the real data distribution, ensuring that the generative model does not produce artifacts or unrealistic outputs. The manifold of real data is constructed by creating \(k\)-nearest neighbor \cite{cover1967nearest} spheres centered at each real data point. These spheres capture the density and locality of real data points in the feature space. In camera domain, it ensures that the generated trajectory closely match the fidelity of real-world trajectories. It helps to verify that the model does not produce unrealistic or physically infeasible trajectories.

\begin{equation}
\text{Precision} = \frac{1}{M} \sum_{j=1}^M \mathbf{1}_{Y_j \in \text{manifold}(X_1, \dots, X_N)}
\end{equation}

Where:  
\begin{itemize}
    \item \(M\): Number of generated samples.
    \item \(N\): Number of real samples.
    \item \(\mathbf{1}_{\cdot}\): Indicator function, returning 1 if the condition inside holds and 0 otherwise.
    \item \(\text{manifold}(X_1, \dots, X_N)\): The union of neighborhood spheres around the real data points.
\end{itemize}

\subsubsection{\textbf{Recall} \cite{naeem2020reliable}}\label{q_metric_17} \hfill

\noindent Recall quantifies the diversity of the generated data by evaluating the proportion of the real data manifold that is covered by the generated samples. This metric ensures that the generative model captures the variability inherent in the real data, avoiding mode collapse and ensuring that diverse samples are represented. The recall metric depends on the ability of generated samples to cover the regions of the real data manifold. The \(k\)-nearest neighbor spheres around generated samples determine whether real samples are sufficiently represented within these spheres. In the context of camera trajectory generation, recall ensures that the generative model produces a diverse set of trajectories that spans the range of possible paths observed in real-world data. This is crucial for applications where diversity in camera movement is essential.

\begin{equation}
\text{Recall} = \frac{1}{N} \sum_{i=1}^N \mathbf{1}_{X_i \in \text{manifold}(Y_1, \dots, Y_M)}
\end{equation}

Where:  
\begin{itemize}
    \item \(N\): Number of real samples.
    \item \(M\): Number of generated samples.
    \item \(\mathbf{1}_{\cdot}\): Indicator function.
    \item \(\text{manifold}(Y_1, \dots, Y_M)\): The union of neighborhood spheres around the generated data points.
\end{itemize}

\subsubsection{\textbf{Density} \cite{naeem2020reliable}} \label{q_metric_18} \hfill

\noindent Density enhances the precision metric by accounting for the relative density of generated samples within the real data manifold. Unlike precision, which evaluates fidelity as a binary outcome, density provides a more nuanced measure by considering how densely generated samples populate the neighborhoods of real data points. The parameter \(k\) controls the granularity of the neighborhood estimation. Density rewards regions where real samples are densely packed and penalizes overestimation due to outliers. In evaluating camera trajectory generation, density measures how well the generated trajectories fill the regions of real trajectories. This provides an indication of both fidelity and coverage of densely populated areas in real trajectory datasets, which is crucial for applications requiring precision and robustness.

\begin{equation}
\text{Density} = \frac{1}{kM} \sum_{j=1}^M \sum_{i=1}^N \mathbf{1}_{Y_j \in B(X_i, \text{NND}_k(X_i))}
\end{equation}

Where:  
\begin{itemize}
    \item \(k\): Number of nearest neighbors considered.
    \item \(M\): Number of generated samples.
    \item \(N\): Number of real samples.
    \item \(B(X_i, \text{NND}_k(X_i))\): Neighborhood sphere centered at \(X_i\), with a radius determined by the distance to its \(k\)-th nearest neighbor (\(\text{NND}_k\)).
\end{itemize}

\subsubsection{\textbf{Coverage} \cite{naeem2020reliable}} \label{q_metric_19} \hfill

\noindent Coverage improves upon the recall metric by focusing on the proportion of real data points that are represented in the neighborhoods of generated samples. Unlike recall, which may overestimate due to outliers, coverage provides a robust measure of diversity by assessing whether each real sample has at least one nearby generated sample. Coverage requires that for each real data point, there exists at least one generated sample within its neighborhood sphere. This metric provides a bounded value between 0 and 1, making it robust to variability in data distributions. Coverage ensures that the generated camera trajectories adequately represent the variability in real trajectories. This guarantees that all important modes in real-world trajectories are captured, avoiding the exclusion of significant patterns.

\begin{equation}
\text{Coverage} = \frac{1}{N} \sum_{i=1}^N \mathbf{1}_{\exists j \text{ such that } Y_j \in B(X_i, \text{NND}_k(X_i))}
\end{equation}

Where:  
\begin{itemize}
    \item \(N\): Number of real samples.
    \item \(M\): Number of generated samples.
    \item \(B(X_i, \text{NND}_k(X_i))\): Neighborhood sphere around \(X_i\), with radius defined by its \(k\)-th nearest neighbor (\(\text{NND}_k\)).
\end{itemize}

\subsubsection{\textbf{Fr\'{e}chet Inception Distance} \cite{heusel2017gans}} \hfill \label{q_metric_1}

\noindent The Fr\'{e}chet Inception Distance (FID) is a metric introduced to evaluate the quality of generative models, particularly Generative Adversarial Networks (GANs) \cite{goodfellow2014generative}, by measuring the similarity between the distributions of generated and real-world data. FID improves upon earlier metrics by comparing the statistical properties of these distributions rather than relying solely on the generated data's diversity and clarity \cite{naeem2020reliable}. Mathematically, FID computes the Wasserstein-2 distance \cite{vaserstein1969markov} between two multivariate Gaussian distributions: one representing the real data and the other representing the generated data. These distributions are derived from the feature embeddings of the data obtained through a pre-trained Inception-v3 network \cite{heusel2017gans}, specifically from its last pooling layer. 
FID measures the similarity between the distribution of real and generated trajectory frames. Applied to camera trajectory evaluation, it assesses how realistic and visually coherent the generated frames are in comparison to ground-truth sequences. The FID is defined as:

\begin{equation}FID(\mathcal{P}_r, \mathcal{P}_g) = \|\mu_r - \mu_g\|_2^2 + \text{Tr}\left(\sqrt{\Sigma_r + \Sigma_g - 2\left(\Sigma_r \Sigma_g\right)}\right)\end{equation}

where:
\begin{itemize}
    \item $\mathcal{P}_r$, $\mathcal{P}_g$ are the real and generated data distributions, respectively, derived from the Inception-v3 network,
    \item $\mu_r, \mu_g$: Mean vectors of the embeddings for the real and generated data, respectively.
    \item $\Sigma_r, \Sigma_g$: Covariance matrices of the embeddings for the real and generated data.
\end{itemize}

\subsubsection{\textbf{Fr\'{e}chet Video Distance} \cite{unterthiner2018towards}} \label{q_metric_2} \hfill

\noindent The \textit{Fr\'{e}chet Video Distance (FVD)} is a metric designed to evaluate the quality of generative video models by measuring the distance between the distribution of real videos and the distribution of videos generated by a model. Introduced in the paper, FVD extends the Fr\'{e}chet Inception Distance \cite{unterthiner2018towards} to account for both spatial and temporal aspects of video data. Unlike frame-level metrics such as PSNR \cite{korhonen2012peak, moreno2013towards} or SSIM \cite{brunet2011mathematical}, FVD evaluates the spatiotemporal consistency of videos. 

Let \( \mathcal{P}_g \) and \( \mathcal{P}_g \) denote the distributions of real and generated videos, respectively. The FVD between these distributions is analogous to the FID, differing only in its parameterization. \( \mu_r \) and \( \mu_g \) represent the means of the distributions \( \mathcal{P}_r \) and \( \mathcal{P}_g \), capturing both spatial and temporal characteristics of video data. Similarly, \( \Sigma_r \) and \( \Sigma_g \) denote the covariance matrices of \( \mathcal{P}_r \) and \( \mathcal{P}_g \), respectively, which encode the variability of spatiotemporal features within the real and generated video distributions. This metric assumes that the distributions \( \mathcal{P}_r \) and \( \mathcal{P}_g \) follow a multivariate Gaussian distribution in the chosen feature space. The feature representations are extracted from a pre-trained neural network.

\subsubsection{\textbf{Fr\'{e}chet CLaTr Distance} \cite{courant2025exceptional}} \label{q_metric_3} 

\hfill

\noindent Courant et al. introduced CLaTr (Contrastive Language-Trajectory) embedding which is a robust evaluation metric designed to assess the alignment between textual descriptions and generated camera trajectories. It leverages contrastive learning to enhance the correlation between language and trajectory data, thereby improving the accuracy and reliability of trajectory generation models. The Fr\'{e}chet CLaTr Distance (\text{FDCLaTr}) measures the similarity between the distribution of real and generated camera trajectories in the CLaTr embedding space \cite{courant2025exceptional}.

\subsubsection{\textbf{CLaTr-Score} \cite{courant2025exceptional}} \label{q_metric_4} \hfill

\noindent The CLaTr-Score evaluates the semantic and geometric alignment between a generated camera trajectory and its textual description. It is calculated as:
\begin{equation}
\text{CLaTr-Score} = \frac{T \cdot C}{\|T\| \|C\|},
\end{equation}
where $T, C$ are normalized embeddings of trajectory and text,

\subsubsection{\textbf{Visual Continuity} \cite{galvane2018directing}} \label{q_metric_13} \hfill

\noindent Smoothness in cinematography refers to the continuity and fluidity of camera motion, characterized by gradual changes in position, velocity, and orientation \cite{chen2024dreamcinema}. On the other hand, visual continuity ensures seamless transitions between frames by maintaining consistent framing and avoiding abrupt changes in composition or perspective, thereby preserving aesthetic and narrative coherence. To achieve visual continuity, the camera trajectory is optimized to minimize deviations from desired framing parameters over time, ensuring consistency in on-screen position, size, and orientation of targets. 

The camera must maintain the desired framing of targets, defined by on-screen position $(x_f, y_f)$, target size $s_f$, and orientation $o_f$. The total cost function combines the framing error and transition smoothness:
\begin{align}
E_{\text{total}} = \sum_{i=0}^N \Big[ & \, \alpha_p \big( (x_i - x_f)^2 + (y_i - y_f)^2 \big) \notag \\
& + \alpha_s (s_i - s_f)^2 \notag \\
& + \alpha_o (o_i - o_f)^2 \Big] + \beta \sum_{i=0}^{N-1} \left( \|\dot{x}_{i+1} - \dot{x}_i\| + \|o_{i+1} - o_i\| \right)
\end{align}

where:
\begin{itemize}
    \item $(x_i, y_i)$: Actual on-screen position of the target at frame $i$.
    \item $s_i$: Actual size of the target at frame $i$.
    \item $o_i$: Actual orientation of the target at frame $i$.
    \item $\alpha_p, \alpha_s, \alpha_o$: Weights for position, size, and orientation terms.
    \item $\beta$ is a weight balancing framing error and smooth transitions.
\end{itemize}

\subsubsection{\textbf{Drone-Specific Metrics} \cite{rousseau2018quadcopter, jeon2019online}} \label{q_metric_15} \hfill

\noindent Drone-based systems require specific metrics to evaluate the performance of camera trajectory generation accurately. Ping is utilized to measure communication delay between the drone and control systems, ensuring real-time responsiveness \cite{galvane2018directing, bonatti2020autonomous}. Computation Time is evaluated to determine the latency of trajectory generation algorithms on drone hardware. Energy Efficiency \cite{bonatti2020autonomous} is assessed by analyzing battery consumption in relation to trajectory complexity. Stability Index \cite{galvane2018directing, bonatti2020autonomous} quantifies trajectory smoothness to reduce visual disruptions, while Collision Risk Assessment evaluates the likelihood of trajectory-induced collisions \cite{burg2022real, burg2020real}. These metrics are generally used for drone-specific performance in cinematography.

Table \ref{tab:quantitative} summarizes this section by presenting each metric and its corresponding formula, with general metrics above the camera specific metrics.

\newcommand{\rowheight}{5ex} 
\definecolor{lightblue}{rgb}{0.7, 0.9, 1}
\definecolor{lightpink}{rgb}{1, 0.8, 0.8}

\definecolor{lightgreen}{rgb}{0.8, 1.0, 0.8} 
\definecolor{lightpink}{rgb}{1.0, 0.8, 0.8} 

\begin{table*}%
\caption{Quantitative Metrics}
\centering
\label{tab:quantitative}
\begin{center}
\resizebox{0.9\textwidth}{!}{%
\begin{tabular}{lcll}
  \toprule
  Metric & Trend & Formula & Introduced in\\ \midrule

  \rule{0pt}{\rowheight} \hyperref[PSNR]{Peak Signal-to-Noise Ratio}
  & \cellcolor{lightgreen} $ \uparrow $ & $
\text{PSNR} = 10 \cdot \log_{10} \left( \frac{\text{MAX}^2}{\text{MSE}} \right) $ & \makecell[l]{\cite{korhonen2012peak}}\\

  \rule{0pt}{\rowheight} \hyperref[SSIM]{Structural Similarity Index}
  & \cellcolor{lightgreen} $ \uparrow $ & $
\text{SSIM}(x, y) = \frac{(2 \mu_x \mu_y + C_1)(2 \sigma_{xy} + C_2)}{(\mu_x^2 + \mu_y^2 + C_1)(\sigma_x^2 + \sigma_y^2 + C_2)} $ & \makecell[l]{\cite{brunet2011mathematical}}\\

  \rule{0pt}{\rowheight} \hyperref[q_metric_8]{Dynamic Time Warping}
  & \cellcolor{lightpink} $ \downarrow $   & $ \text{DTW}(X, Y) = \min_P \sum_{(i, j) \in P} D(i, j) $ 
  & \makecell[l]{\cite{muller2007dynamic}} \\

  \rule{0pt}{\rowheight} \hyperref[q_metric_5]{CLIP-Score}
  & \cellcolor{lightgreen} $ \uparrow $ & $ \text{CLIP-S}(c, v) = w \cdot \max(\cos(c, v), 0) $ 
  & \makecell[l]{\cite{radford2021learning}}\\

  \rule{0pt}{\rowheight} \hyperref[q_metric_6]{Natural Image Quality Evaluator}
  & \cellcolor{lightpink} $ \downarrow $  & $ NIQE(\nu_1, \nu_2, \Sigma_1, \Sigma_2) = \sqrt{ (\nu_1 - \nu_2)^T \left( \frac{\Sigma_1 + \Sigma_2}{2} \right)^{-1} (\nu_1 - \nu_2) } $
  & \makecell[l]{\cite{mittal2012making}}\\

  \rule{0pt}{\rowheight} \hyperref[q_metric_7]{\makecell[l]{Blind/Referenceless Image Spatial\\Quality Evaluator}}
  & \cellcolor{lightpink} $ \downarrow $   & $ \hat{I}(i, j) = \frac{I(i, j) - \mu(i, j)}{\sigma(i, j) + C} $
  & \makecell[l]{\cite{mittal2012no}}\\

  \rule{0pt}{\rowheight} \hyperref[q_metric_9]{Flow Error}
   & \cellcolor{lightpink} $ \downarrow $  & $ \text{Flow Error} = \frac{1}{N} \sum_{(x, y, t)} \| \mathbf{F}_g(x, y, t) - \mathbf{F}_r(x, y, t) \|_2 $ 
  & \makecell[l]{\cite{yang2024direct}} \\

  \rule{0pt}{\rowheight} \hyperref[q_metric_11]{Average Precision}
  & \cellcolor{lightgreen} $ \uparrow $ & $ \text{AP} = \int_0^1 \text{Precision}(r) \, dr $ 
  & \makecell[l]{\cite{zhu2004recall}}\\

  \rule{0pt}{\rowheight} \hyperref[q_metric_12]{Average Endpoint Error}
  & \cellcolor{lightpink} $ \downarrow $  & $ \text{AEE} = \frac{1}{N} \sum_{i=1}^{W} \sum_{j=1}^{H} \sqrt{(u_{i,j} - u_{i,j}^{\text{GT}})^2 + (v_{i,j} - v_{i,j}^{\text{GT}})^2} $ 
  & \makecell[l]{\cite{sharmin2012optimal}}\\

  \rule{0pt}{\rowheight} \hyperref[q_metric_16]{Precision}
  & \cellcolor{lightgreen} $ \uparrow $ & $ \text{Precision} = \frac{1}{M} \sum_{j=1}^M \mathbf{1}_{Y_j \in \text{manifold}(X_1, \dots, X_N)}
 $ & \makecell[l]{\cite{naeem2020reliable}}\\

  \rule{0pt}{\rowheight} \hyperref[q_metric_17]{Recall}
  & \cellcolor{lightgreen} $ \uparrow $ & $ \text{Recall} = \frac{1}{N} \sum_{i=1}^N \mathbf{1}_{X_i \in \text{manifold}(Y_1, \dots, Y_M)}
 $ & \makecell[l]{\cite{naeem2020reliable}}\\

  \rule{0pt}{\rowheight} \hyperref[q_metric_18]{Density}
  & \cellcolor{lightgreen} $ \uparrow $ & $ \text{Density} = \frac{1}{kM} \sum_{j=1}^M \sum_{i=1}^N \mathbf{1}_{Y_j \in B(X_i, \text{NND}_k(X_i))}
 $ & \makecell[l]{\cite{naeem2020reliable}}\\

  \rule{0pt}{\rowheight} \hyperref[q_metric_19]{Coverage}
  & \cellcolor{lightgreen} $ \uparrow $ & $
\text{Coverage} = \frac{1}{N} \sum_{i=1}^N \mathbf{1}_{\exists j \text{ such that } Y_j \in B(X_i, \text{NND}_k(X_i))} $ & \makecell[l]{\cite{naeem2020reliable}}\\


  \rule{0pt}{\rowheight} \hyperref[q_metric_1]{
 Fréchet Inception Distance}
  & \cellcolor{lightpink} $ \downarrow $ & $ FID(\mathcal{P}_r, \mathcal{P}_g) = \|\mu_r - \mu_g\|_2^2 + \text{Tr}\left(\sqrt{\Sigma_r + \Sigma_g - 2\left(\Sigma_r \Sigma_g\right)}\right) $ 
  & \makecell[l]{\cite{heusel2017gans}} \\
  
  \rule{0pt}{\rowheight} \hyperref[q_metric_2]{Fréchet Video Distance}
  & \cellcolor{lightpink} $ \downarrow $ & $ FVD(\mathcal{P}_r, \mathcal{P}_g) = \| \mu_r - \mu_g \|^2 + \mathrm{Tr}\left(\sqrt{\Sigma_r + \Sigma_g - 2 \left(\Sigma_r \Sigma_g\right)} \right) $ 
  & \makecell[l]{\cite{unterthiner2018towards}} \\
  
  \rule{0pt}{\rowheight} \hyperref[q_metric_3]{Fr\'{e}chet CLaTr Distance}
  & \cellcolor{lightpink} $ \downarrow $ & $ FDCLaTr(\mathcal{P}_r, \mathcal{P}_g) = \| \mu_r - \mu_g \|^2 + \mathrm{Tr}\left(\sqrt{\Sigma_r + \Sigma_g - 2 \left(\Sigma_r \Sigma_g\right)} \right) $ 
  & \makecell[l]{\cite{courant2025exceptional}} \\
  
  \rule{0pt}{\rowheight} \hyperref[q_metric_4]{CLaTr-Score}
  & \cellcolor{lightgreen} $ \uparrow $  & $ \text{CLaTr-Score} = \frac{T \cdot C}{\|T\| \|C\|} $
  & \makecell[l]{\cite{courant2025exceptional}}\\

  \rule{0pt}{\rowheight} \hyperref[q_metric_13]{Visual Continuity}
  & \cellcolor{lightpink} $ \downarrow $ & $ E_{\text{total}} = E_{\text{framing}} + \beta \sum_{i=0}^{N-1} \left( \|\dot{x}_{i+1} - \dot{x}_i\| + \|o_{i+1} - o_i\| \right)
 $ & \makecell[l]{\cite{galvane2015camera}}\\

  \bottomrule
\end{tabular}
}
\end{center}
\bigskip\centering
\footnotesize \emph{Note:} Each formula is explained in detail within its corresponding section. Metrics above the horizontal line are general, while those below are specific.
\end{table*}%

\subsection{Qualitative Metrics}
Qualitative evaluation of camera trajectory generation methods focuses on subjective assessments that capture the perceptual and aesthetic quality of the generated trajectories. These metrics complement quantitative measures by addressing how well the generated trajectories align with human expectations and professional standards in practical applications. In this field, three primary categories of qualitative metrics are recognized and will be explored in the subsequent subsections:

\subsubsection{Visual Comparison}
By visually comparing the outputs of a method to a baseline, this approach enables evaluators to assess differences in smoothness, framing, and scene coverage \cite{courant2025exceptional}. This straightforward method effectively highlights areas in which the technique demonstrates strengths or weaknesses, particularly in instances where numerical metrics may not adequately capture subtle nuances.

\subsubsection{User Study}
User studies gather subjective opinions by asking participants to rank or choose the most appealing trajectory among results from different methods \cite{wang2024dancecamera3d}. These studies provide insights into general audience preferences, serving as a reliable indicator of how well a method meets end-user expectations.

\subsubsection{Expert Feedback}
Expert feedback involves evaluations from professionals with extensive experience in cinematography \cite{nageli2017real}. Experts assess trajectories against industry standards, focusing on elements like visual storytelling, framing techniques, and aesthetic appeal. Their input is invaluable for refining methods and ensuring high-quality results. 

To summarize this section, Table \ref{tab:qualitative} presents the categories of qualitative metrics along with the papers that utilize the corresponding metrics for evaluation.

\begin{table}%
\caption{Qualitative Metrics}
\label{tab:qualitative}
\begin{minipage}{\columnwidth}
\begin{center}
\begin{tabular}{lc}
  \toprule
  Metric    & Papers\\ \midrule
  Visual Comparison     & \makecell[c]{
  \cite{courant2025exceptional} \\ \cite{li2024director3d} \\ \cite{jiang2024cinematographic} \\ \cite{wang2023jaws} \\ \cite{yang2024direct} \\ \cite{jiang2024cinematic} \\ \cite{wang2024motionctrl} \\ \cite{hu2024motionmaster} \\ \cite{galvane2014narrative} \\ \cite{louarn2018automated} \\ \cite{yoo2021virtual} \\ \cite{kim2012detecting}}\\ \midrule
  User Study            & \makecell[c]{
  \cite{wang2024dancecamera3d} \\ \cite{wu2018thinking} \\ \cite{guo2023animatediff} \\ \cite{bai2024uniedit} \\ \cite{gebhardt2021optimization} \\ \cite{chen2016learning} \\ \cite{burelli2015adaptive} \\ \cite{lino2011director} \\ \cite{liang2012script} \\ \cite{bonatti2021batteries} \\ \cite{wang2024dancecamanimator}
  }\\ \midrule
  Expert Feedback       & \makecell[c]{
  \cite{nageli2017real} \\ \cite{galvane2018directing}
  }\\
  \bottomrule
\end{tabular}
\end{center}
\bigskip\centering
\end{minipage}
\end{table}%

\section{Datasets} \label{datasets_section}
A significant challenge in camera trajectory generation using deep learning models is the accessibility of high-quality, application-specific datasets. Such datasets are essential for training models that can generalize across diverse environments and scenarios, ensuring robustness and reliability. In this section, we explore the types of datasets used in this field, focusing on their strengths and limitations.

\subsection{Synthetic Datasets}
Obtaining low-level camera parameters, such as focal length, aperture, and sensor size, along with accurate trajectory data, can be difficult and time-consuming. Beside that, real-world datasets often suffer from imbalances \cite{courant2025exceptional}, where certain types of camera movements or scene complexities are underrepresented, leading to biased models that may not generalize well to diverse real-world scenarios. To address these limitations, researchers have increasingly turned to synthetic datasets, which offer cost-effectiveness, availability, and control over data generation. By simulating realistic camera movements, lighting conditions, and scene content, synthetic datasets can provide a rich and diverse source of training data \cite{jiang2020example, wang2023jaws, wang2024dancecamera3d, burelli2015adaptive}. 

However, the generalizability of models trained on synthetic data to real-world scenarios remains an open question. Several studies have explored the use of synthetic datasets for camera trajectory generation, including \cite{yang2024direct, xian2023neural, wu2023secret, yu2023automated}. While these studies have demonstrated promising results, further research is needed to evaluate the limitations and biases associated with synthetic data. It is crucial to investigate factors such as the realism of synthetic data, the diversity of training scenarios, and the domain gap between synthetic and real-world data to ensure the effectiveness of models trained on synthetic datasets. In the following, we introduce some of the commonly used synthetic datasets and their applications in camera trajectory generation.

\textbullet \textbf{ Batteries, camera, action! \cite{bonatti2021batteries}:} The dataset used in this study, comprises 200 video clips generated within the AirSim photo-realistic simulator. These clips feature a diverse range of aerial shots parameterized by spherical coordinates and annotated using minimal perceptual units for shot variations. Semantic scores for 15 descriptors, such as "calm" or "exciting," were obtained through crowd-sourced pairwise comparisons involving 500 participants. The dataset's design emphasizes perceptual and cinematic relevance, facilitating the creation of a semantic control space for mapping descriptors to camera trajectory parameters. This dataset was validated across simulated and real-world scenarios to ensure robustness and generalizability.

\textbullet \textbf{ CCD \cite{jiang2024cinematographic}:} The CCD dataset, is a synthetic collection designed for virtual cinematography, featuring 25,000 sequences with over 4.5 million frames and 200,000 textual annotations. These annotations describe key cinematic parameters such as shot angles, scales, and view directions, enabling precise control over static, dynamic, and orbit-based camera movements across diverse speeds like slow motion and fast-paced sequences. It provides balanced coverage of cinematic styles, making it valuable for training machine learning models. However, its synthetic nature limits real-world applicability, as it omits dynamic multi-subject interactions, broader narrative contexts, and emotional depth. Textual annotations lack vocabulary richness, and stationary subjects restrict learning intricate camera-subject interactions, reducing adaptability to complex, real-world filmmaking scenarios requiring creative and narrative flexibility.

\subsection{Real Datasets}
Real datasets are critical in training camera trajectory generation models by providing authentic movement patterns that capture the subtle dynamics and physical constraints inherent in real-world camera operations. Unlike synthetic data, real datasets incorporate natural camera behaviors, scene-specific constraints, and cinematographic principles that emerge from human operators' expertise and practical filming considerations. While some datasets focus on high-level cinematographic features such as shot types, camera angles, and motion categories \cite{bruckert2023look}, this section specifically examines datasets that provide precise camera trajectories through exact position and orientation data for each frame of video clips.

\textbullet \textbf{ RealEstate10k \cite{zhou2018stereo}:} The RealEstate10k dataset introduced in 2018, derived from over 7,000 curated real estate video clips on YouTube. These videos, ranging from 1 to 10 seconds in duration, capture both indoor and outdoor scenes, with precise metadata including camera position, orientation, and field of view for each frame. The dataset was created through a four-stage pipeline, leveraging manual selection, motion estimation techniques like ORB-SLAM2 \cite{mur2017orb}, for optimization, and final filtering for quality assurance. Advantages include its substantial scale, diversity in scene types, and smooth camera movements, which enhance its utility for training camera trajectory models. However, limitations exist, such as its focus on simple, static camera motions typical of real estate videos, lack of semantic descriptions for camera actions, and restricted environmental diversity, excluding natural or urban settings. Furthermore, its suitability for generating complex or dynamic movements, such as those involving subject interactions or rapid changes. 

\textbullet \textbf{ Example-Driven \cite{jiang2020example}:} The dataset introduced by Hongda Jiang et al. (2020), referred to as the Cinematic Feature Dataset, underpins their development of a novel camera motion controller for virtual cinematography. This dataset comprises a combination of synthetic and real film data, capturing essential cinematic features such as camera poses, character configurations, and dynamic interactions across diverse scenes. The dataset's strengths lie in its detailed annotation and its utility in learning complex cinematographic patterns applicable to two-character interactions. However, its limitations include a focus on simplified scenes with a maximum of two characters and the lack of representation for high-frequency camera movements or background motion dynamics. 

\textbullet \textbf{ Augmented RealEstate \cite{wang2024motionctrl}:} The paper authored by Zhouxia Wang et al. (2024) introduces two datasets, the augmented-RealEstate10K. The augmented-RealEstate10K dataset includes over 60,000 videos with annotated camera poses, supplemented by synthesized captions using Blip2. This dataset aids camera motion control but is limited by its narrow domain diversity.

\textbullet \textbf{ DCM \cite{wang2024dancecamera3d}:} The paper authored by Zixuan Wang et al. (2024) introduces the DCM (Dance-Camera-Music) dataset, the first of its kind to integrate 3D camera movement with dance motion and music audio. This dataset includes 108 paired sequences from the anime community, spanning 3.2 hours across four music genres and offering rich annotations for camera keyframes, dance joints, and audio features. By providing synchronized camera trajectories and music-dance alignments. Its advantages include the inclusion of diverse shot types and human-centric camera characteristics. However, it faces limitations, such as the reliance on animator-edited data, which may restrict spontaneity, and challenges in generalizing from anime contexts to real-world settings.

\textbullet \textbf{ E.T. \cite{courant2025exceptional}:} The E.T. (Exceptional Trajectories) dataset is a significant resource for text-to-camera trajectory generation, derived from the CMD dataset \cite{bain2020condensed}. It features 115,000 samples from 16,210 unique scenes, totaling over 11 million frames and 120 hours of cinematic footage. Each sample includes synchronized camera and subject trajectories, with textual captions describing both camera motion and motion relative to the subject. Unlike synthetic datasets, E.T. is based on real movie footage, capturing complex 6 degree of freedom movements and offering a rich vocabulary of over 1,000 words. However, it suffers from imbalances favoring simple motions, lacks professional cinematic terminology, and is limited to single-human subjects without contextual details like subject attributes and environmental factors. These limitations reduce its utility for advanced, real-world filmmaking applications.

In summary, the datasets discussed provide diverse approaches to addressing challenges in camera trajectory generation, each tailored to specific applications and methodologies. These datasets vary in scale, composition, and the types of trajectories they capture, ranging from synthetic sequences with detailed parameterization to real-world datasets emphasizing diversity and realism. While some datasets prioritize control and repeatability, others focus on naturalistic motion and broader applicability. In the following Table \ref{tab:dataset_comparison}, we present a comparative analysis of these datasets, highlighting their key features and differences to provide an overview of their contributions and can not used for various research objectives.

\newcommand{\rowheightdataset}{5ex} 

\newcommand{\cmark}{\ding{51}}%
\newcommand{\xmark}{\ding{55}}%

\begin{table*}%
\caption{Dataset Comparison}
\label{tab:dataset_comparison}
\begin{center}
\resizebox{0.9\textwidth}{!}{%
\begin{tabular}{lcccccccccc}
  \toprule
  Dataset & \#Samples & \#Frames & \#Hours & Domain & \makecell[c]{Character\\Traj.} & \makecell[c]{Camera\\Traj.} & \#Vocabulary & Prompt & \makecell[c]{Dataset\\Link} \\ \midrule 

  \rule{0pt}{\rowheightdataset} \makecell[l]{E.T.\\\cite{courant2025exceptional}} & 115K & 11M & 120 H & Real / Movie & \makecell[c]{YES\\(115K)} & \makecell[c]{YES\\(230K)} & 1790 & \ding{51} & \href{https://www.lix.polytechnique.fr/vista/projects/2024_et_courant}{Link}\\  

  \rule{0pt}{\rowheightdataset} \makecell[l]{DCM\\\cite{wang2024dancecamera3d}} & 108 & 345K & 3.2 H & Synthetic / Dance & NO & YES & NO & \ding{55} & \href{https://github.com/Carmenw1203/DanceCamera3D-Official/tree/master/dataset}{Link}\\

  \rule{0pt}{\rowheightdataset} \makecell[l]{CCD\\\cite{jiang2024cinematographic}} & 25K & 4.5M & 50 H & Synthetic & NO & \makecell[c]{YES\\(25K)} & 48 & \ding{51} & \href{https://github.com/jianghd1996/Camera-control}{Link}\\
  
  \rule{0pt}{\rowheightdataset} \makecell[l]{\cite{bonatti2021batteries}} & 200 & \makecell[c]{NA.} & <1 H & \makecell[c]{Synthetic /\\Semantic Trajectory} & NO & \makecell[c]{NA.} & \makecell[c]{NA.} & \ding{51} & \makecell[c]{NA.}\\

  \rule{0pt}{\rowheightdataset} \cite{jiang2020example} & 2.16M & 86M & \makecell[c]{NA.} & \makecell[c]{10\% Real (Movies)\\90\% Synthetic} & NO & YES & NO & \ding{55} & \makecell[c]{NA.}\\
  
  \rule{0pt}{\rowheightdataset} \makecell[l]{RealEstate10K\\\cite{zhou2018stereo}} & 7K & 11M & 121 H & Real / YouTube & NO & YES & NO & \ding{55} & \href{https://tinghuiz.github.io/projects/mpi/}{Link}\\

  \bottomrule
\end{tabular}
}

\end{center}
\bigskip\centering
\footnotesize\emph{Sources:} \cite{courant2025exceptional, wang2024dancecamera3d, jiang2024cinematographic, jiang2020example, zhou2018stereo, wang2024motionctrl, bonatti2021batteries} 
\end{table*}%

\section{Limitations and Future Direction} \label{limitations_section}
Automated camera trajectory generation systems are a critical component of virtual cinematography and related fields. However, existing approaches face significant challenges that limit their applicability and effectiveness in real-world scenarios. This section outlines the key limitations of current methodologies and proposes future directions for advancing research and practical applications in this domain.

\subsection{Limited Availability and Diversity of Datasets}
The progress of automated camera trajectory generation is hindered by the lack of comprehensive and diverse datasets. Most available datasets, as pointed in Section \ref{datasets_section}, focus on narrow scenarios or predefined settings, limiting their ability to generalize to broader use cases. The majority of these datasets fail to capture complex, dynamic environments or incorporate detailed annotations for advanced cinematic properties such as framing, timing, or motion. Additionally, data collection processes are often resource-intensive, involving substantial technical and financial investments. This scarcity of high-quality datasets constrains the training and evaluation of machine learning models, thereby impeding the development of robust, real-world-ready systems.

\subsection{Computational Complexity in High-Dimensional Models}
Optimization-based methods for camera trajectory generation often involve high-dimensional search spaces, such as 7-DOF \cite{Christie2008camera}. While these models provide precise and detailed control over camera movements, their computational requirements are prohibitively high, especially for real-time applications. The iterative processes required to explore such large solution spaces lead to significant delays, making these methods impractical for time-sensitive scenarios \cite{bonatti2020autonomous}. Similarly, when employing neural network models for camera trajectory generation, it is crucial to ensure that these models are lightweight and efficient, as they are often intended for deployment on embedded devices with limited computational resources.

\subsection{Rigidity of Rule-Based Systems}
Rule-based methods are widely appreciated for their adherence to established cinematic principles \cite{christie2009camera, chen2014autonomous}. However, their inherent rigidity poses significant challenges in dynamic and creative contexts. These systems rely on static, predefined rules that limit their adaptability to novel scenarios or evolving artistic requirements \cite{kennedy2002planning}. When confronted with situations that deviate from their encoded heuristics, rule-based approaches struggle to produce visually coherent and contextually relevant outputs \cite{he1996virtual}. The lack of flexibility also restricts their ability to innovate or accommodate user-driven customization, which is increasingly demanded in professional and amateur filmmaking environments. There remains a notable absence of hybrid systems capable of leveraging contemporary heuristics while delivering robust and accurate results in novel scenarios.

\subsection{Challenges in Dynamic Environments}
Handling dynamic environments, such as those involving moving subjects, obstacles, changing lighting conditions, or potential occlusions, remains a significant challenge for automated systems. Most existing methods assume static or predictable scenes, which limits their applicability to complex, real-world scenarios like sports, live events, or outdoor filmmaking. In these settings, cameras must continuously adapt to evolving conditions, ensuring smooth movements, collision avoidance, occlusion avoidance, and adherence to cinematic principles. Despite the advancements in the field \cite{burg2021real, liu2017planning}, existing systems frequently struggle to seamlessly integrate these requirements, resulting in disruptions to visual quality, such as obstructed views or reliance on manual intervention.

\subsection{Insufficient Integration of Aesthetic Objectives}
While technical accuracy is a focus of most camera trajectory generation systems, the integration of aesthetic principles is often neglected. Many systems prioritize parameters such as stability and framing precision while ignored critical artistic elements like rhythm, emotion, and storytelling. This oversight results in outputs that are technically sound, but, lack the emotional and narrative depth required for professional-grade cinematography. Bridging this gap between technical execution and artistic intent is crucial for advancing the field and meeting the expectations of modern audiences.

\subsection{Camera Trajectory is More than a Numerical Sequence}
Camera trajectory not only defines how the camera moves within a real or virtual environment but also serves as a powerful tool to evoke emotions and guide the viewer's attention \cite{bonatti2021batteries}. By carefully controlling motion, orientation, and timing, it establishes narrative flow, enhances dramatic effects, and conveys mood \cite{sudabathula2024emotion}. These neglected aspects are essential in storytelling, shaping how audiences perceive and interact with visual content. However, there is a clear lack of integrated camera trajectory generation systems that holistically address these dimensions. Critical areas such as the representation of such systems, the availability of high-quality datasets, the development of robust generative models, and the establishment of comprehensive evaluation metrics remain under explored and warrant significant attention.

Future research can enhance automated camera trajectory generation by advancing semantic understanding, expanding multi-subject support, improving dataset diversity, refining evaluation metrics, and exploring long-term opportunities. 



\section{Conclusion}
The field of automated camera trajectory generation has witnessed remarkable advancements, drawing from a diverse spectrum of methodologies such as rule-based systems, optimization techniques, machine learning, and hybrid approaches. These methods have collectively tackled challenges related to computational efficiency, adaptability, and cinematic quality. By systematically reviewing key contributions and methodologies within this survey, we have demonstrated how these approaches address core challenges and contribute to the field's evolution. Specifically, we have synthesized insights from foundational principles and SOTA advancements, providing a cohesive understanding of existing solutions and emerging trends.

One of the most active areas of research in this field is the application of machine learning methods, which have emerged as a hot topic due to their adaptability and capacity for learning complex cinematic patterns. Machine learning approaches, particularly those leveraging deep learning and generative models, enable the synthesis of flexible, creative, and context-aware \& multi-domain \cite{wang2024dancecamera3d, courant2025exceptional} camera trajectories. These models are increasingly capable of integrating aesthetic principles and responding to dynamic environments, offering transformative potential for both professional filmmaking and interactive applications.

Challenges in automated cinematography, as discussed in \ref{limitations_section}, include limited dataset diversity, which hampers models' ability to generalize across real-world scenarios, and underrepresentation of dynamic environments, multi-subject interactions, and cinematic attributes like rhythm and storytelling. Future research must address these limitations by enhancing dataset diversity, utilizing synthetic generation techniques, bridging the gap between synthetic and real-world data, and leveraging advanced neural architectures such as visual-language models for generating cinematographic specific description for existing ones. Real-time systems with adaptive behaviors, multi-subject interactions, and adherence to cinematic principles, combined with emerging technologies like 3D scene modeling \cite{liu2024reconx, zhang2024monst3r}, hold the potential to deliver solutions that are both technically proficient and artistically compelling, revolutionizing filmmaking and immersive media.

%
%
%
%

\bibliographystyle{ACM-Reference-Format}
\bibliography{sample-bibliography}

\end{document}